\documentclass[acmtog, authorversion]{acmart}
\AtBeginDocument{%
  \providecommand\BibTeX{{%
    \normalfont B\kern-0.5em{\scshape i\kern-0.25em b}\kern-0.8em\TeX}}}

\copyrightyear{2024}
\acmYear{2024}
\setcopyright{acmlicensed}\acmConference[SA Conference Papers '24]{SIGGRAPH Asia 2024 Conference Papers}{December 3--6, 2024}{Tokyo, Japan}
\acmBooktitle{SIGGRAPH Asia 2024 Conference Papers (SA Conference Papers '24), December 3--6, 2024, Tokyo, Japan}
\acmDOI{10.1145/3680528.3687704}
\acmISBN{979-8-4007-1131-2/24/12}

\usepackage{multirow}
\usepackage{url}

\usepackage{xspace}

\newcommand{\eg}{\textsl{e.g.}~}

\newcommand{\avatarStage}{MultiFLARE}

\newcommand{\numVideos}{N}
\newcommand{\flameshape}{\beta}
\newcommand{\flamepose}{\theta}
\newcommand{\flameexpr}{\psi}
\newcommand{\intrinsicMLP}{\mathcal{M}}
\newcommand{\lightMLP}{\mathcal{L}}
\newcommand{\albedo}{\rho}
\newcommand{\roughness}{r}
\newcommand{\specularInt}{k}
\newcommand{\diffuseShading}{l_d}
\newcommand{\specularShading}{l_s}
\newcommand{\n}{n}
\newcommand{\refvec}{\omega}

\newcommand{\w}{0}
\begin{document}

\title{SPARK: Self-supervised Personalized Real-time Monocular Face Capture}


\author{Kelian Baert}
\affiliation{%
  \institution{Technicolor Group}
  \city{Paris}
  \country{France}
}
\affiliation{%
  \institution{University Rennes}
  \city{Rennes}
  \country{France}
}
\email{kelian.baert@technicolor.com}

\author{Shrisha Bharadwaj}
\affiliation{%
  \institution{Max Planck Institute for Intelligent Systems}
  \city{Tübingen}
  \country{Germany}}
\email{shrisha.bharadwaj@tuebingen.mpg.de}

\author{Fabien Castan}
\affiliation{%
  \institution{Technicolor Group}
  \city{Paris}
  \country{France}
}
\email{fabien.castan@technicolor.com}

\author{Benoit Maujean}
\affiliation{%
  \institution{Technicolor Group}
  \city{Paris}
  \country{France}
}
\email{benoit.maujean@technicolor.com}

\author{Marc Christie}
\affiliation{%
  \institution{University Rennes, IRISA, CNRS, Inria}
  \city{Rennes}
  \country{France}}
\email{marc.christie@irisa.fr}

\author{Victoria F. Abrevaya}
\affiliation{%
  \institution{Max Planck Institute for Intelligent Systems}
  \city{Tübingen}
  \country{Germany}}
\email{victoria.abrevaya@tuebingen.mpg.de}

\author{Adnane Boukhayma}
\affiliation{%
  \institution{Inria, University Rennes, IRISA, CNRS}
  \city{Rennes}
  \country{France}}
\email{adnane.boukhayma@gmail.com}

\renewcommand{\shortauthors}{Baert et al.}

\begin{abstract}
    Feedforward monocular face capture methods seek to reconstruct posed faces from a single image of a person. Current state of the art approaches have the ability to regress parametric 3D face models in real-time across a wide range of identities, lighting conditions and poses by leveraging large image datasets of human faces. These methods however suffer from clear limitations in that the underlying parametric face model only provides a coarse estimation of the face shape, thereby limiting their practical applicability in tasks that require precise 3D reconstruction (aging, face swapping, digital make-up,...).
    In this paper, we propose a method for high-precision 3D face capture taking advantage of a collection of unconstrained videos of a subject as prior information.
    Our proposal builds on a two stage approach. We start with the reconstruction of a detailed 3D face avatar of the person, capturing both precise geometry and appearance from a collection of videos. We then use the encoder from a pre-trained monocular face reconstruction method, substituting its decoder with our personalized model, and proceed with transfer learning on the video collection. Using our pre-estimated image formation model, we obtain a more precise self-supervision objective, enabling improved expression and pose alignment. This results in a trained encoder capable of efficiently regressing pose and expression parameters in real-time from previously unseen images, which combined with our personalized geometry model yields more accurate and high fidelity  mesh inference.   

    Through extensive qualitative and quantitative evaluation, we showcase the superiority of our final model as compared to state-of-the-art baselines, and demonstrate its generalization ability to unseen pose, expression and lighting. 
\end{abstract}




\begin{teaserfigure}
 \includegraphics[width=\textwidth]{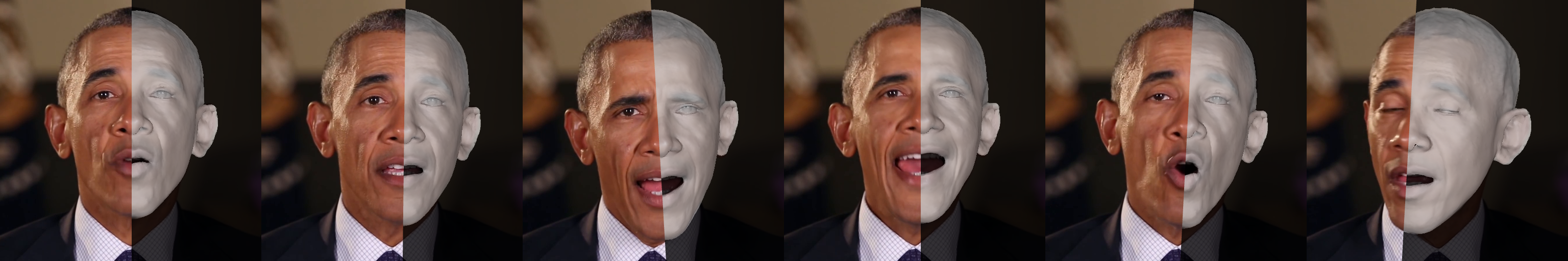}
 \caption{After a personalization step, SPARK can reconstruct accurate faces with expressive details on unseen images \textbf{in real-time}.}
 \Description{Several images of Barack Obama's face with reconstructed geometry. Each image is cut in half, with the left side showing the input image and the right side showing a render of the estimated geometry.}
 \label{fig:teaser}
\end{teaserfigure}

\received{19 May 2024}
\received[revised]{6 September 2024}
\received[accepted]{?}

\maketitle

%%%%%%%%%%% sections %%%%%%%%%%
\section{Introduction}

3D facial performance capture is a key component in several applications, including immersive telepresence in AR or VR and visual effects for the entertainment industry. Producing high-quality results, however, often requires large financial, time and resource investments. 
This involves not only expensive 3D capture equipment  \cite{beelerHighqualitySingleshotCapture2010, debevec2000acquiring}, precise marker-based tracking systems~\cite{bennett2014adopting} or head-mounted displays~\cite{brito2019recycling}, but also extensive hours of capture sessions from the actor. Marker-less capture setups are promising solutions for simplifying this pipeline, however high-quality results still rely on complex rigs~\cite{helman2020ilm} or large, personalized training datasets~\cite{laine2017production, wu2018deep}.  

At the other end of the spectrum are 3D reconstruction methods that can operate on images or videos from affordable consumer-grade hardware. The main idea is to use statistical models of the 3D face --- so-called 3D Morphable Models (3DMMs) --- which are fitted to RGB images or 2D landmarks using optimization-based \cite{andrusFaceLabScalableFacial2020, zielonkaMICAMetricalReconstruction2022} or learning-based methods \cite{feng2021learning, danecek2022EMOCA, retsinasSMIRK3DFacial2024}. The prior knowledge from the statistical model helps to overcome the ill-posed nature of the problem, and the development of learning-based techniques has enabled unprecedented robustness to pose, illumination and occlusions. However, this comes at the cost of lower geometric quality, providing only a coarse approximation of the shape and expression which falls well short of high-end systems. 

Neural head avatars were recently proposed to address some of the limitations of 3DMM-based methods. The goal here is to train a model from a single monocular video, that can afterwards be animated with novel expressions \cite{zheng2022imavatar, zheng2023pointavatar, Grassal_2022_nha}  and optionally environments \cite{bharadwaj2023flare}. The key idea is to harness advances in neural rendering~\cite{munkberg2022extracting, hasselgren2022shape} which allow to better model geometric and appearance details. While this results in images of higher quality, the estimated 3D geometry still remains behind production systems. Furthermore, the computational costs of such approaches prevent them from running in real-time or even close to real-time contexts.

In this paper we aim to bridge the gap between these two worlds by answering the following: \textbf{given a \emph{collection} of unconstrained videos of a person as prior information, can we build a personalized real-time tracker that regresses higher quality geometry?}
Video collections offer a valuable middle ground between high-end capture systems and consumer-grade solutions. Without requiring complex, non-portable hardware, we can leverage multiple videos with a variety of illuminations and potentially more diverse poses to better constrain the problem of 3D face reconstruction.

More specifically, we propose SPARK, a novel method to build \emph{personalized}, high-quality 3D facial models that can be tracked in real-time. Our approach consists of two key steps: (1) constructing a personalized 3D mesh and deformation model from a collection of \emph{in-the-wild videos}, and (2) training a fast, regression-based tracker that can provide accurate reconstructions given a novel image. The challenge here is two-fold. First, we need to design a system with the capacity to distill high-fidelity geometry from contents with very diverse lighting conditions, color tones and appearances. To this end, we leverage recent advances in unsupervised efficient inverse neural rendering (\eg \cite{bharadwaj2023flare, zheng2022imavatar, zielonka2022insta}) and adapt them to the multi-video case.  
Second, we need to recover a high fidelity topologically consistent 3D facial geometry in real time. Rooted in transfer learning, our idea is to consolidate the benefits of state-of-the-art generalizable 3D face capture methods (\eg \cite{danecek2022EMOCA,feng2021learning}) that were trained on a large in-the-wild image corpora, and propose an efficient method for adapting to the specific subject, leveraging the results from the previous step. 

Through quantitative and qualitative evaluations, our experimental results show that we can generalize successfully to unseen lighting, expressions and poses, and are capable of producing accurate geometries in real time. Our method outperforms competing techniques with similar inference times offering, more accurate poses and expressions, together with more expressive meshes.

In summary, our contributions are:
\begin{itemize}
    \item A new method using multiple monocular videos of a person to estimate the 3D shape and range of deformations of a face, with fine geometric details essential for facial expressiveness.
    \item A transfer learning approach that enables fine-tuning a face tracking method for a specific subject using a pre-estimated personalized geometry and appearance model, while still being able to generalize to different lightings, poses and expressions.
    \item Thorough evaluations of our proposed approach, including two new metrics for evaluating the posed geometry without 3D ground truth. 
\end{itemize}

\section{Related Work}

\paragraph{Head avatars from video}

There have been many efforts towards the democratization of animatable 3D head avatar creation via commodity sensors, as opposed to costly traditional capture systems \cite{debevec2000acquiring, beeler2011high, riviere2020single, ghosh2011multiview}. Traditional methods \cite{garrido2016reconstruction, thies2016face2face} utilize statistical models \cite{blanzvetter1999} to reconstruct 3D shape and appearance, but they result in relatively coarse reconstructions. NerFACE~\cite{Gafni_2021_nerface} pioneered in using dynamic neural radiance fields (NeRF) \cite{Mildenhall_2020_nerf} to create head avatars. IMavatar~\cite{zheng2022imavatar} achieves precise geometry recovery by using implicit surfaces, simultaneously learning canonical head geometry and expression deformations. However, methods relying on implicit representations often suffer from inefficiencies in training and rendering. PointAvatar~\cite{zheng2023pointavatar} adopts a similar deformation model but uses an explicit point cloud representation, facilitating faster rasterization and improved image quality. Recently, several approaches~\cite{Gao2022nerfblendshape, zielonka2022insta, Xu2022manvatar} have leveraged InstantNGP~\cite{mueller2022instant} to accelerate radiance field queries, enabling avatar reconstruction within 5 to 20 minutes. Contemporary work proposes to build avatars through Gaussian Splatting (3DGS) \cite{kerbl20233d} by anchoring gaussians to a coarse underlying control geometry \cite{ma20243d,xu2023gaussianheadavatar,qian2023gaussianavatars}. However, there is no clearly established mechanism to robustly extract topologically consistent detailed meshes from these models. Neural Head Avatar \cite{Grassal_2022_nha} creates mesh-based avatars containing full head and hair geometry. Despite this, the resulting geometry is relatively coarse, with many details rendered in the texture space. The recent FLARE \cite{bharadwaj2023flare} constructs high-quality mesh-based avatars within 15 minutes, while decomposing appearance into albedo, roughness, and extrinsic illumination. We build on this method in our avatar reconstruction in-the-wild stage, by extending it to a novel multi monocular video setting, in addition to introducing several improvements to enhance efficiency and expressiveness.

\paragraph{3D monocular face capture}

Model-free methods can learn to directly infer meshes \cite{deng2020retinaface,feng2018prn,ruan2021sadrnet, dou2017endtoend,alp2017densereg,abrevaya2019decoupled,Jung2021,Sela2017,Szabo2019,Wei2019,Zeng2019_DF2Net,Wu2020}, voxels \cite{jackson2017large}, or Signed Distance Functions \cite{yenamandra2021i3dmm}. However, they require extensive 3D training data, and they can also suffer from limited expressiveness and generalization due to the synthetic/real domain gap \cite{dou2017endtoend,Sela2017,Zeng2019_DF2Net}, with many of them relying on synthetic training data.
3DMMs (\eg BFM \cite{paysan20093d}, FaceWarehouse \cite{cao2013facewarehouse}, FLAME \cite{FLAME:SiggraphAsia2017}) can be fitted to images through test-time analysis-by-synthesis optimization \cite{aldrian2012inverse, Bas2017fitting,Blanz2002,Koizumi2020_UMDFA,Ploumpis2020,blanzvetter1999,romdhanivetter2005,li2013realtime, cao2014displaced,garrido2016reconstruction,thies2015realtime,thies2016face2face,thies2016facevr,gerig2018morphable}, but these are not useful for time-sensitive applications.
Recent deep learning based models offer robust and fast inference, either in their supervised \cite{cao2015real, AnhTran2017,tran2018extreme,chang2018expnet,guo2020towards,kim2018inverse,richardson2016synthetic,Zhu2016_3DDFA, zielonkaMICAMetricalReconstruction2022, zhang2023accurate} or weakly/self-supervised \cite{deng2019accurate,Liu2017,RingNet:CVPR:2019,tewari17MoFA, tewari2018self,tewari2019fml,feng2021learning,shang2020self,yang2020facescape,abrevaya2020cross,alp2017densereg,wood2022denselandmarks,genova2018unsupervised, danecek2022EMOCA} forms. Most notably, recent self-supervised models DECA \cite{feng2021learning} and EMOCA \cite{danecek2022EMOCA} offer arguably state-of-the-art animatable geometry inference performance. Their encoders predict 3DMM, camera and spherical harmonics coefficients used to render and compare a morphable geometry to images in the wild for large scale training. Despite their additional mesh detail regression, they still showcase expression generalization and alignment issues, and their generic FLAME \cite{FLAME:SiggraphAsia2017} based geometry can lack expressiveness and fidelity. To remedy these issues, and in contrast with  existing literature, we propose to personalize such models using solely a few monocular videos.  

\section{Method}

\begin{figure*}[t]
    \includegraphics[width=\linewidth]{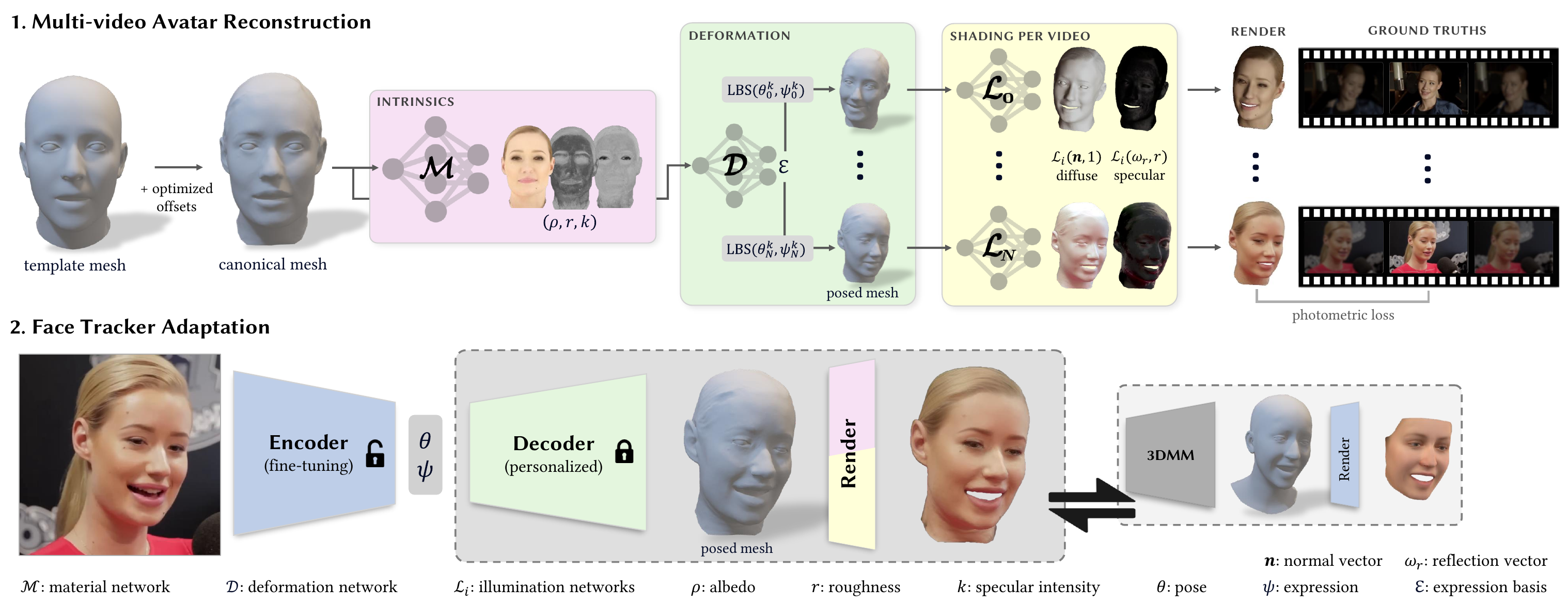}
    \caption{Illustration of our-two stage adaptation process. In stage 1, we rely on a collection of different video sources of the same person to build a personalized geometry decoder through inverse rendering. In stage 2, the 3DMM of a generalizable feedforward face capture network is swapped with the new decoder, and the encoder is tuned by reconstructing the same adaptation video frames leveraging the pre-estimated reflectance function for each video.}
    \label{fig:method}
    \Description{A diagram showcasing the architecture of our two-stage method. The top part details the avatar reconstruction stage, while the bottom part shows the face tracker adaptation.}
\end{figure*}

At inference, our system takes as input an image of a subject and generates a 3D face reconstruction including pose and expression in real time. At training time, our work follows a two-stage approach as illustrated in Fig.~\ref{fig:method}. First we leverage an unconstrained collection of video sequences of the same subject to build a personalized decoder using inverse rendering in a stage called \avatarStage{}, in which we disentangle the color into illumination and intrinsic material (see top row in Fig.~\ref{fig:method}). Second, we adapt a generalizable feedforward 3D face capture model by replacing its decoder with the custom \avatarStage{} geometry model, and tuning its encoder on the video collection using the pre-computed reflectance functions. In the following, we detail these two stages.

\subsection{\avatarStage{}: Inverse rendering from a video collection}
\label{sec:multiflare}

Feedforward face reconstruction methods leverage a generic statistical model for geometry and appearance, serving as a differentiable image formation layer for image based self-supervised learning. We replace this with a new differentiable decoder built efficiently from our adaptation data through inverse rendering. Conjointly, the inverse rendering allows us to estimate a reflectance function per adaptation sequence. We detail in the following the optimization leading to these personalized geometry and reflectance functions.   

\paragraph{Geometry and Deformation}

We use a triangular mesh as our geometry representation and build on the FLAME 3DMM~\cite{FLAME:SiggraphAsia2017}. The personalized shape is obtained in canonical space by optimizing the template vertices, similarly to PointAvatar~\cite{zheng2023pointavatar} and FLARE~\cite{bharadwaj2023flare}. A given point $x_c$ in canonical space is then deformed for pose and expression using the following equation: 
\begin{eqnarray}
    \lefteqn{FLAME(x_c, \mathcal{P}, \mathcal{E}, \mathcal{W}, \flamepose, \flameexpr) =} \nonumber \\
    & & LBS( x_c + B_P(\flamepose; \mathcal{P}) + B_E(\flameexpr; \mathcal{E}), J(\flameshape), \flamepose, \mathcal{W})
    \label{eq:deformation}
\end{eqnarray}
where LBS is the standard linear blend-skinning function, $J$ is the joint regressor used to compute joint locations from mesh vertices, $\mathcal{W}\in\mathbb{R}^{n_V\times n_j}$ are per-vertex blend-skinning weights, $B_P(\cdot)$ and $B_E(\cdot)$ compute pose and expression offsets given pose correctives $\mathcal{P}$ and expression blendshapes $\mathcal{E}$. $\flameshape$, $\flamepose$ and $\flameexpr$ are respectively pre-computed FLAME parameters for shape, pose and expression. We follow IMavatar \cite{zheng2022imavatar} and FLARE in personalizing the expression blendshapes by training a deformation network $\mathcal{D}$ that predicts the expression basis $\mathcal{E}\in\mathbb{R}^{n_e}$ given canonical vertex location $x_c$:
\begin{eqnarray}
    \mathcal{D}(\gamma(x_c)):\mathbb{R}^3\rightarrow \mathcal{E}
    \label{eq:deformer}
\end{eqnarray}
We apply sinusoidal positional encoding $\gamma(\cdot)$ to map canonical positions into a higher dimensional space, enabling the network to model higher frequency variations. $\mathcal{D}$ is initialized in a separate supervised pre-training stage that minimizes $||\mathcal{D}(\gamma (x)) - \mathcal{E}||_2$ at the positions of the canonical FLAME vertices, such that the network initially mimics the FLAME expression basis $\mathcal{E}$. Note that this stage is completed in under a minute, as we are only performing forward and backward passes through $\mathcal{D}$ and not rendering.

We perform remeshing \cite{Botsch2004ARA} of the canonical geometry after a fixed number of training iterations to increase its resolution.

\paragraph{Illumination}
Modelling the appearance is challenging in our setting as we have $\numVideos{}$ videos of each subject with varying illumination conditions. Moreover, the personalized canonical geometry is dependent on the learning of the appearance, as we optimize the geometry through deferred shading. Hence, to handle multiple videos, we disentangle the color into illumination and intrinsic materials such as albedo, roughness and specular intensity. However, our dataset by design has in-the-wild videos that are low-dynamic range and have uncontrolled lighting conditions such as over/under-exposed frames. Thus, it is non-trivial to model the illumination and disadvantageous to represent it using HDRI maps.
To tackle this, we follow FLARE and split the rendering equation into a diffuse term using a Lambertian model and a specular term through the Cook-Torrance microfacet model \cite{cook1982reflectance}, which is evaluated using the neural split-sum approximation. More specifically, the illumination is represented by a single MLP controlled using Integrated Directional Encoding \cite{verbin2022refnerf} (IDE) to mimic different mipmap levels regulated by the surface roughness.
We kindly refer the readers to FLARE \cite{bharadwaj2023flare} for more details and for the sake of simplicity, we denote the neural split-sum approximation (including IDE and the precomputed look-up table \textit{FG-LUT}) as $\lightMLP$, such that:
\begin{align}
     \lightMLP(\n, \mathbf{1}) &= \diffuseShading &
     \lightMLP(\refvec, \roughness)  &= \specularShading
\end{align}
where $\textbf{n}$ is the surface normal, $\refvec$ is the reflection vector, $\roughness$ is the roughness, $\diffuseShading$ represents the diffuse shading and $\specularShading$ represents the specular shading. To adapt to our setting, we extend this to 
$\numVideos{}$ approximations given by $\{\lightMLP^1, \dots, \lightMLP^{\numVideos{}}\}$ and optimize the parameters for each illumination seperately: $\{\diffuseShading^1, \dots \diffuseShading^{\numVideos{}}\}$ and $\{\specularShading^1, \dots \specularShading^{\numVideos{}}\}$.

\paragraph{Intrinsic material properties}
We learn the albedo, roughness and specular intensity through a MLP $\intrinsicMLP$ in canonical space as introduced by FLARE. Contrary to the illumination, which is learned separately for each video, we make the assumption that the person's intrinsic appearance varies little between videos and learn a single set of material properties. $\intrinsicMLP$ takes canonical points ($x_c$) as inputs and predicts the albedo $\albedo$, roughness $\roughness$ and specular intensity $\specularInt$:
\begin{equation}
    \intrinsicMLP(x_c): \mathbb{R}^3 \rightarrow \albedo, \roughness, \specularInt
    \label{eq:material}
\end{equation}
The final color for the $i^{th}$ video is calculated as follows:
\begin{equation}
    C = \albedo \cdot \diffuseShading^i + \specularInt \cdot \specularShading^i
    \label{eq:reflectance}
\end{equation}
Thus, we enforce consistency in the intrinsic materials by design by using the same network $\intrinsicMLP$ to calculate the final color for $\numVideos{}$ illuminations. The learnable illumination, along with shared subject-specific geometry and material properties are optimized through inverse rendering using differentiable rasterization.

%\subsection{Personalizing a generalizable feedforward 3D face capture model}
\subsection{Face Tracker Adaptation}
\label{sec:method_transfer_learning}

Standard 3D face reconstruction networks (\eg \cite{danecek2022EMOCA,feng2021learning,retsinasSMIRK3DFacial2024}) typically consist of an encoder, that we denote $E_\phi$, with parameters $\phi$. They predict shape, pose and expression parameters, among others. Combined with a 3D morphable model and a relatively simple appearance representation, such methods learn self-supervisedly from images through inverse rendering. We wish to benefit from the feedforward prediction capabilities of such encoders, whose large-scale training enables generalization to a large variety of illuminations, poses and expressions, and associate this ability with our superior expressive personalized geometry decoder, as opposed to a vanilla 3DMM. Without loss of generality, we build on EMOCA \cite{danecek2022EMOCA}, which uses the FLAME 3DMM.

From the previous stage, we recover a personalized geometric model, through the identity-specific canonical positions $x_c$ and expression deformation basis $\mathcal{E}$  (Eq.~\ref{eq:deformation}). We also obtain a personalized reflectance function (Eq.~\ref{eq:reflectance}) allowing us to render this custom geometry, with intrinsic materials of the subject's face ($\albedo, \roughness, \specularInt$), along with lighting for the training sequences ($\diffuseShading^i, \specularShading^i$). We now freeze our canonical geometry, appearance and deformation models. We compute the expression basis $\mathcal{E}$ and intrinsic materials $(\rho,r,k)$ according to Eq.~\ref{eq:deformer} and Eq.~\ref{eq:material} respectively.
Whilst our personalized \avatarStage{} geometry model is built starting from FLAME, its final canonical geometry and deformation basis deviate from their original definition due to their personalization (Section \ref{sec:multiflare}). In fact, simply swapping FLAME for our personalized estimations in EMOCA leads to very suboptimal results, as can be seen in Table \ref{tab:results_ablation}.

Hence, inspired by few-shot adaptation literature  (\eg \cite{gao2024clip,zhou2022learning}), we propose to tune the encoder in order to adapt its output latent space ($\flamepose$ and $\flameexpr$) to our new personalized geometry representation. Specifically, we update the weights of the last ResNet block~\cite{heDeepResidualLearning2016} of the backbone and the entire MLP head, for both the coarse shape encoder and the expression encoder of EMOCA. This allows us to adapt to the modified regression task while retaining the general features learned in earlier layers \cite{lee2022surgical}, as our objective can be akin to alleviating output level shift in the regression task. 

\subsection{Training details}

In this section, we provide information regarding the training of both stages of our method. More details are available in our supplemental material.

\subsubsection{MultiFLARE}

\paragraph{FLAME regularization}
Previous methods, namely the aforementioned IMavatar, PointAvatar and FLARE, regularize the deformations using the FLAME values at the nearest vertex, which can be computed for arbitrary mesh topologies and various geometry representations. Our original canonical geometry uses the FLAME topology and only changes at the remeshing step. Thus, we update the FLAME expression basis by projecting the remeshed vertex positions on the original mesh and using barycentric interpolation. We then use those updated per-vertex values directly in place of the nearest neighbors, keeping the same loss formulation. We find this improves stability for learning the deformation network, especially for identities whose canonical geometry deviates heavily from the FLAME template.

\paragraph{Objective function}
Our full objective function is as follows: 
\begin{eqnarray}
    \lefteqn{L = \lambda_{\text{RGB}}L_{\text{RGB}} + \lambda_{vgg}L_{vgg} + \lambda_{\text{mask}}L_{\text{mask}} + \lambda_{\text{FLAME}}L_{\text{FLAME}} +} \nonumber \\
    & & \lambda_{\text{laplacian}}L_{\text{laplacian}} + \lambda_{\text{normal}}L_{\text{normal}} + \lambda_{\text{smooth}}L_{\text{smooth}} + \\
    & & \lambda_r L_r + \lambda_{\text{spec}}L_{\text{spec}} + \lambda_{\text{light}}L_{\text{light}} \nonumber
    \label{eq:loss}
\end{eqnarray}

where $L_{\text{RGB}}$ and $L_{vgg}$ are respectively a photometric L2 loss in log space and a perceptual loss \cite{Simonyan_2014_vgg} between masked ground-truth and rendered image, $L_\text{mask}$ is an L2 loss between ground truth and rasterized binary mask, $L_\text{laplacian}$ is a laplacian smoothness regularizer for the canonical vertices, $L_\text{normal}$ is a cosine similarity loss between neighboring face normals of the canonical geometry. $L_\text{FLAME}$ regularizes the expression blendshapes $\mathcal{E}$ from the deformation network $\mathcal{D}$ using the FLAME basis, $L_r$ and $L_\text{spec}$ enforce a predefined distribution for specular intensity $k$ and surface roughness $r$, $L_\text{light}$ penalizes strong deviations from white light in the diffuse shading and $L_\text{smooth}$ is a smoothness regularization for albedo and roughness. We refer the reader to FLARE for more details on the individual loss functions. We set the relative weights $\lambda_i$ of those terms as follows: $\lambda_{\text{RGB}} = 1.0$, $\lambda_{vgg} = 0.1$, $\lambda_{\text{mask}} = 2.0$, $\lambda_{\text{FLAME}} = 20.0$, $\lambda_{\text{laplacian}}=100.0$, $\lambda_{\text{normal}} = 0.1$, $\lambda_{\text{smooth}} = \lambda_r = \lambda_{\text{spec}} = \lambda_{\text{light}} = 0.01$.

\paragraph{Single-stage training}
We jointly optimize the canonical positions $x_c$, deformation $\mathcal{D}$, material $\mathcal{M}$ and lighting networks ($l_d^i$, $l_s^i$) in a single stage, with the material network equipped from the start with hash-grid encoding \cite{mueller2022instant}. To address the findings of FLARE, wherein the model would overfit to color too quickly resulting in over-smooth shapes, we use \textit{progressive} multi-resolution hash encoding \cite{li2023neuralangelo} and enable the increasingly high-resolution hash levels successively. Through this change, we achieve high quality texture and geometry without having to learn $\mathcal{M}$ and ($l_d^i$, $l_s^i$) from scratch in a second stage. We exponentially decay the learning rate of the canonical vertices at a rate of $\alpha = 0.998$ per iteration. Additionally, we conjointly fine-tune the pre-estimated pose and expression parameters to allow for better alignment for frames where the pre-estimated values are imprecise.

\subsubsection{Face Tracker Adaptation}

\paragraph{Objective function}
The transfer learning stage described in Section~\ref{sec:method_transfer_learning} is performed through gradient descent using the following combined loss, borrowed from the coarse stage training of EMOCA v2 \cite{danecek2022EMOCA, filntisis2023spectre}, and applied to the training video frames of the identity of interest:
\begin{gather} 
    \phi^* = \text{argmin}_\phi \; L(\phi)\\
    L(\phi) = \lambda_{\text{emo}}L_{\text{emo}} + \lambda_{\text{pho}}L_{\text{pho}} + \lambda_{\text{lmk}}L_{\text{lmk}} + \lambda_{\text{eye}}L_{\text{eye}} +\\
    \lambda_{\text{mc}}L_{\text{mc}} + \lambda_{\psi}L_{\psi} + \lambda_{\text{lipr}}L_{\text{lipr}} \nonumber
\end{gather}
where $L_{\text{emo}}$, $L_{\text{pho}}$, $L_{\text{lmk}}$, $L_{\text{eye}}$, $L_{\text{mc}}$, $L_{\psi}$ are respectively the emotion consistency loss, photometric loss, landmarks reprojection loss, relative landmarks-based eye closure loss and mouth closure loss, and expression regularization loss as introduced in EMOCA. $L_{\text{lipr}}$ is the lip reading loss based on \cite{filntisis2023spectre} introduced in EMOCA v2. $\lambda_i$ designates the respective weight of each objective. We remind that the self-supervised rendering loss $L_{\text{pho}}$ uses our personalized geometry model (Eq.~\ref{eq:deformation}) in combination with our reflectance (Eq.~\ref{eq:reflectance}), computed using the corresponding lighting $l_d^i$, $l_s^i$ for each adaptation video.

\subsection{Test-time inference}
\label{sec:inference}

After the encoder adaptation, geometry prediction for any unseen image $I$ of the target identity is obtained by combining the inferred parameters  $(\flamepose^*,\flameexpr^*) = E_{\phi^*}(I)$ with our personalized \avatarStage{} geometry model (Eq.~\ref{eq:deformation}). As our neural geometry model is pre-computed after the personalization stage, we can perform inference in real time on standard GPUs (forward pass through $E_{\phi^*}$, a Resnet-50~\cite{heDeepResidualLearning2016} network).

\section{Experiments}

\begin{figure*}[t]
    \setlength{\tabcolsep}{0.0em} % horizontal padding
    \def\arraystretch{0.0}{ % vertical padding
    \renewcommand{\w}{0.085\linewidth}
    \begin{tabular}[t]{ccccccc}
        Input & DECA & EMOCA & EMOCA-t & SMIRK & SPARK
        & SPARK
        \\
        \includegraphics[width=\w]{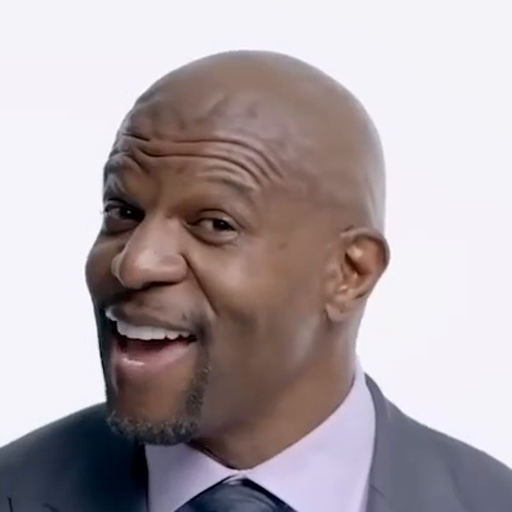} &
        \includegraphics[width=\w]{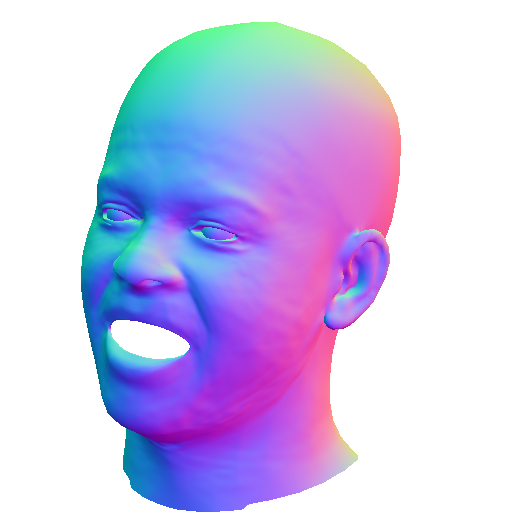} &
        \includegraphics[width=\w]{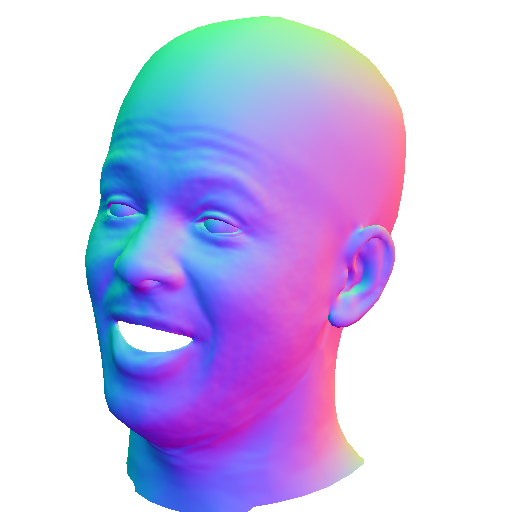} &
        \includegraphics[width=\w]{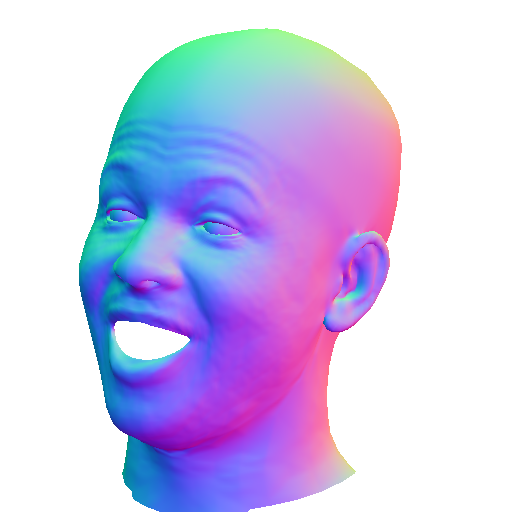} &
        \includegraphics[width=\w]{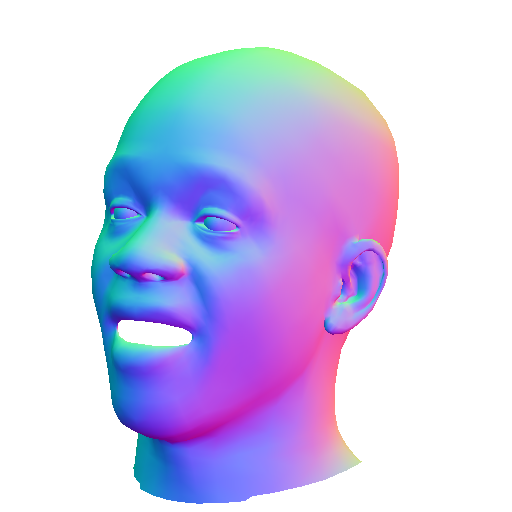} &
        \includegraphics[width=\w]{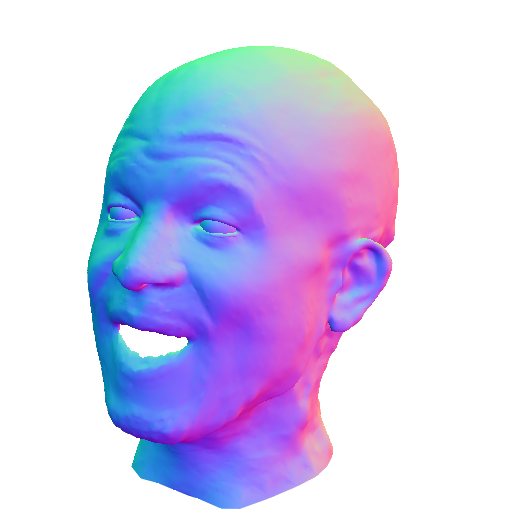}
        & \includegraphics[width=\w]{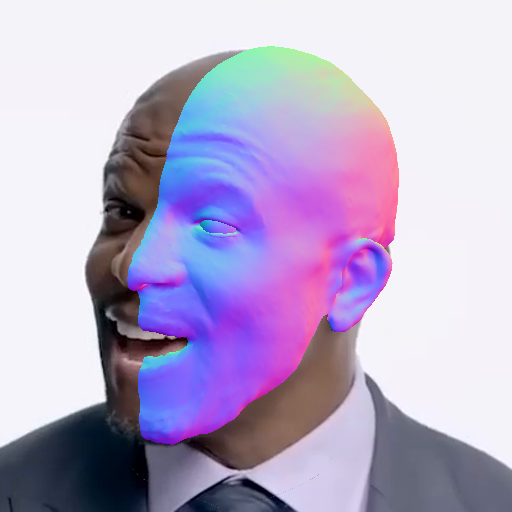}
        \\
        \includegraphics[width=\w]{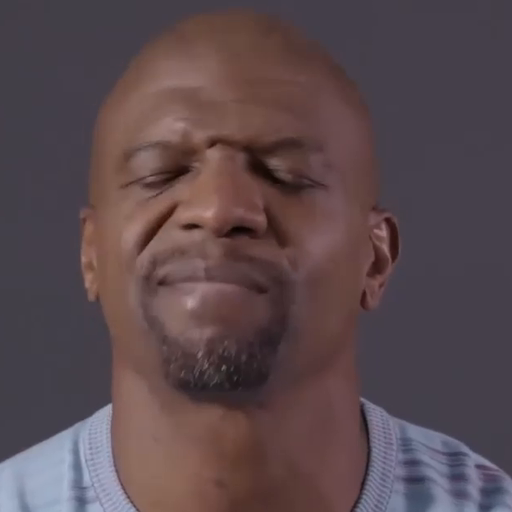} &
        \includegraphics[width=\w]{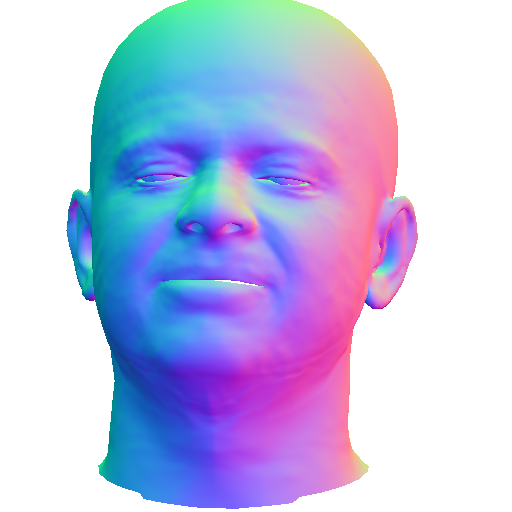} &
        \includegraphics[width=\w]{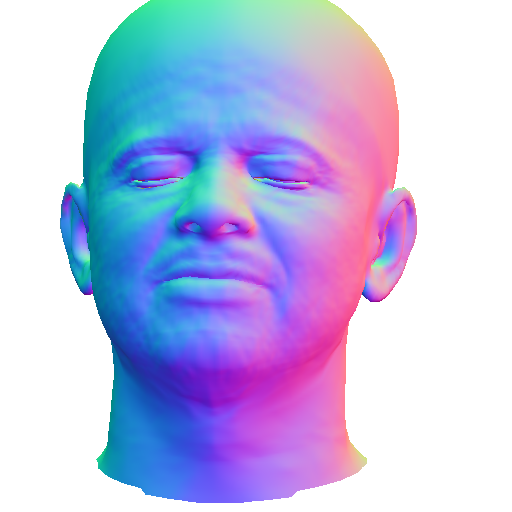} &
        \includegraphics[width=\w]{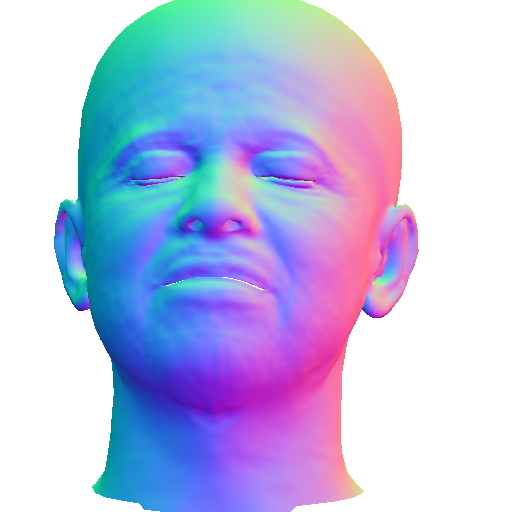} &
        \includegraphics[width=\w]{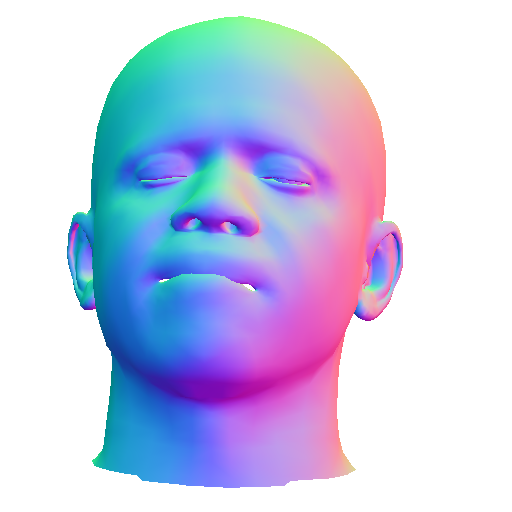} &
        \includegraphics[width=\w]{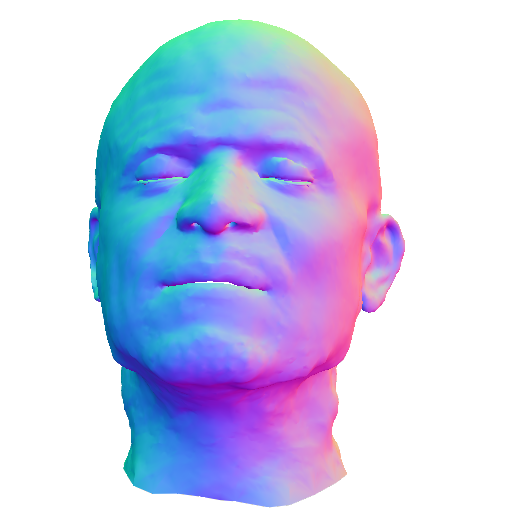}
        & \includegraphics[width=\w]{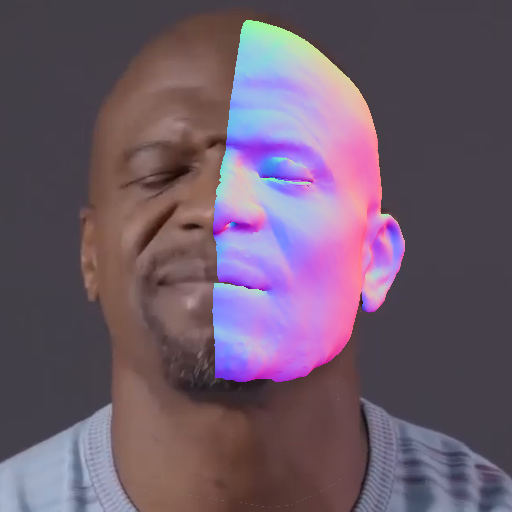}
        \\
        \includegraphics[width=\w]{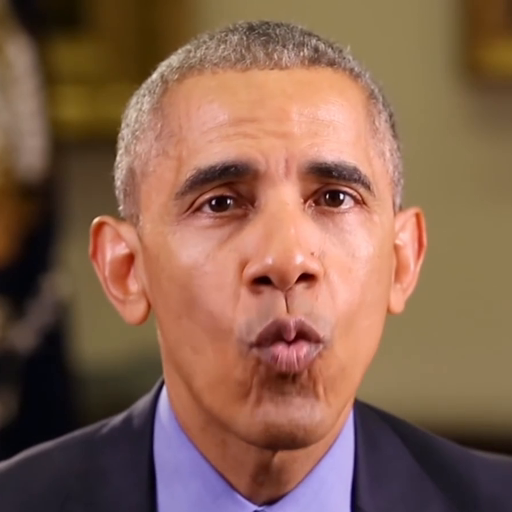} &
        \includegraphics[width=\w]{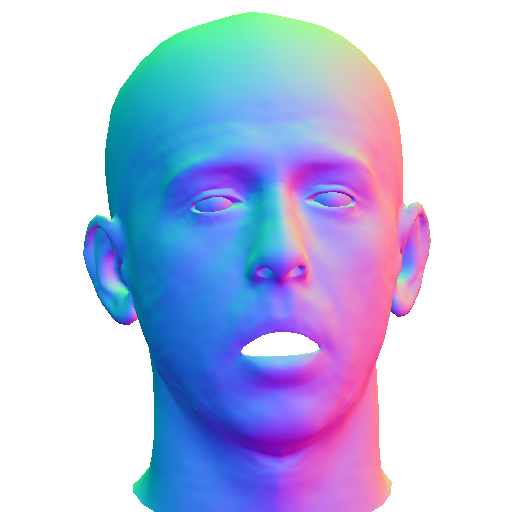} &
        \includegraphics[width=\w]{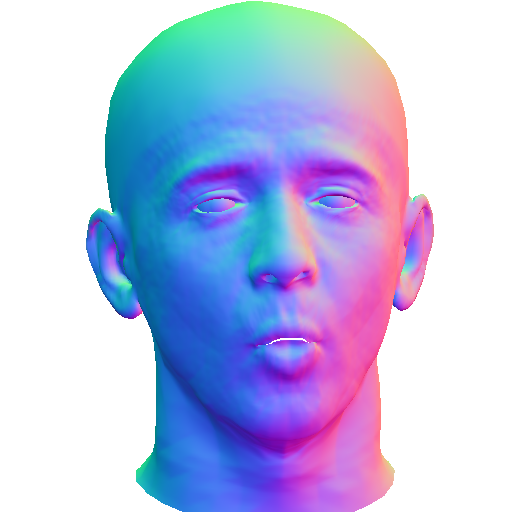} &
        \includegraphics[width=\w]{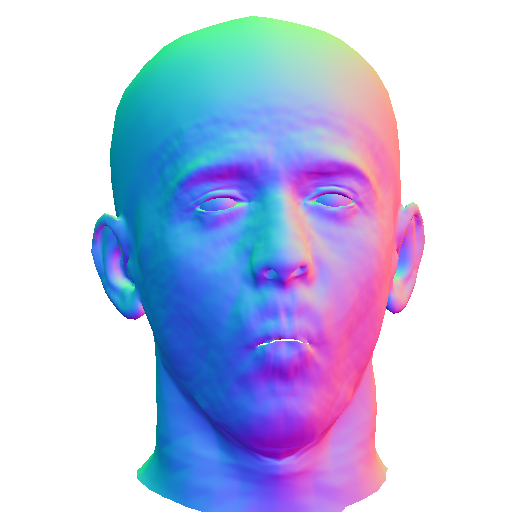} &
        \includegraphics[width=\w]{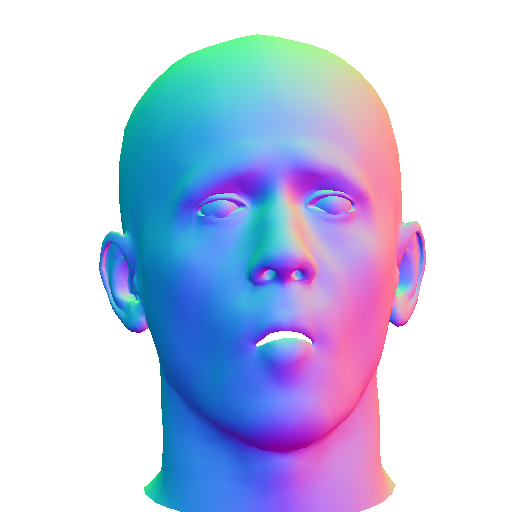} &
        \includegraphics[width=\w]{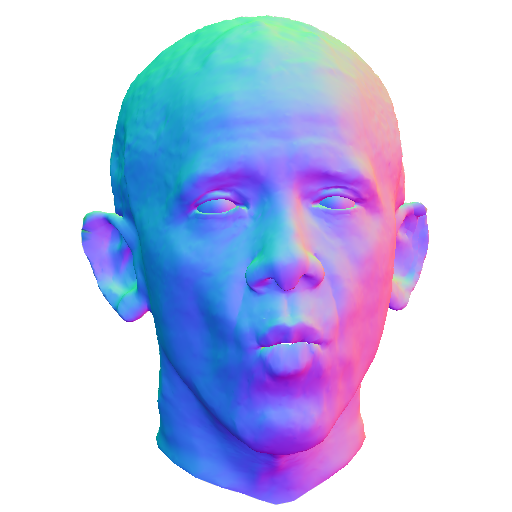}
        & \includegraphics[width=\w]{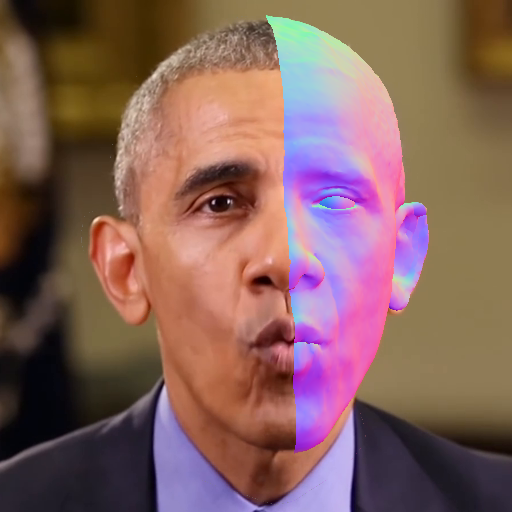}
        \\
        \includegraphics[width=\w]{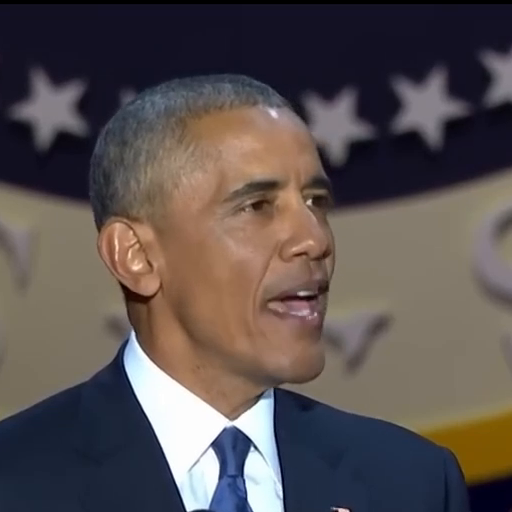} &
        \includegraphics[width=\w]{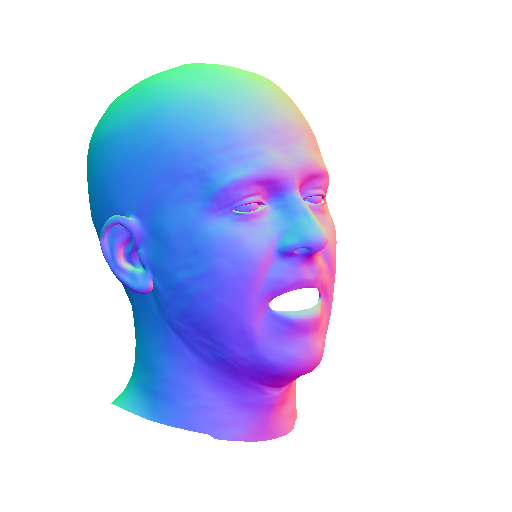} &
        \includegraphics[width=\w]{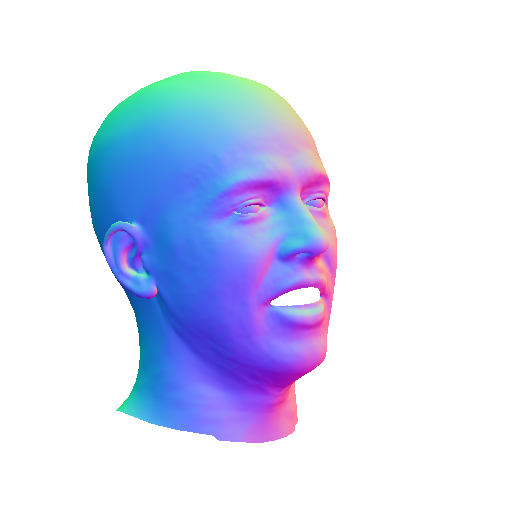} &
        \includegraphics[width=\w]{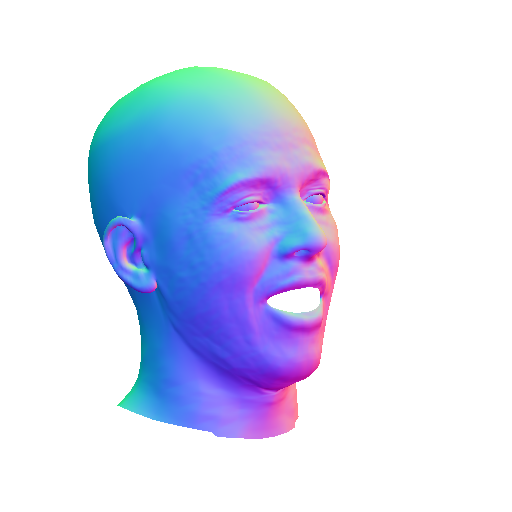} &
        \includegraphics[width=\w]{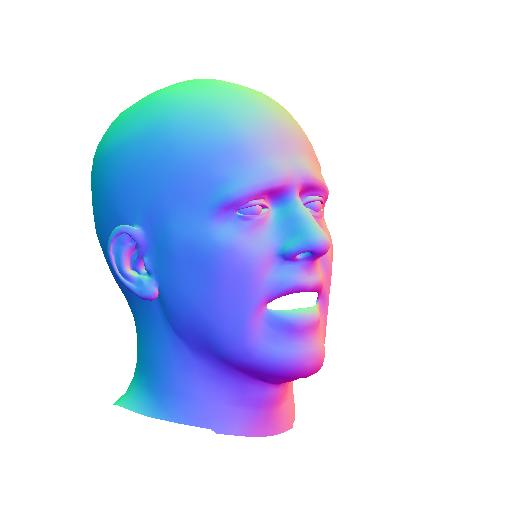} &
        \includegraphics[width=\w]{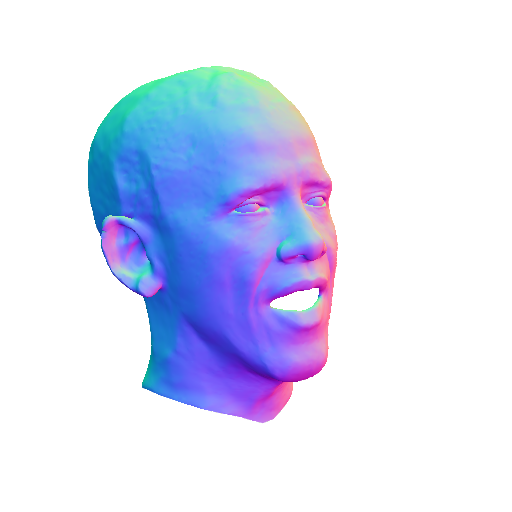}
        & \includegraphics[width=\w]{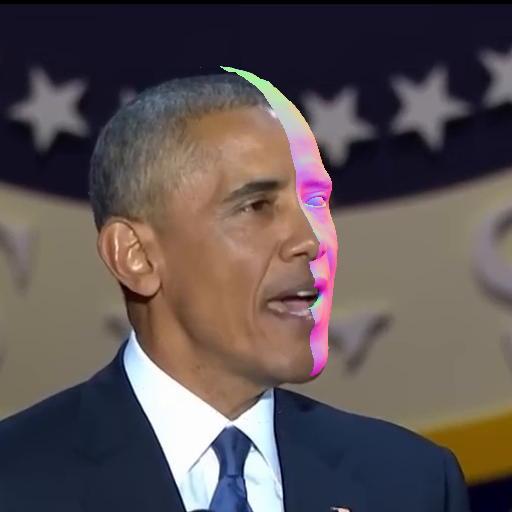}
        \\
        \includegraphics[width=\w]{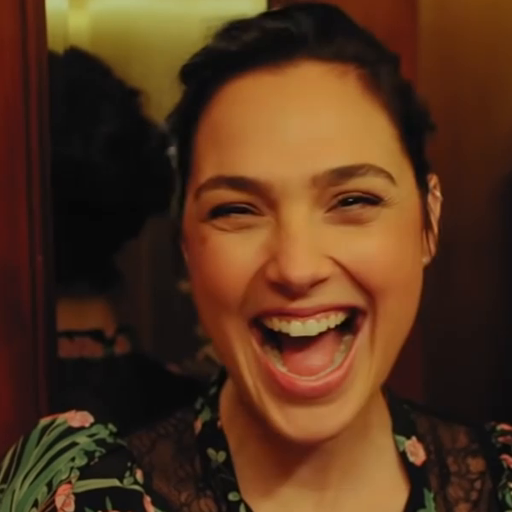} &
        \includegraphics[width=\w]{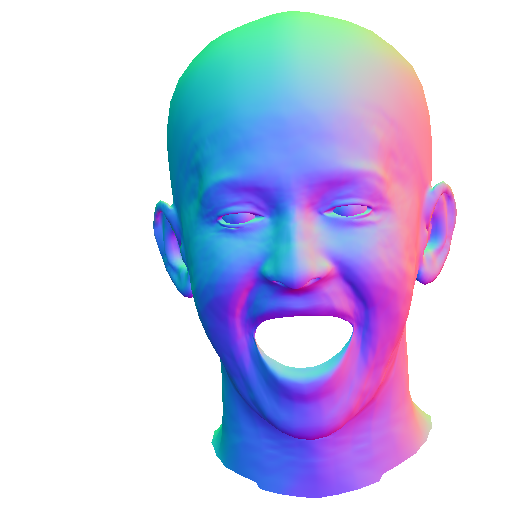} &
        \includegraphics[width=\w]{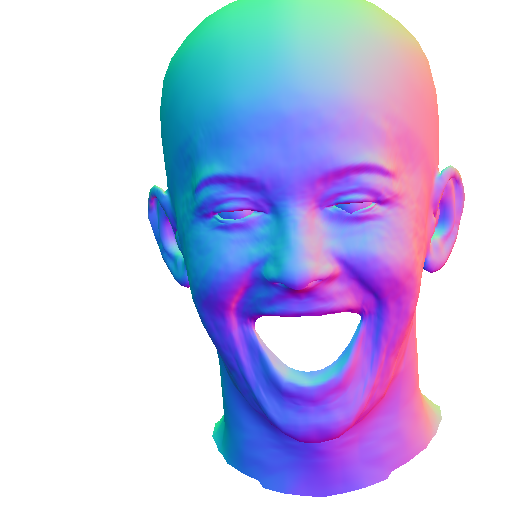} &
        \includegraphics[width=\w]{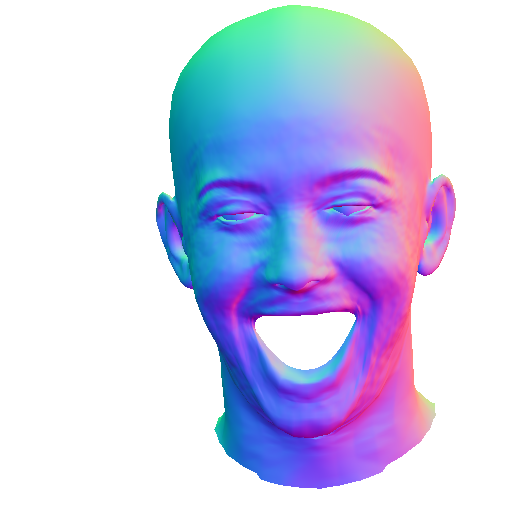} &
        \includegraphics[width=\w]{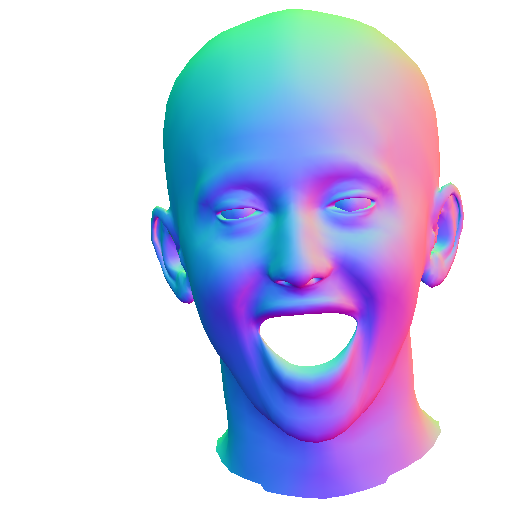} &
        \includegraphics[width=\w]{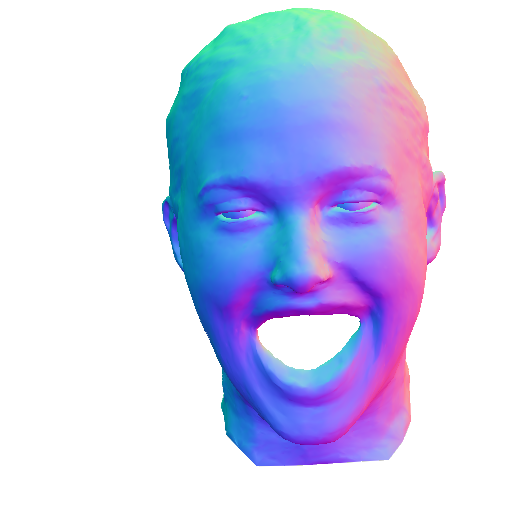}
        & \includegraphics[width=\w]{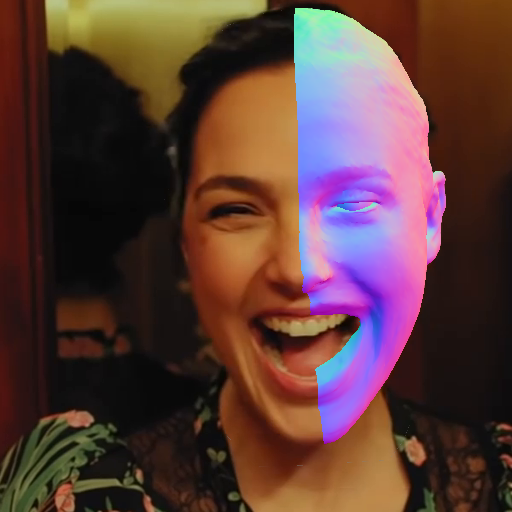}
        \\
        \includegraphics[width=\w]{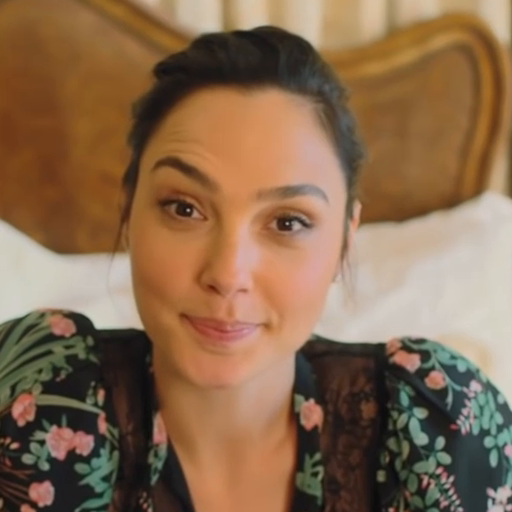} &
        \includegraphics[width=\w]{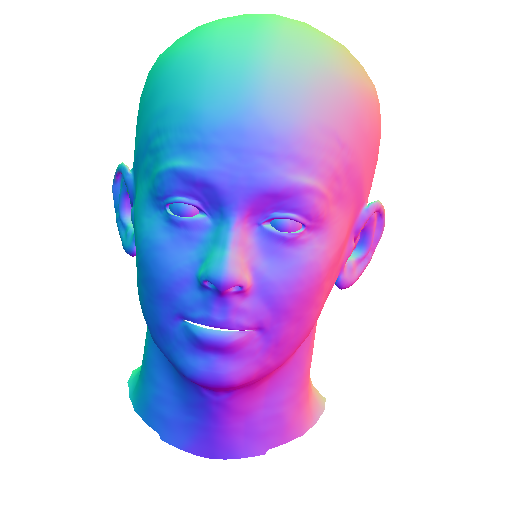} &
        \includegraphics[width=\w]{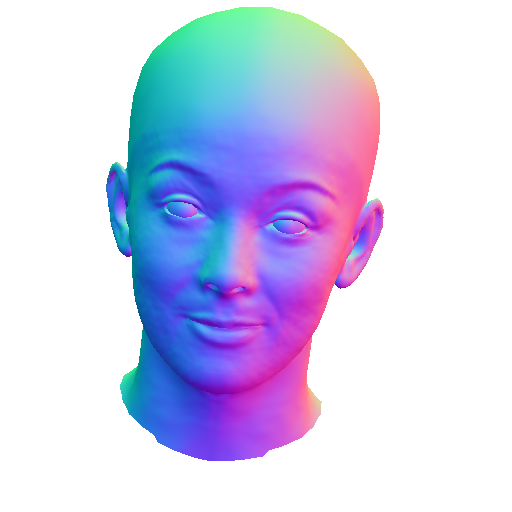} &
        \includegraphics[width=\w]{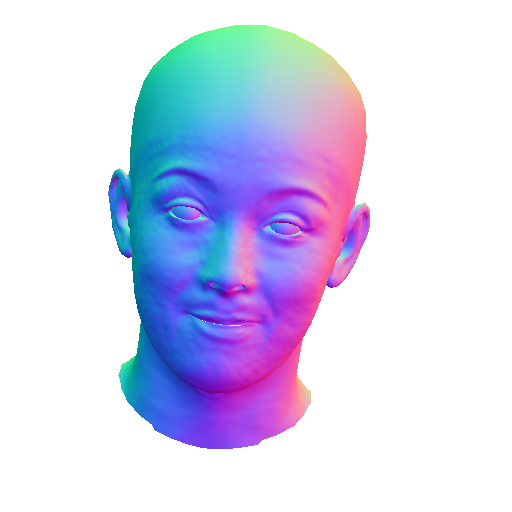} &
        \includegraphics[width=\w]{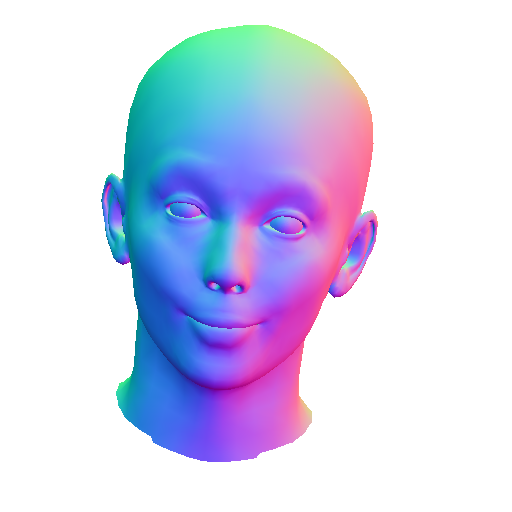} &
        \includegraphics[width=\w]{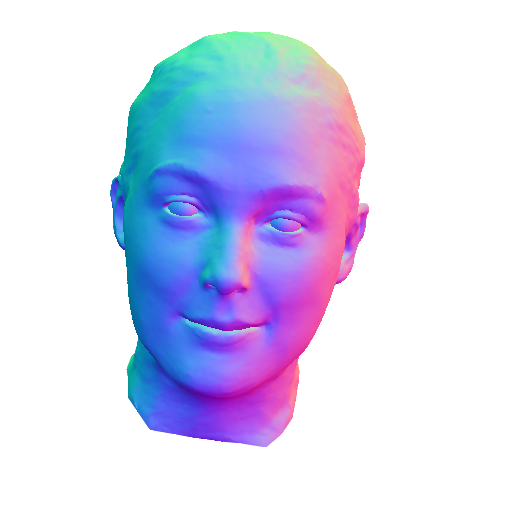}
        & \includegraphics[width=\w]{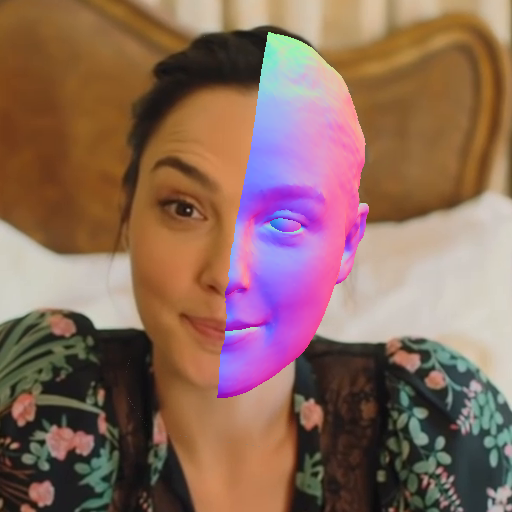}
        \\
        \includegraphics[width=\w]{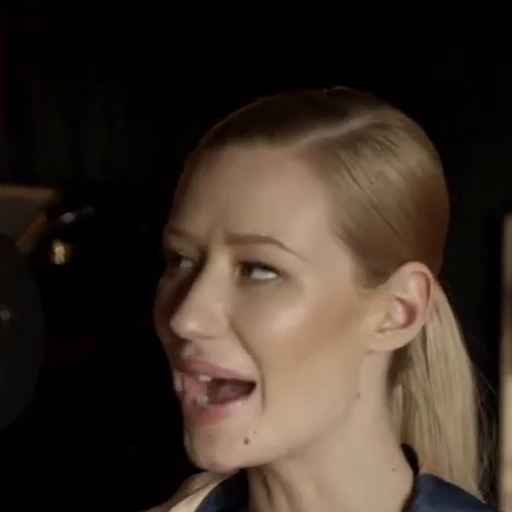} &
        \includegraphics[width=\w]{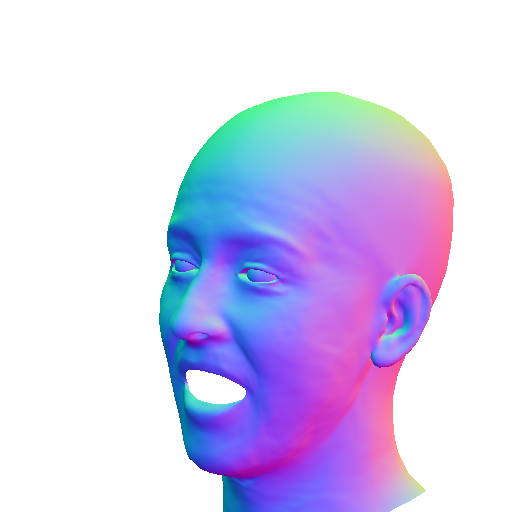} &
        \includegraphics[width=\w]{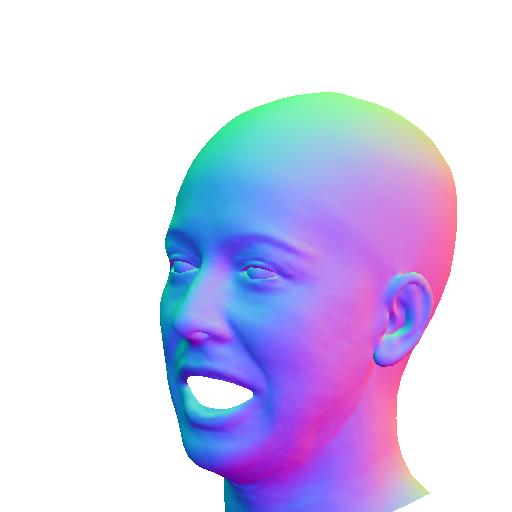} &
        \includegraphics[width=\w]{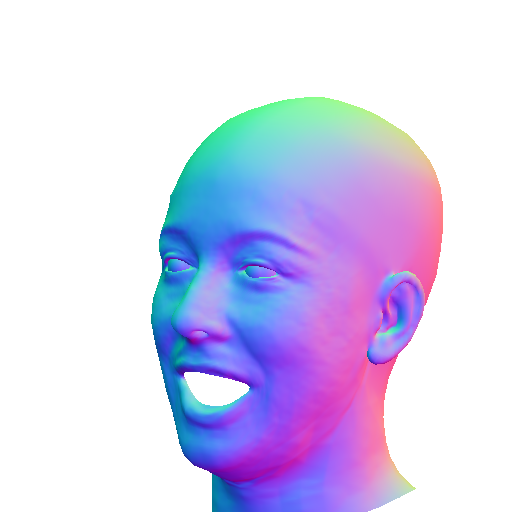} &
        \includegraphics[width=\w]{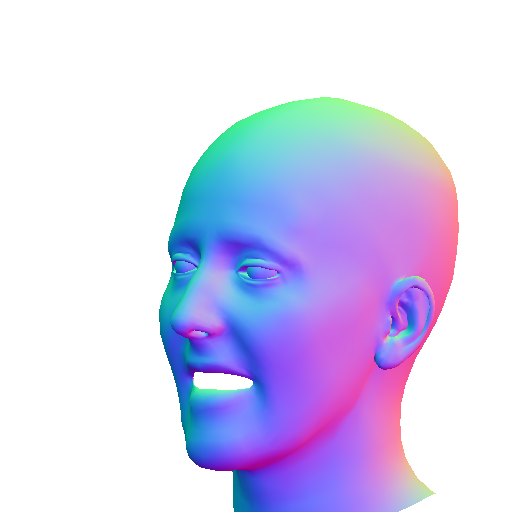} &
        \includegraphics[width=\w]{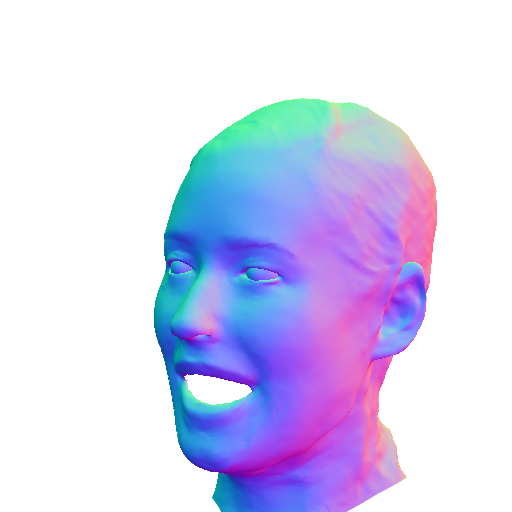}
        & \includegraphics[width=\w]{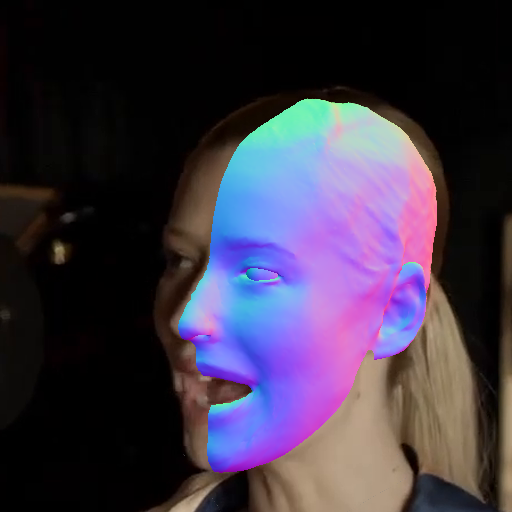}
        \\
        \includegraphics[width=\w]{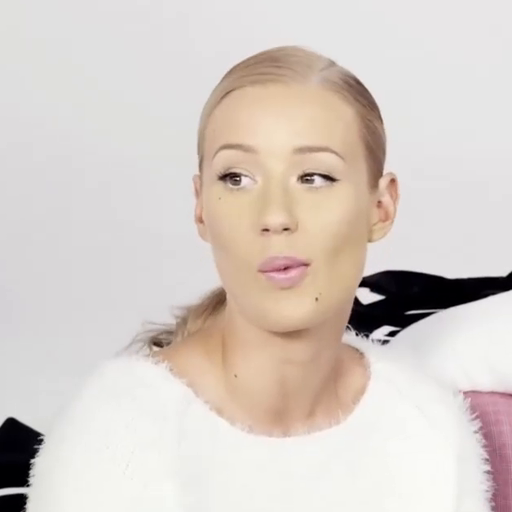} &
        \includegraphics[width=\w]{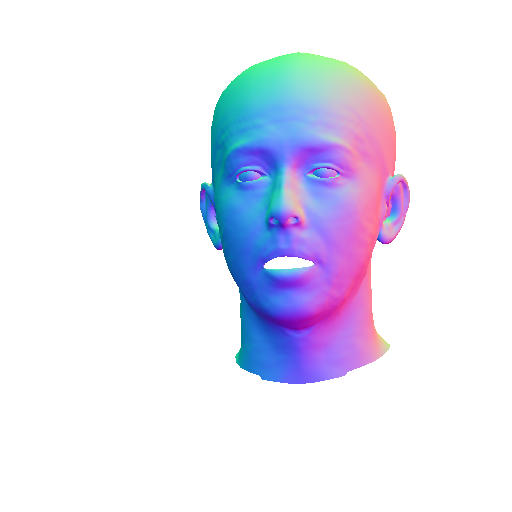} &
        \includegraphics[width=\w]{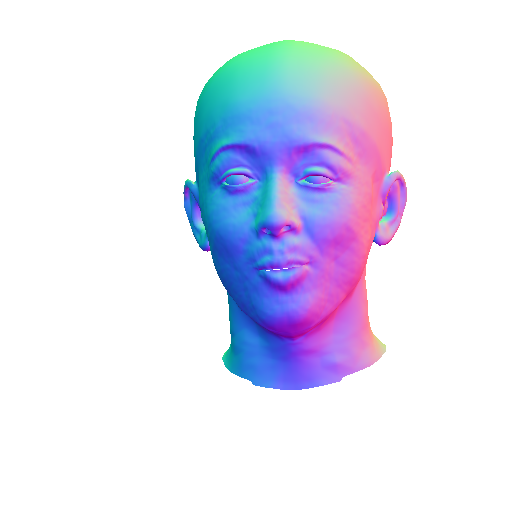} &
        \includegraphics[width=\w]{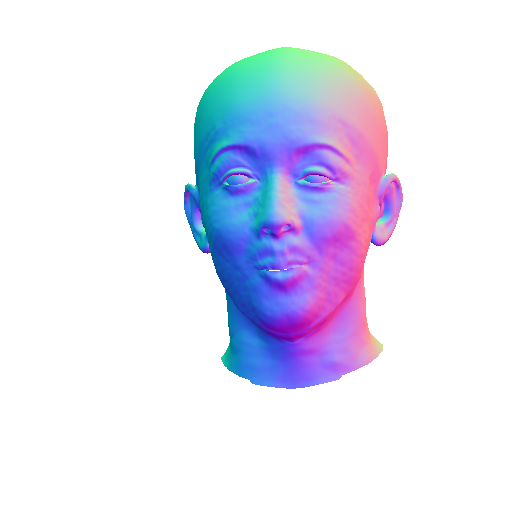} &
        \includegraphics[width=\w]{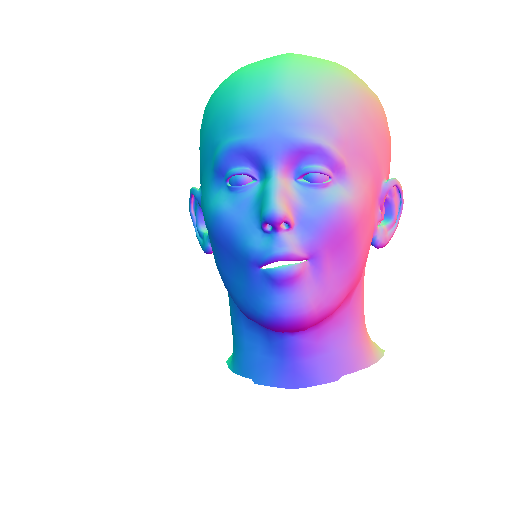} &
        \includegraphics[width=\w]{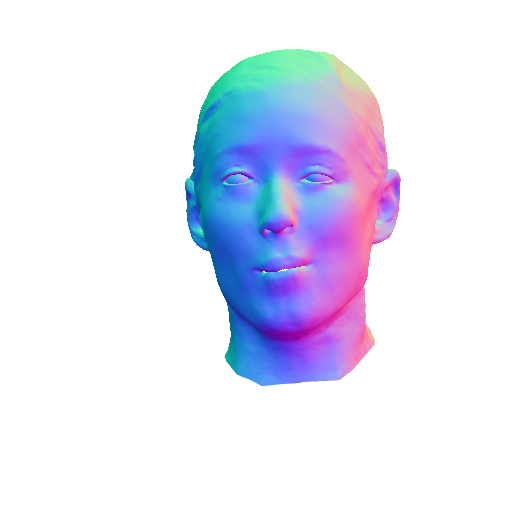}
        & \includegraphics[width=\w]{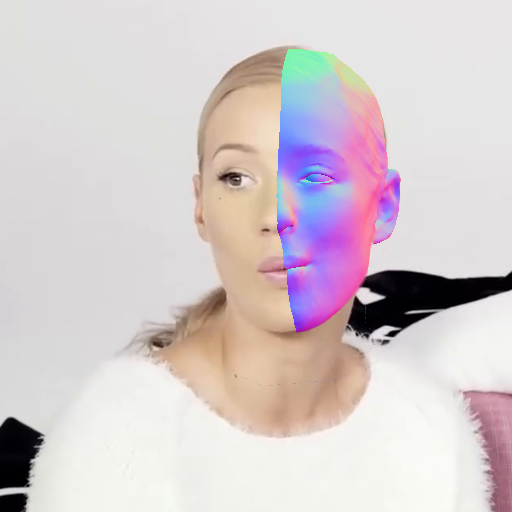}
        \\
        \includegraphics[width=\w]{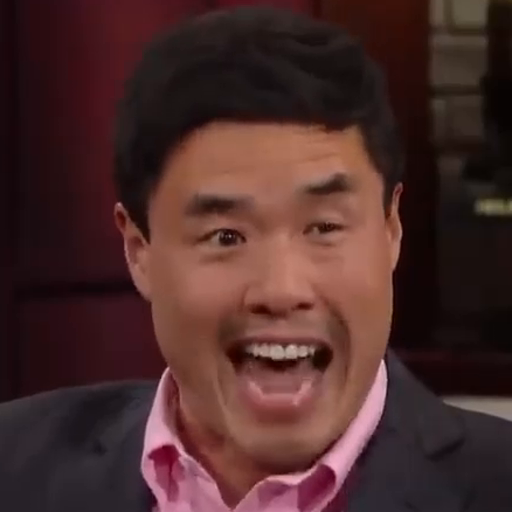} &
        \includegraphics[width=\w]{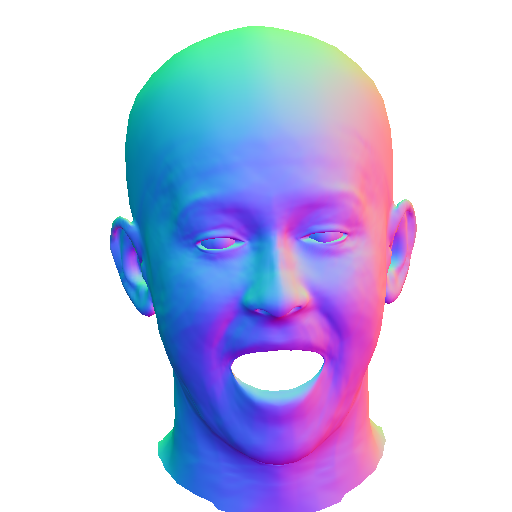} &
        \includegraphics[width=\w]{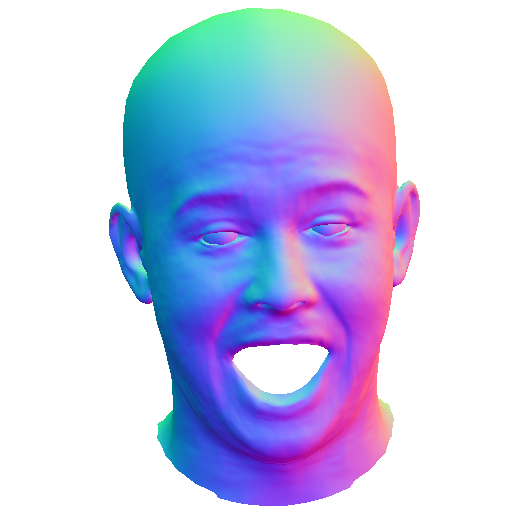} &
        \includegraphics[width=\w]{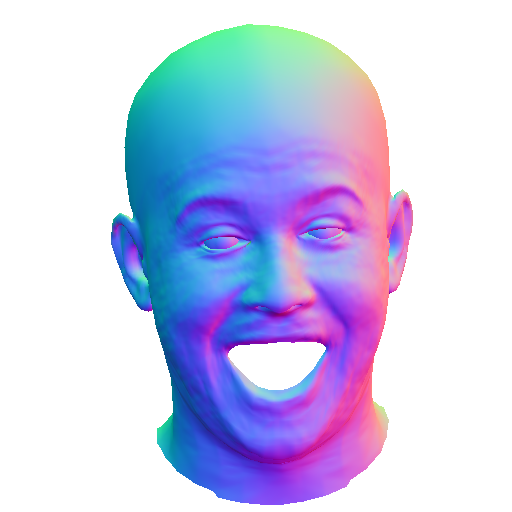} &
        \includegraphics[width=\w]{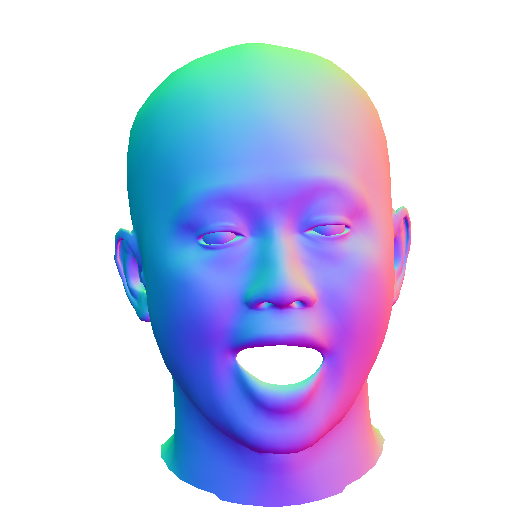} &
        \includegraphics[width=\w]{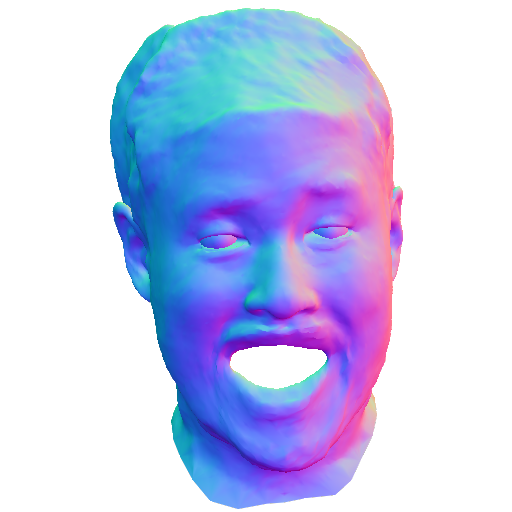}
        & \includegraphics[width=\w]{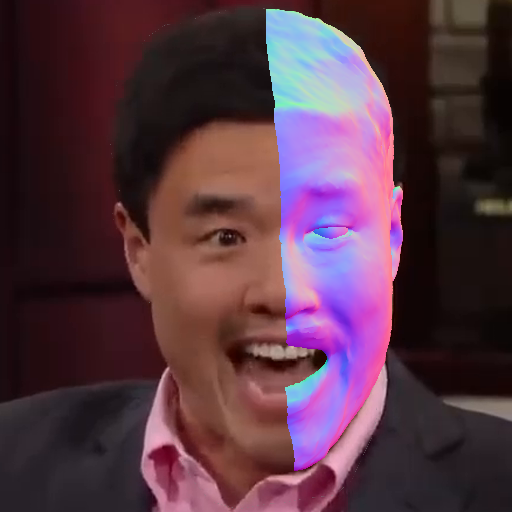}
        \\
        \includegraphics[width=\w]{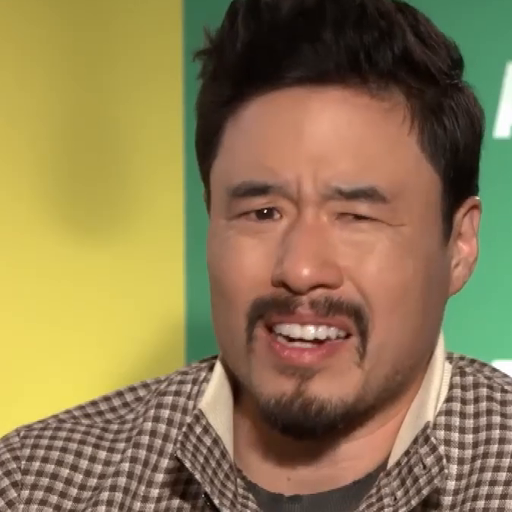} &
        \includegraphics[width=\w]{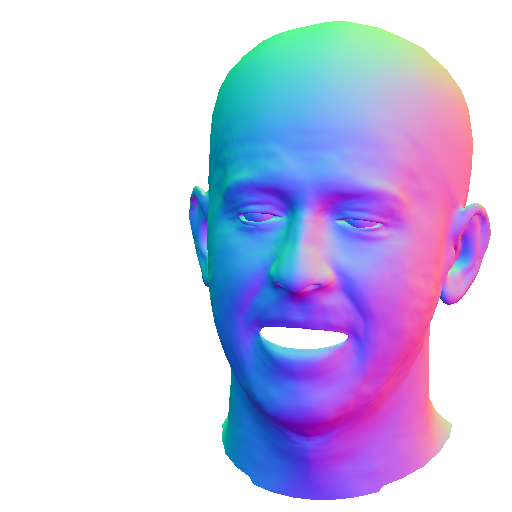} &
        \includegraphics[width=\w]{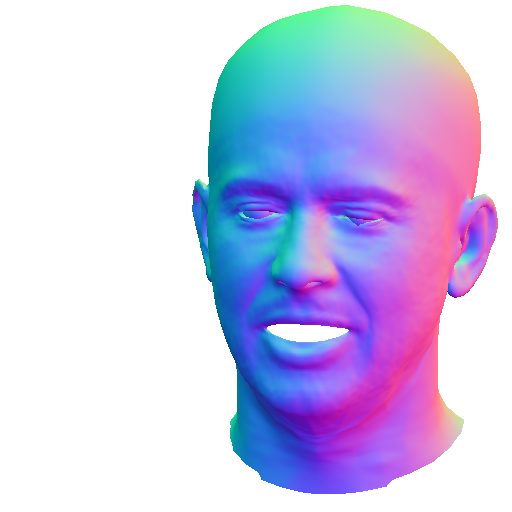} &
        \includegraphics[width=\w]{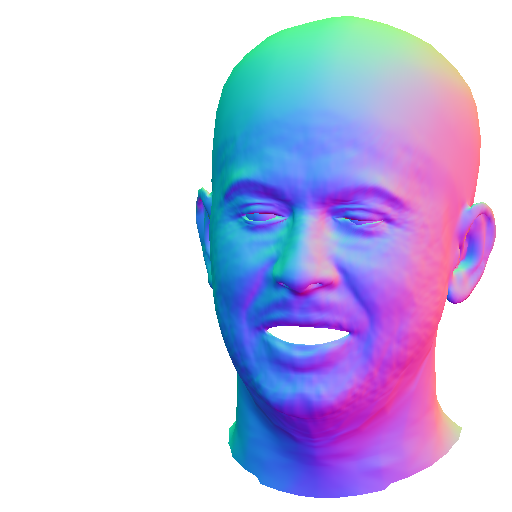} &
        \includegraphics[width=\w]{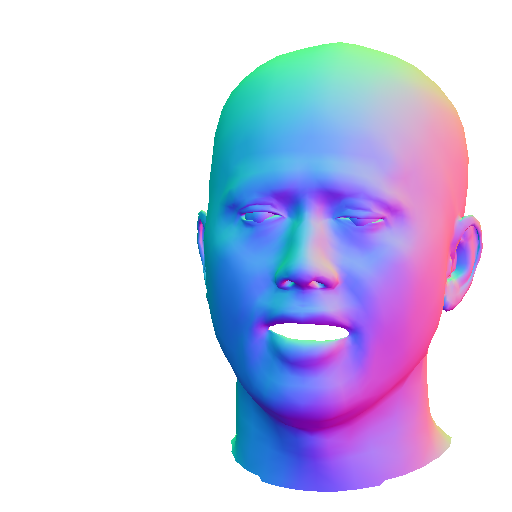} &
        \includegraphics[width=\w]{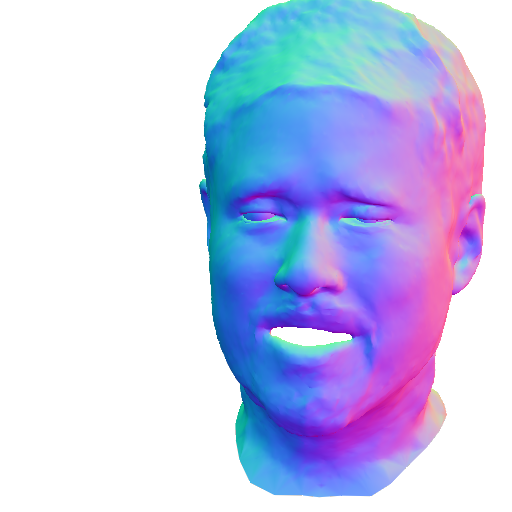}
        & \includegraphics[width=\w]{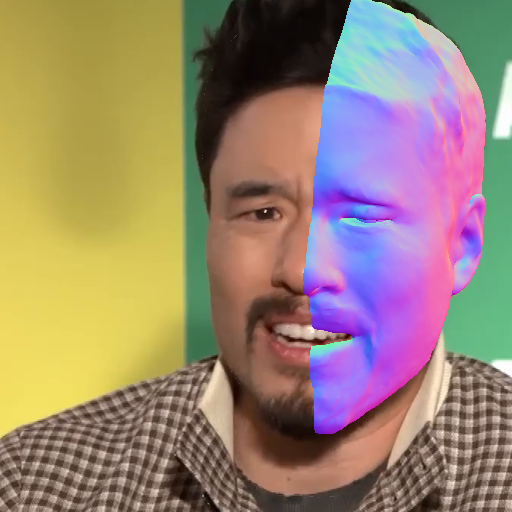}
    \end{tabular}
    }
    \caption{Comparison to multiple state of the art face reconstruction methods. From left to right: input image, DECA \cite{feng2021learning}, EMOCA \cite{danecek2022EMOCA}, EMOCA fine-tuned,  SMIRK \cite{retsinasSMIRK3DFacial2024} and SPARK. See Fig. \ref{fig:qualitative_more} for more examples. More results are also provided in our supplemental videos.}
    \label{fig:qualitative_main}
    \Description{A grid of images showing, from left to right, the ground truth image, the reconstruction using various real-time face tracking methods, our reconstruction, and our result overlaid on top of half of the input face. 10 examples are shown.}
\end{figure*}

In this section we demonstrate our method on 6 datasets of different subjects, chosen for diversity and ease of access to in-the-wild videos.
Each dataset is made up of 6 to 12 in-the-wild videos with durations ranging between 10 seconds and 1 minute. These videos are cut from interviews, talks and dialogue scenes. We provide sources for the videos in our supplemental material. For practicality purposes, we subsample all sequences at a 1:4 ratio for training. We train \avatarStage{} at a resolution of 512 $\times$ 512. The images are further cropped and resized to 224 $\times$ 224 for the transfer learning stage, a standard resolution used by many recent feedforward face reconstruction methods.

We compare our method with state-of-the-art feedforward face capture methods: DECA \cite{feng2021learning}, EMOCA \cite{danecek2022EMOCA} and SMIRK \cite{retsinasSMIRK3DFacial2024}. Additionally, for fair comparison, we evaluate against EMOCA fine-tuned, per identity, on the data used for building our adapted feedforward model.

\subsection{Qualitative Evaluation}

\paragraph{Multi-video avatar reconstruction}
Given \numVideos{} videos, SPARK first reconstructs a head avatar using \avatarStage{}. Fig.~\ref{fig:qualitative_multiflare} shows examples of the reconstructed image, the albedo, shading and geometry for a few subjects. By leveraging multiple monocular sequences of a person, we are able to disentangle the intrinsic face albedo from the shading and recover fine geometric details. Additionally, our improved \textit{deformation} network (Eq.~\ref{eq:deformer}) is able to model precise subtle expressive details.

\paragraph{Personalized real-time face capture}
At test time, SPARK is able to reconstruct 3D geometry in real-time given unseen images of the person. Figure \ref{fig:qualitative_main} shows qualitative results of SPARK compared against existing methods. Our approach is able to reconstruct faces that match the input image better than the previous state-of-the-art. In addition to more precise overall alignment of facial features, we find that our improved \textit{deformation} model can represent fine dynamic details of the face such as wrinkles and cheek movement under large jaw poses. This demonstrates our method's ability to reconstruct details beyond what 3DMMs such as FLAME can represent, even on unseen images of the person, under illuminations and poses that differ from the training set of videos. More results are available in the supplemental videos, where we also demonstrate how our accurate tracking can be used for face editing tasks such as adding facial hair through texture mapping.

\subsection{Quantitative Evaluation}

Existing face capture methods generally do not evaluate the accuracy of the posed geometry. One common scheme is to evaluate the estimated neutral face using 3D ground truth data, with benchmarks such as NoW~\cite{RingNet:CVPR:2019}. Differently, we seek to evaluate the quality of the estimated poses and expressions. However, to our knowledge there is no dataset that comprises multiple monocular in-the-wild videos  of the same subject in different contexts with per-frame 3D ground truths for evaluation. We also argue that even recent dense landmark prediction models remain too sparse to accurately evaluate state of the art dense geometric face capture methods. Additionally, since landmarks are often used to train such methods, the specific landmarks layout chosen for evaluation could yield an unfair advantage to some methods over others (overfitting). Thus, we introduce two new metrics for evaluating the posed geometry independently of albedo and shading.

\begin{figure}
    \renewcommand{\w}{0.18\linewidth}
    \setlength{\tabcolsep}{0.0em} % horizontal padding
    \def\arraystretch{0.8}{ % vertical padding
    \begin{tabular}{ccccc}
        & \shortstack{Masked\\ input} & \shortstack{Predicted\\ mask} & \shortstack{EMOCA\\ render} & \shortstack{SPARK\\ render} \\
        \multirow{2}{0.27\linewidth}[1.8cm]{
            \includegraphics[width=\linewidth]{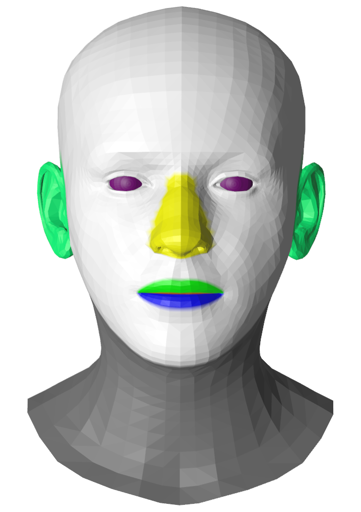}
        } &
        \includegraphics[width=\w]{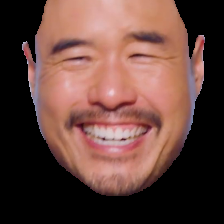} &
        \includegraphics[width=\w]{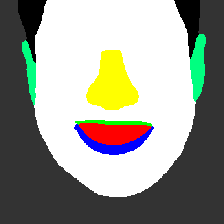} &
        \includegraphics[width=\w]{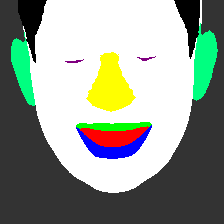} &
        \includegraphics[width=\w]{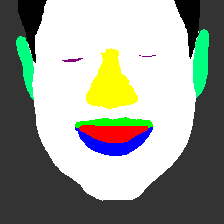} \\
        &
        \includegraphics[width=\w]{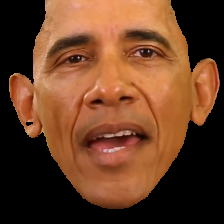} &
        \includegraphics[width=\w]{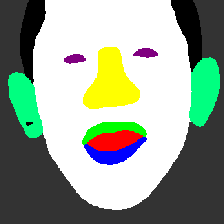} &
        \includegraphics[width=\w]{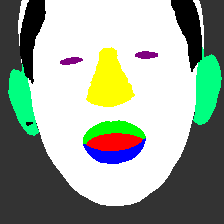} &
        \includegraphics[width=\w]{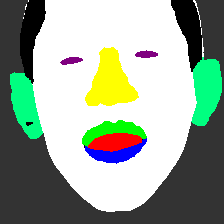}
    \end{tabular}
    }
    \caption{Illustration of our semantic Intersection-over-Union metric. Left: manually annotated semantic masks for FLAME. Right: two examples with ground-truth segmentation, a render of EMOCA's tracked mesh and our tracked mesh.}
    \label{fig:semantic_iou_examples}
    \Description{On the left, the template FLAME mesh is shown with color masks on the ears, nose, eyes, upper lip, lower lip and neck. On the right, two examples are shown with the input face, the predicted pseudo ground-truth segmentation mask, and the segmentation mask rendered by tracking the geometry with EMOCA and SPARK.}
\end{figure}

\begin{figure}
    \renewcommand{\w}{0.2\linewidth}
    \newcommand{\wDouble}{0.4\linewidth}
    \setlength{\tabcolsep}{0.0em} % horizontal padding
    \def\arraystretch{0.5}{ % vertical padding
    \begin{tabular}{ccc}
        \small{$I_{t-k}$} & \small{$I_{t}$} & \small{Warped image $I^{\text{w}}_{t}$} \\
        \includegraphics[width=\w]{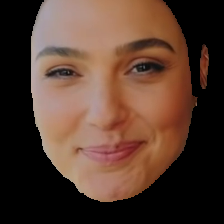} &
        \includegraphics[width=\w]{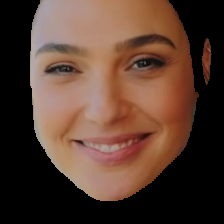} &
        \includegraphics[width=\w]{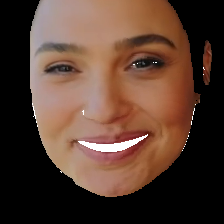} \\
        \multicolumn{2}{c}{\includegraphics[width=\wDouble]{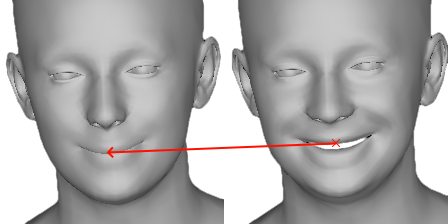}}
    \end{tabular}
    }
    \caption{Illustration of our image warping metric. The background and hair are masked out in both frames. We use the tracked geometry to backtrack each rasterized pixel of image $I_t$ to a pixel in image $I_{t-k}$. The white area is the occlusion mask, where we do not compare pixels.}
  \label{fig:warp_illustration}
  \Description{A face is shown for two consecutive frames of a video. Below are the tracked meshes for these frames, with a red arrow pointing from one point of the upper lip in the geometry of the second frame, to the corresponding surface point in the geometry of the first frame. On the right, the result of the per-pixel warping is shown, resembling the second frame but having been reconstructed using only pixels from the first frame.}
\end{figure}

\paragraph{Semantic IoU metric}
We use state-of-the-art semantic segmentation method BiSeNet~\cite{yuBiSeNetBilateralSegmentation2018} to predict semantic masks for the following head regions: skin, hair, nose, ears, eyes, upper lip, lower lip and mouth interior. We mask out the hair region, as hair is not modeled by current feedforward face capture methods. We manually annotate the FLAME template mesh with corresponding semantic masks, as shown in Fig.~\ref{fig:semantic_iou_examples}, allowing us to render a semantic segmentation from the posed geometry. We compute the mean intersection over union (IoU) over all the semantic classes of a given frame:
$$IoU_s = \frac{1}{C} \sum_{i=1}^{C}\frac{|M_i \cap \hat{M}_i|}{|M_i \cup \hat{M}_i|}$$
where $C$ is the number of classes considered, $\hat{M}_i$ are the respective pseudo ground-truth semantic masks and $M_i$ are the rasterized masks. In Fig.~\ref{fig:semantic_iou_examples} we show examples of the predicted and rasterized masks. The semantic IoU error measures the alignment of various face regions independently of external factors such as scene illumination, motion blur and high frequency variation on the skin. However, it is sensitive to the accuracy of the semantic segmentation.

\paragraph{Geometry-based image warping metric}
We introduce a second metric directly based on the input images to measure the accuracy of tracked geometry. The video sequence is sampled at fixed intervals $k$. For each pair of consecutively sampled frames ($I_{t-k}$, $I_{t}$), we use the tracked geometry to reconstruct image $I_{t}$ from pixels of $I_{t-k}$. More precisely, we use rasterized barycentric coordinates to map each pixel in $I_{t}$ to a surface point on the tracked geometry. We then project the corresponding point on the geometry of $I_{t-k}$ to screen space and retrieve the pixel value, forming a new warped image $I^t_w$. The image warping error is defined as follows:
$$W_{\text{PSNR}}(t) = \text{PSNR}(I_{t} \cdot o_{t-k,t}, I_t^{\text{w}} \cdot o_{t-k,t})$$
where $o_{t-k,t}$ is a binary occlusion mask for surface points that were not visible in $I_{t-k}$.
Note that we use semantic segmentation masks to remove the background and hair from both images. In Fig.~\ref{fig:warp_illustration}, we illustrate the warping on a pair of images. While this metric can be computed for any arbitrary pair of images, we find that it is more reliable with a short frame interval, reducing potential issues such as aliasing, incorrect shading and large occlusion masks. We empirically choose an interval of 170 ms (5 frames at 30 frames per second).

We perform k-fold cross-validation on every subject, with k chosen such that we always leave out 2 or 3 sequences for evaluation. All other sequences are used for training. We report the semantic IoU and image warping metric as introduced in the previous section, as well as landmarks reprojection errors averaged across subjects in Table \ref{tab:results_main}. We also report per-subject results in our supplemental material. We outperform the competition numerically with similar inference times, which confirms our superior qualitative results.

\begin{table}
    \setlength{\tabcolsep}{0.3em} % horizontal padding
    \def\arraystretch{1.0}{ % vertical padding
    \begin{tabular}{cccc}
        \toprule
        Method & \small{Warp PSNR $\uparrow$} & \small{Semantic IoU $\uparrow$} & \small{Landmarks L1 $\downarrow$} \\
        \midrule
        DECA & 29.501 & 0.616 & 0.047\\
        SMIRK & 30.062 & 0.649 & 0.035\\
        EMOCA & 29.979 & 0.644 & 0.036\\
        \small{EMOCA (fine-tuned)} & 30.275	& 0.673 & \textbf{0.027}\\
        \midrule
        Ours & \textbf{30.647} & \textbf{0.702} & \textbf{0.027}\\
        \bottomrule
    \end{tabular}
    }
    \caption{Quantitative results assessing the accuracy of the captured geometry on unseen data.}
    \label{tab:results_main}
\end{table}

\begin{table}
    \setlength{\tabcolsep}{0.1em} % horizontal padding
    \def\arraystretch{1.0}{ % vertical padding
    %\scalebox{0.8}{
    \scalebox{1.0}{
    \begin{tabular}{cccc}
        \toprule
        Ablation & \small{Warp PSNR $\uparrow$} & \small{Semantic IoU $\uparrow$} & \small{Landmarks L1 $\downarrow$} \\
        \midrule
        \footnotesize{w/o transfer learning} & 30.073 & 0.621 & 0.042 \\
        \footnotesize{w/o personalized appearance} & 30.449 & 0.685 & 0.028 \\
        Train MLP only & 30.559	& 0.699 & 0.028 \\
        Train all & 30.354 & 0.691 & 0.031 \\
        \midrule
        Ours & \textbf{30.647} & \textbf{0.702} & \textbf{0.027} \\
        \bottomrule
    \end{tabular}
    }
    }
    \caption{Quantitative results of our ablation study, averaged over 6 subjects with cross-validation.}
    \label{tab:results_ablation}
\end{table}
\begin{table}
    \setlength{\tabcolsep}{0.5em} % horizontal padding
    \def\arraystretch{1.0}{ % vertical padding
    \begin{tabular}{cccc}
        \toprule
        Sequences & Warp PSNR $\uparrow$ & Semantic IoU $\uparrow$ & Landmarks L1 $\downarrow$\\
        \midrule
        1 & 31.574 & 0.665 & 0.037 \\
        2 & 31.659 & 0.681 & 0.036 \\
        4 & 31.911 & 0.695 & 0.033 \\
        8 & \textbf{32.028} & \textbf{0.706} & \textbf{0.030} \\
        \bottomrule
    \end{tabular}
    }
    \caption{Ablation study on the numbers of sequences used for training. Results are averaged for 3 subjects.}
    \label{tab:results_ablation_sequences}
\end{table}

\subsection{Ablation Study}

\begin{figure}[t]
    \setlength{\tabcolsep}{0.2em} % horizontal padding
    \def\arraystretch{0.0}{ % vertical padding
    \renewcommand{\w}{0.2\linewidth}
    \begin{tabular}{cccc}
        $N=1$ & $N=2$ & $N=4$ & $N=8$\\
        \includegraphics[width=\w]{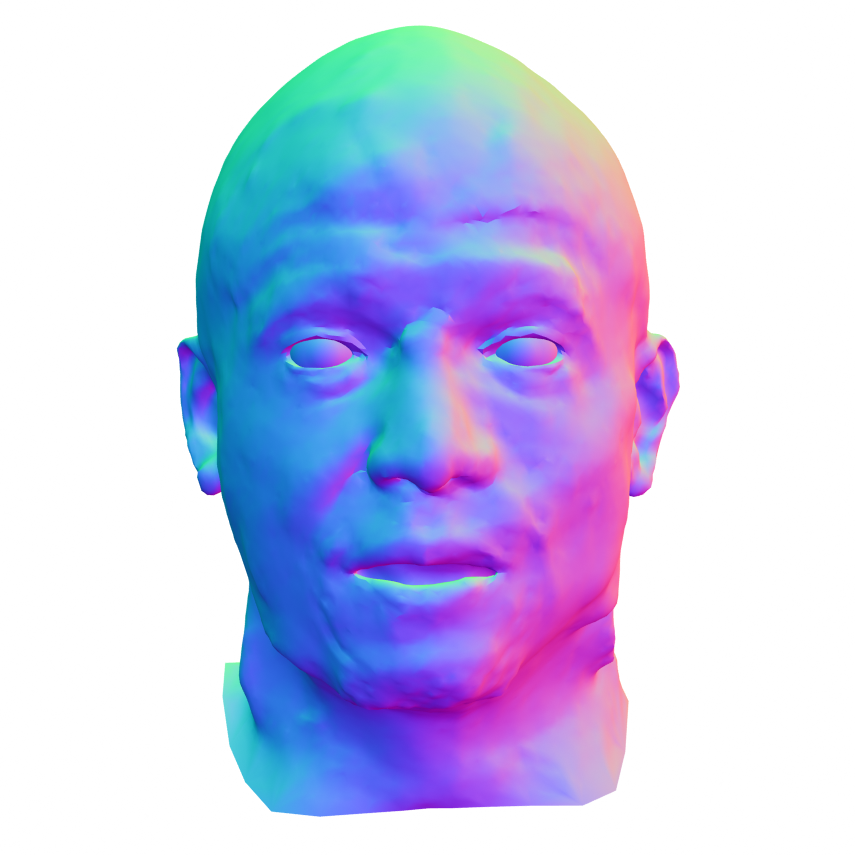} &
        \includegraphics[width=\w]{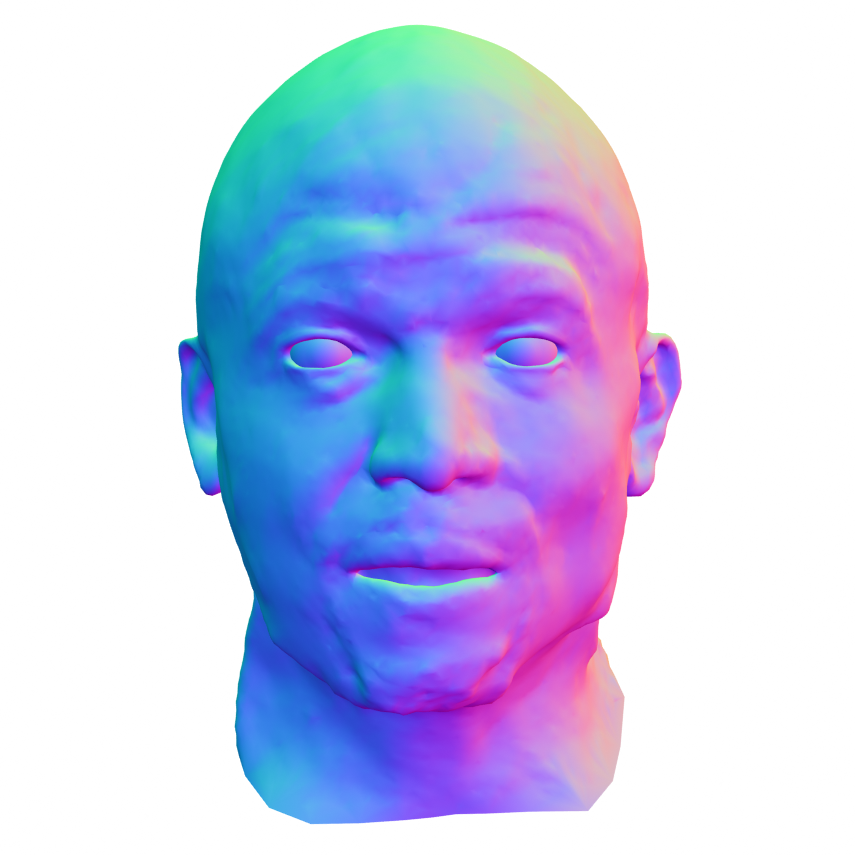} &
        \includegraphics[width=\w]{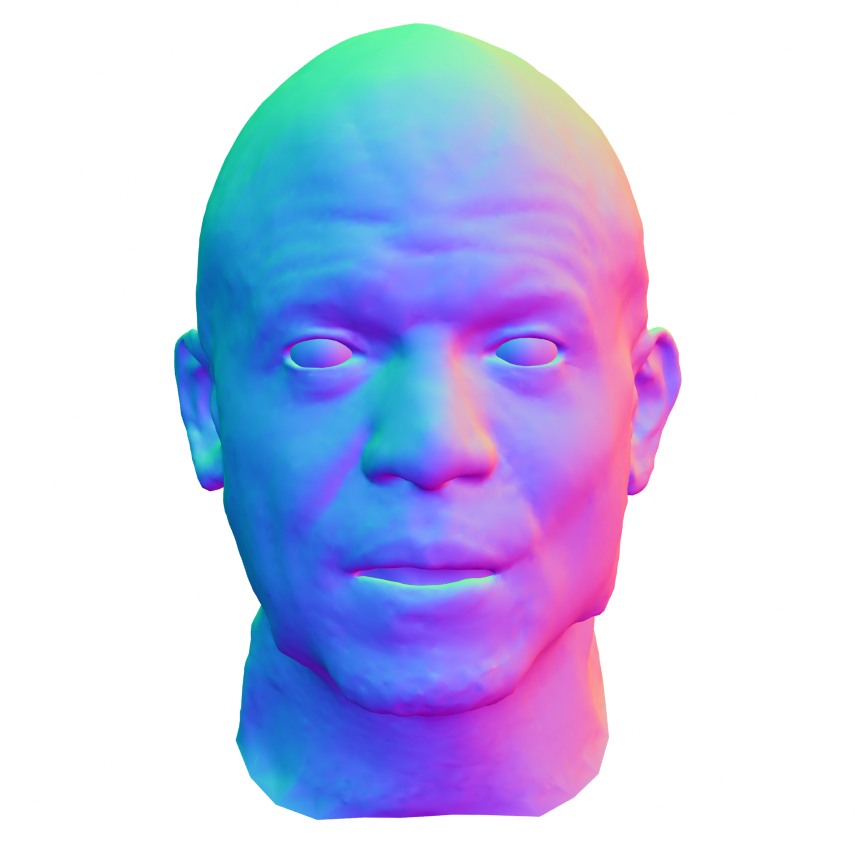} &
        \includegraphics[width=\w]{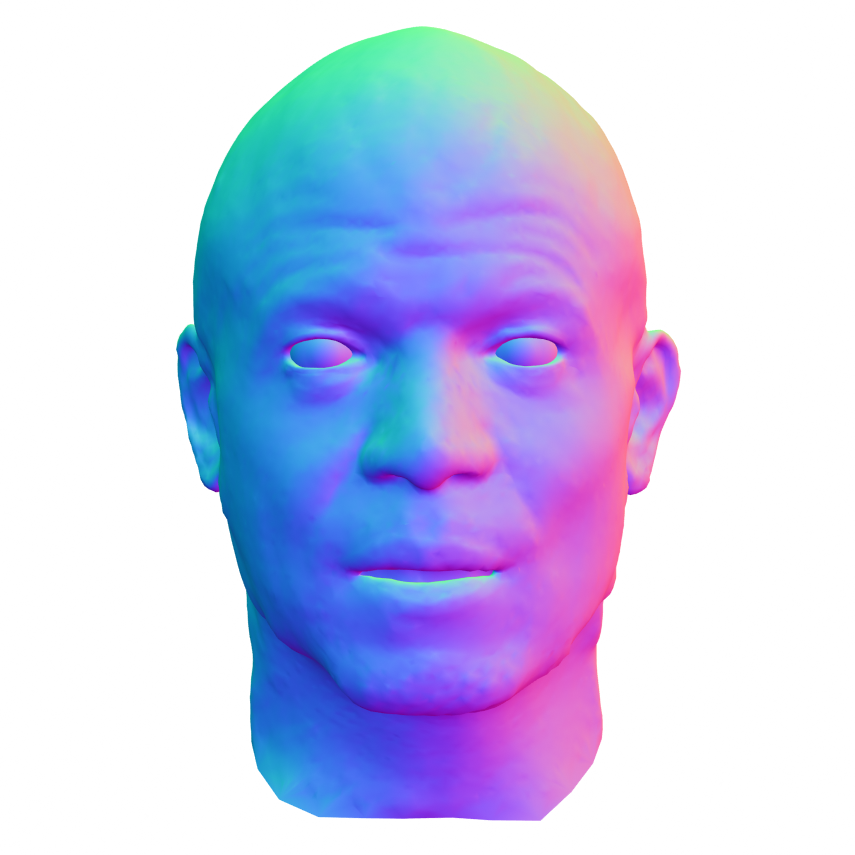} \\
        \includegraphics[width=\w]{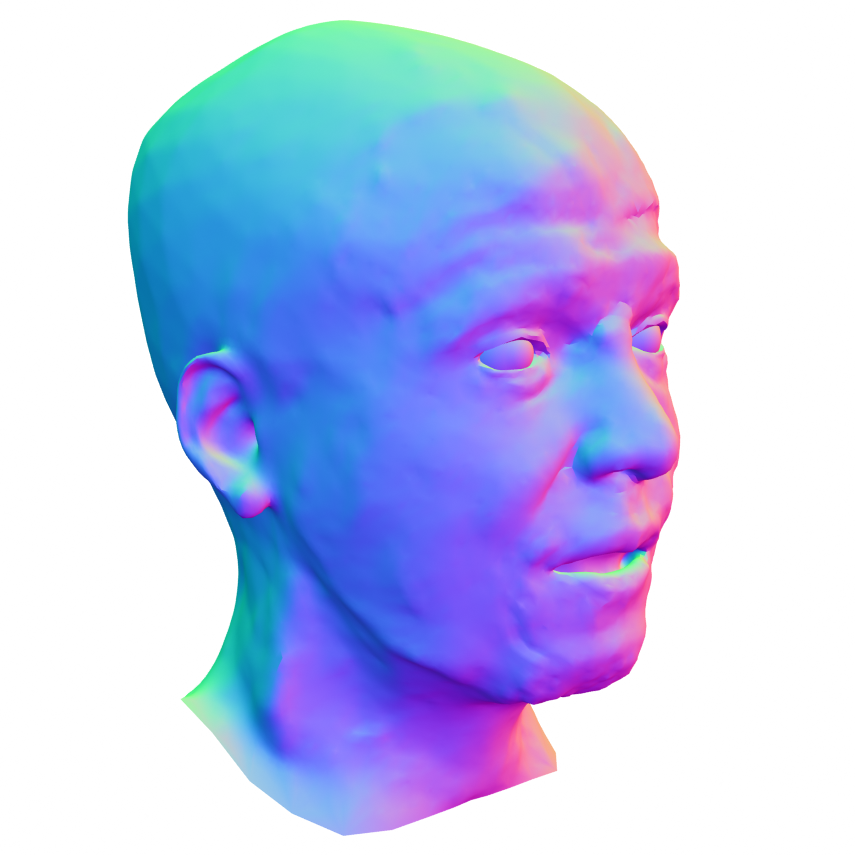} &
        \includegraphics[width=\w]{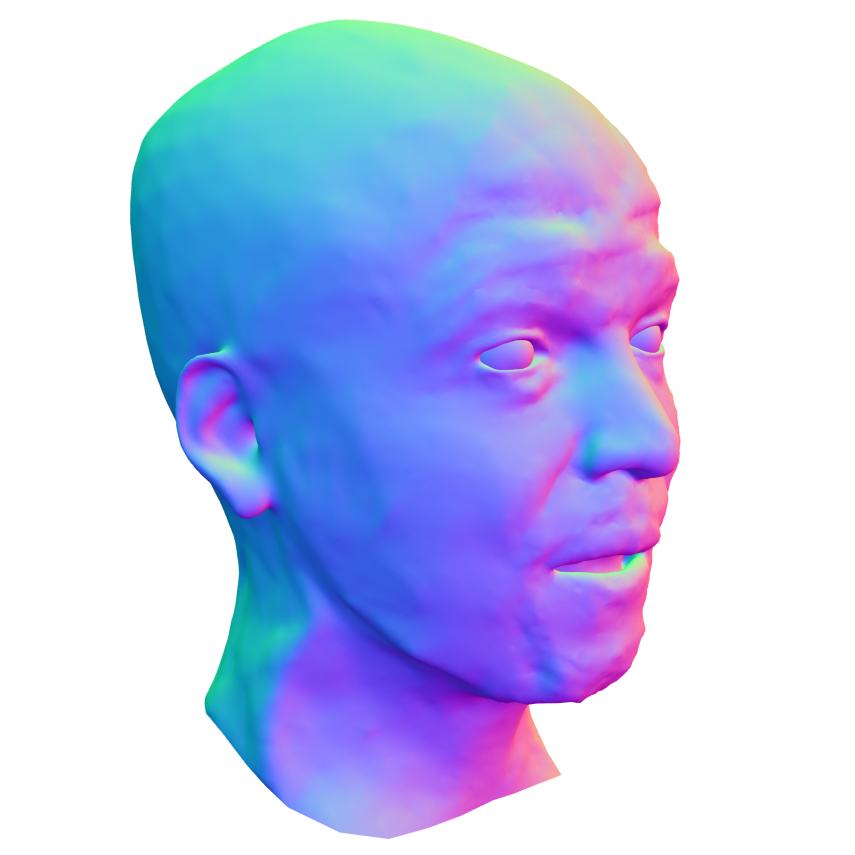} &
        \includegraphics[width=\w]{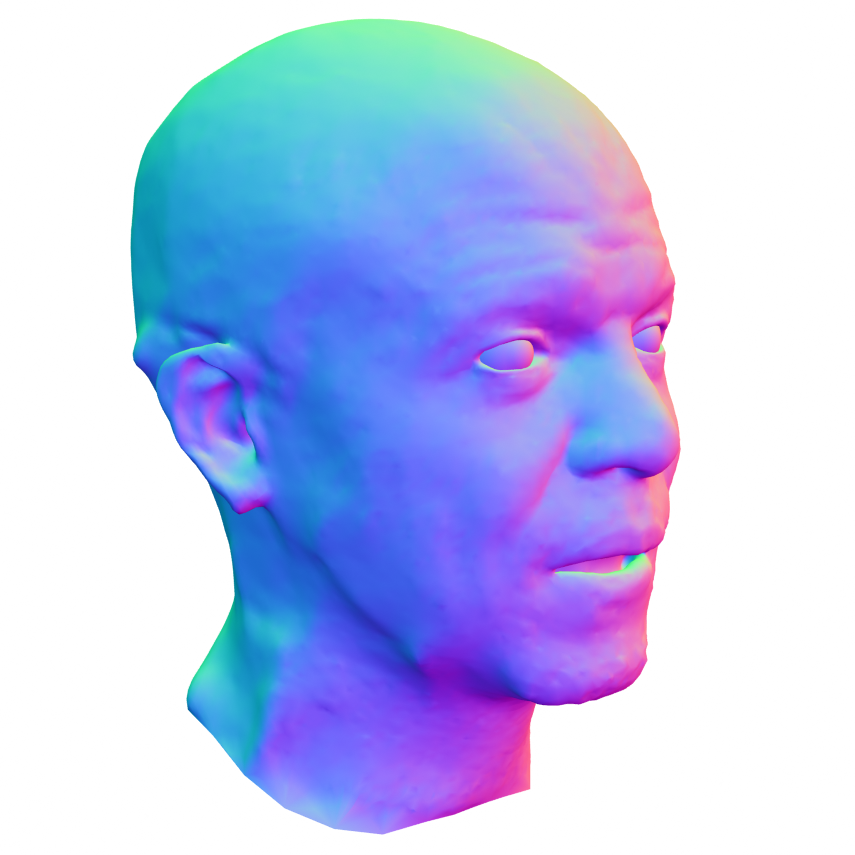} &
        \includegraphics[width=\w]{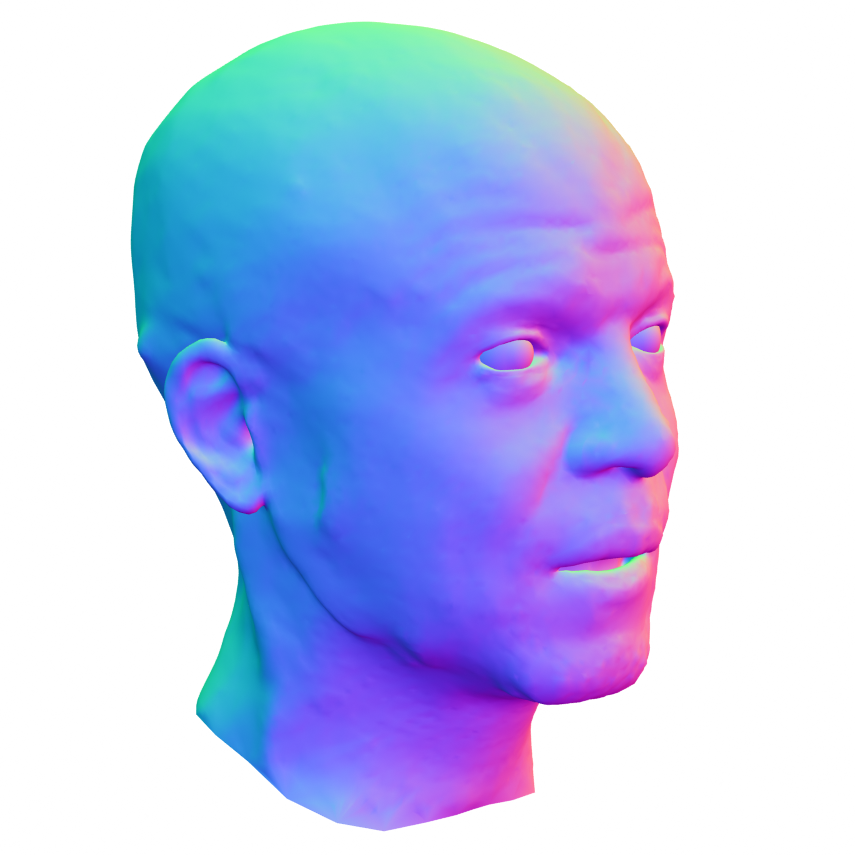} \\
    \end{tabular}
    }
    \caption{Examples of reconstructed canonical geometry from \avatarStage{} for an increasing number of training sequences.}
    \label{fig:ablation_multiflare_sequences}
    \Description{An example of canonical geometry is shown in front and side views, reconstructed using 1, 2, 4 and 8 video sequences of the person, from left to right. The quality greatly improves as the number of videos increases, especially from 1 to 4.}
\end{figure}

\paragraph{Number of sequences}
To our knowledge, we present the first method that uses multiple in-the-wild monocular videos simultaneously for avatar reconstruction. To justify this choice, we ablate the number of sequences we use, both for training \avatarStage{} and for the transfer learning. For 3 subjects for which we have at least 10 sequences, we leave 2 sequences out for evaluation and use 1, 2, 4 or 8 sequences for training. In Fig.~\ref{fig:ablation_multiflare_sequences} we show the reconstructed neutral mesh obtained from \avatarStage{}. In Table \ref{tab:results_ablation_sequences} we report the averaged IoU, image warping PSNR and landmarks reprojection errors over 3 subjects. Using more sequences increases our accuracy on all metrics, although we observe diminishing returns beyond 4. We attribute this improved performance to the more accurate canonical mesh, deformation model and material properties, and to the more diverse training set for transfer learning.

\begin{figure}[t]
    \setlength{\tabcolsep}{0.2em} % horizontal padding
    \def\arraystretch{0.0}{ % vertical padding
    \renewcommand{\w}{0.16\linewidth}
    \begin{tabular}{ccccc}
        Input & $L=0$ & $L=5$ & $L=10$ & $L=20$\\
        \includegraphics[width=\w]{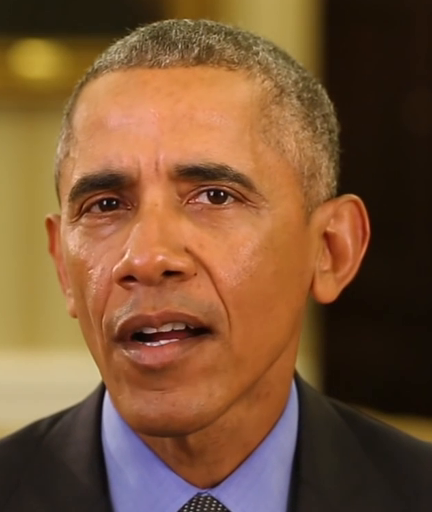} &
        \includegraphics[width=\w]{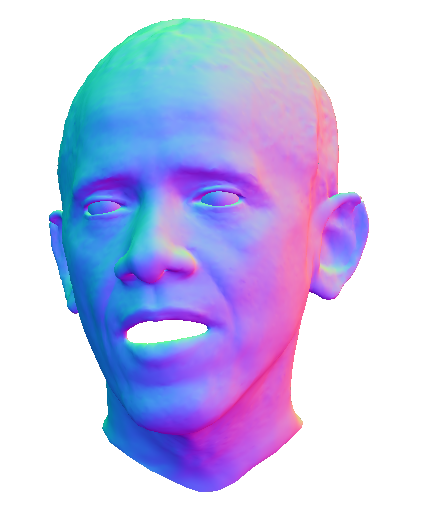} &
        \includegraphics[width=\w]{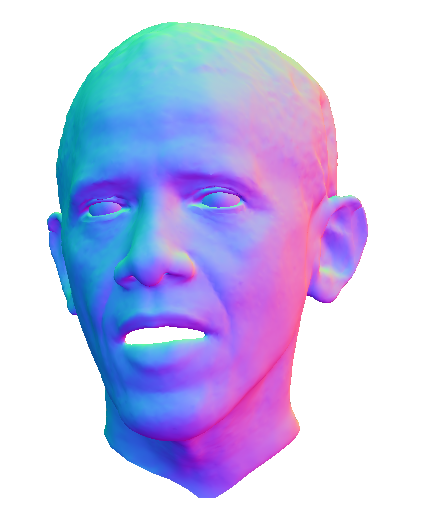} &
        \includegraphics[width=\w]{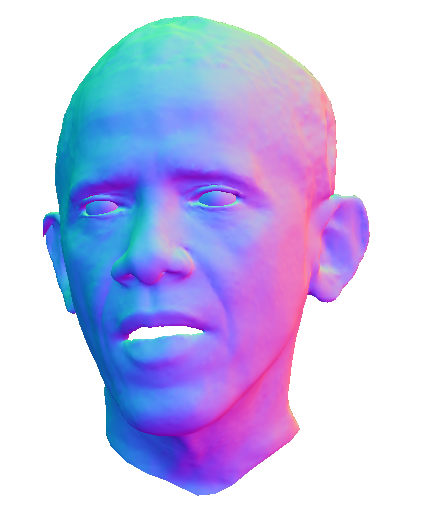} &
        \includegraphics[width=\w]{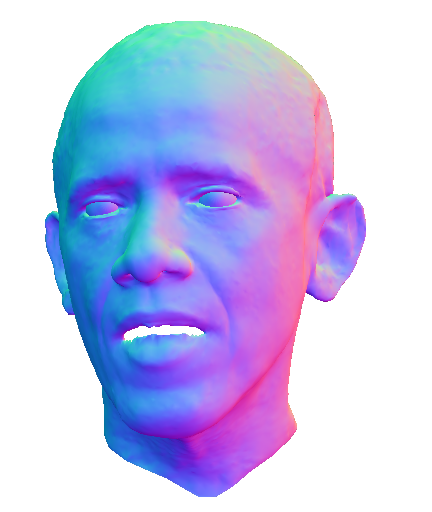} \\
        \includegraphics[width=\w]{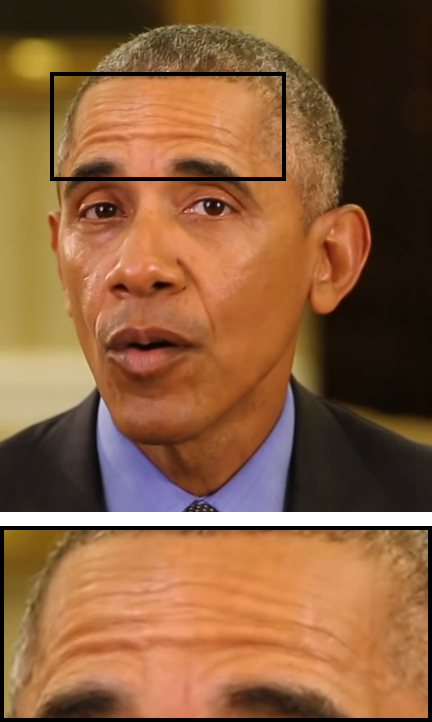} &
        \includegraphics[width=\w]{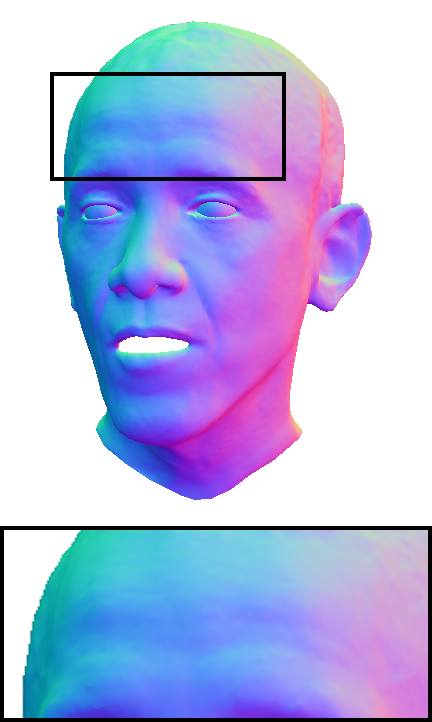} &
        \includegraphics[width=\w]{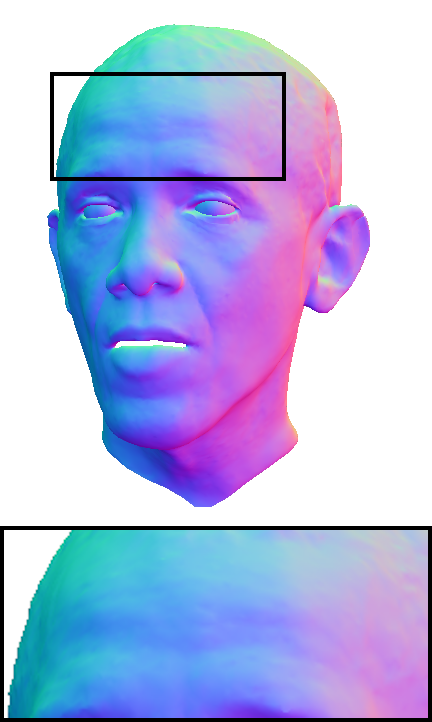} &
        \includegraphics[width=\w]{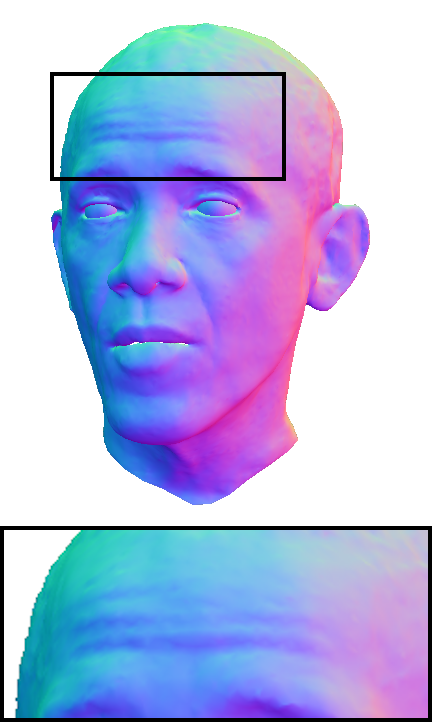} &
        \includegraphics[width=\w]{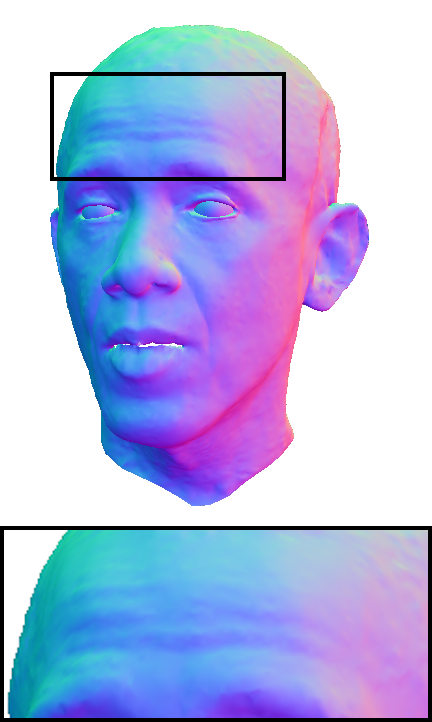} \\
    \end{tabular}
    } 
    \caption{Examples of reconstructed geometry using various levels of positional encoding in the deformer network (Eq \ref{eq:deformer}) of \avatarStage{}.}
    \label{fig:ablation_multiflare_posencoding}
    \Description{The figure shows reconstructed meshes for two poses, after training with L=0, L=5, L=10 or L=20 additional frequencies in the deformation network's positional encoding. Substantial improvements in quality can be observed by increasing L, with diminishing returns after L=10.}
\end{figure}

\paragraph{Positional encoding}
We ablate the number of frequencies used by the positional encoding of the deformer $\mathcal{D}$ (Eq.~$\ref{eq:deformer}$). In Fig.~\ref{fig:ablation_multiflare_posencoding}, we show the varying level of detail that we are able to reconstruct. Without positional encoding, the deformation network is unable to reconstruct the dynamic fine details of the geometry: these details are either lost or baked into the static canonical mesh. With additional frequencies, the fine details are much more defined and dynamic; however, too many high frequencies can lead to a noisy reconstruction.

\paragraph{Transfer learning}: In Table ~\ref{tab:results_ablation}, we report quantitative results for different transfer learning options: training the last backbone layers and the MLPs (our proposed method), training the full backbones and MLPs, and training only the MLP. We find that freezing most of the backbone yields the best results: we keep the pre-trained backbone's ability to generalize to diverse inputs while still allowing the highest-level features to adapt. We also evaluate the impact of using the appearance model estimated in \avatarStage{} instead of the generic model from EMOCA. We show that our more precise appearance learned on the training sequences improves the performance of our final model.

\section{Limitations and Future Work}

The personalized 3D geometry and appearance reconstruction stage of our method inherits several limitations that are specific to mesh-based avatar reconstruction methods. Occlusions and face accessories such as glasses cannot be represented accurately by our method. Furthermore, while we qualitatively show that minor facial hair can be handled, our canonical geometry and deformation model do not differentiate hair from skin, hence large beards and long hair cannot be accurately reconstructed. Similarly to FLARE ~\cite{bharadwaj2023flare}, the simplified integration of the rendering equation, namely the split-sum approximation and pre-filtering of the environment map, results in difficulties for modeling sharp specular highlights and harsh shadows. In turn, this limits the accuracy of our personalized appearance model, thus harming the performance of our transfer-learning in such conditions.

The main limitations of the pre-trained face reconstruction encoder also apply to our method. While we do address some of the limitations that stem from the use of a parametric 3DMM, we are still subject to misalignment on extreme poses and strongly occluded faces.

Future work could explore further refinement of the personalization scheme and the modelling of per-sequence appearance variations to enhance the robustness and applicability of our method across a broader range of scenarios. Our work could also be applied to future state of the art face reconstruction methods, providing an even better starting point for our transfer learning stage.

\section{Ethics}

This research is conducted with a focus on advancing the accuracy and realism of 3D face capture technologies for legitimate and ethical applications, such as improving visual effects in the entertainment industry and enhancing user experiences in virtual and augmented reality. Our work is not intended for the creation of deepfakes or any unconsented face modifications. We firmly oppose the use of our method for malicious purposes, including the generation of misleading or harmful content. Our research is aimed solely at contributing to the scientific community and legitimate industry practices. We advocate for continued discussion and regulation to ensure that advancements in this field are used in ways that respect individual rights and community standards.

\section{Conclusion}
 
In this paper we present a methodology that performs a reconstruction of 3D face geometry and appearance from a collection of sources, which is then exploited to inform a transfer learning process. By replacing the decoder of a pre-trained monocular face reconstruction method with our personalized model, we achieve a more accurate self-supervision objective. This enables more precise expression and pose alignment, resulting in a new encoder capable or reconstructing 3D faces with expressive details from unseen images in real-time.

While current state-of-the-art methods offer real-time regression of parametric 3D face models, they fall short in providing the high-fidelity geometry required for high-end visual effects. By incorporating a more personalized approach, we address these limitations and demonstrate that leveraging unconstrained reference imagery can significantly enhance facial tracking quality.

Our results hold promising implications for the field of visual effects and other applications where reference material of a person is available. Overall, we contribute a novel and effective strategy for improving the fidelity of monocular face capture, paving the way for fully automated production-ready face capture.

\section{Acknowledgments}

This project has received funding from the Association Nationale de la Recherche et de la Technologie under CIFRE agreement No 2022\_1630.

%%%% insert bib %%%%%%
% Bibliography
\bibliographystyle{ACM-Reference-Format}
\bibliography{bibliography}

%%%% figures only page %%%%
\begin{figure*}[t]
    \setlength{\tabcolsep}{0.0em} % horizontal padding
    \def\arraystretch{0.0}{ % vertical padding
    \renewcommand{\w}{0.07\linewidth}
    \begin{tabular}[t]{l@{\hspace{1em}}r}
        \begin{tabular}[t]{ccccccc}
            Input & DECA & EMOCA & EMOCA-t & SMIRK & SPARK
            & SPARK
            \\
            \includegraphics[width=\w]{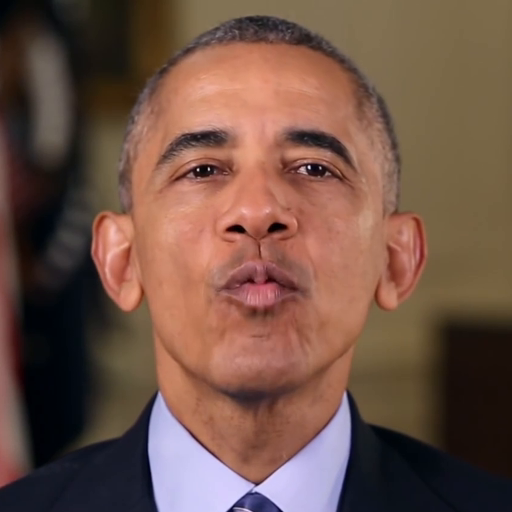} &
            \includegraphics[width=\w]{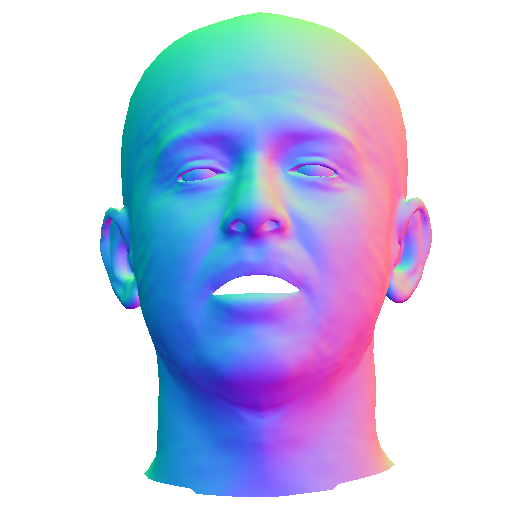} &
            \includegraphics[width=\w]{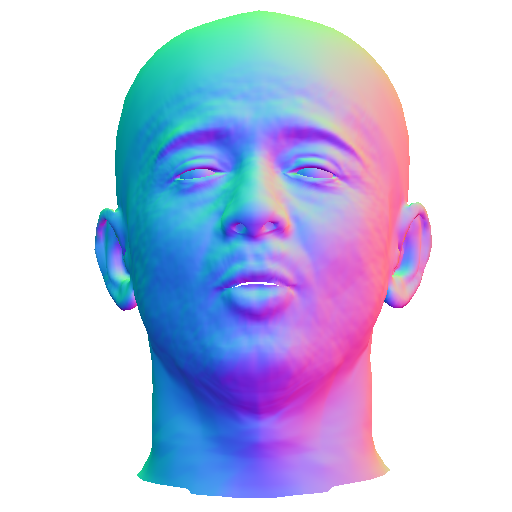} &
            \includegraphics[width=\w]{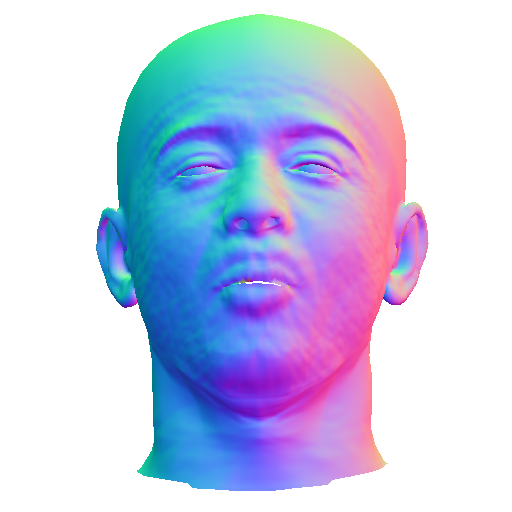} &
            \includegraphics[width=\w]{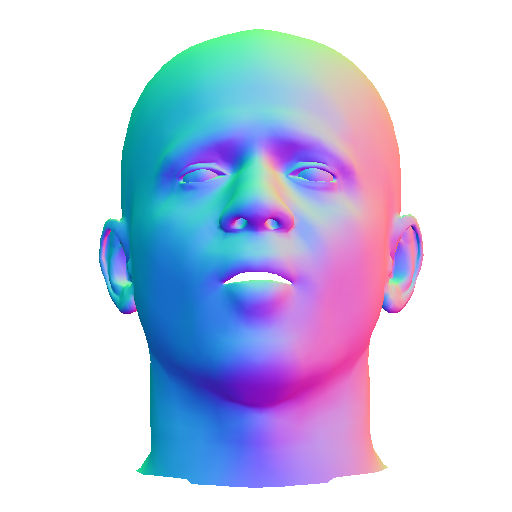} &
            \includegraphics[width=\w]{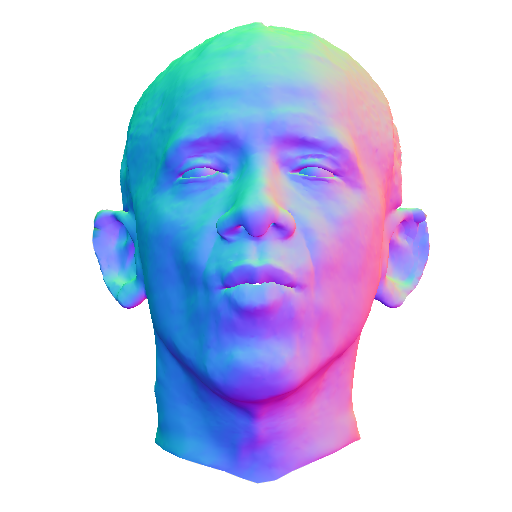}
            & \includegraphics[width=\w]{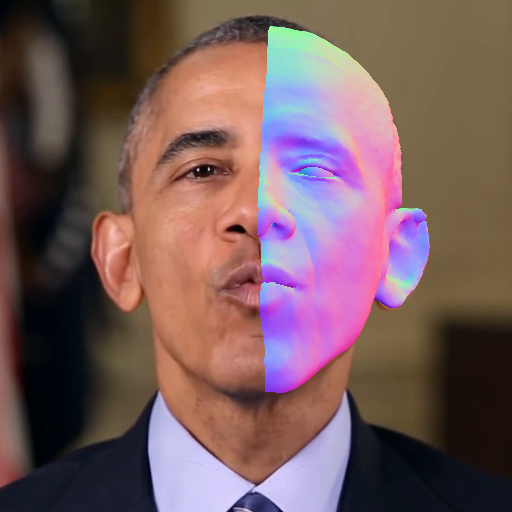}
            \\
            \includegraphics[width=\w]{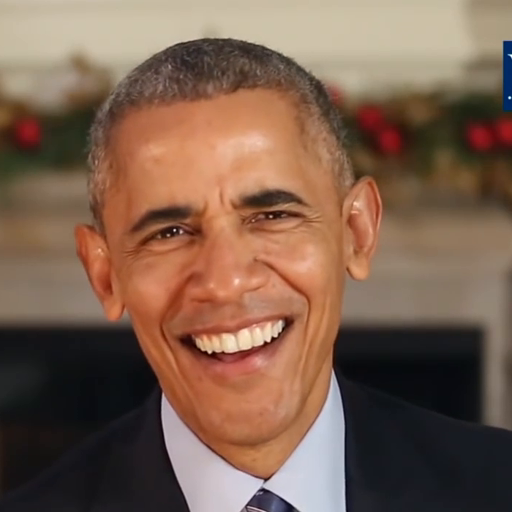} &
            \includegraphics[width=\w]{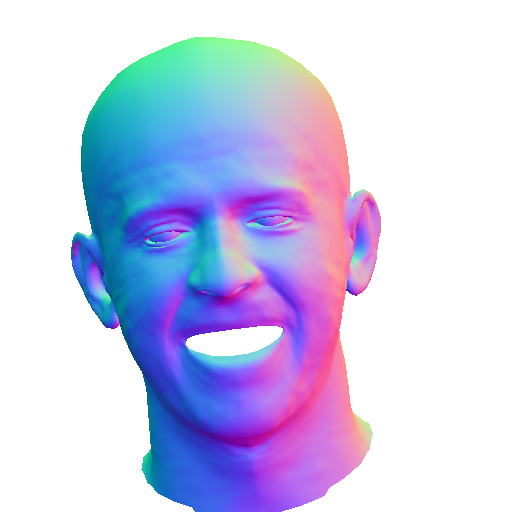} &
            \includegraphics[width=\w]{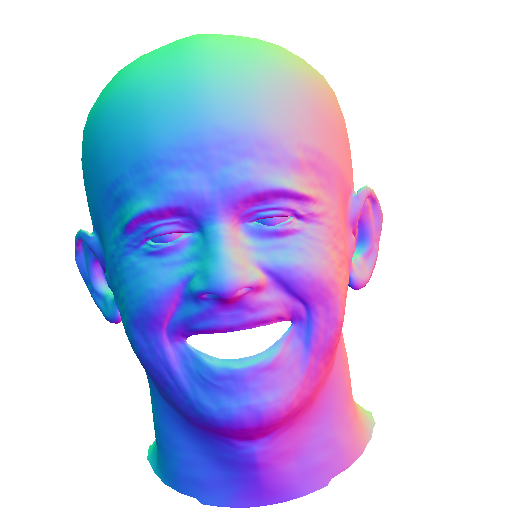} &
            \includegraphics[width=\w]{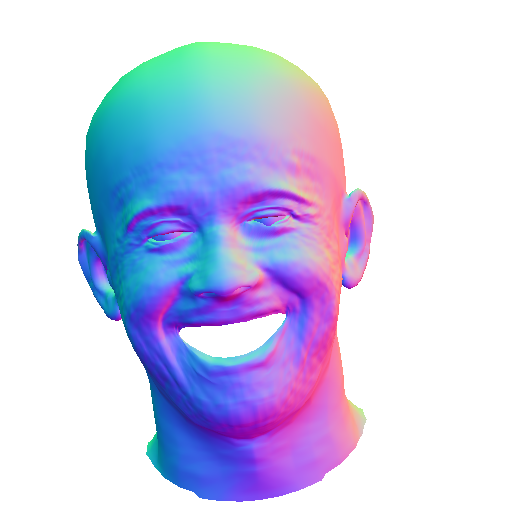} &
            \includegraphics[width=\w]{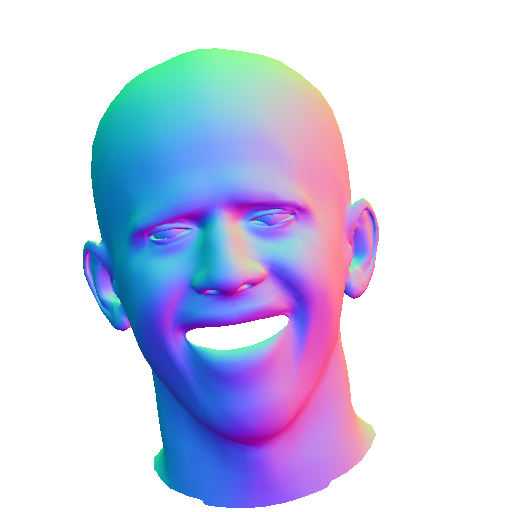} &
            \includegraphics[width=\w]{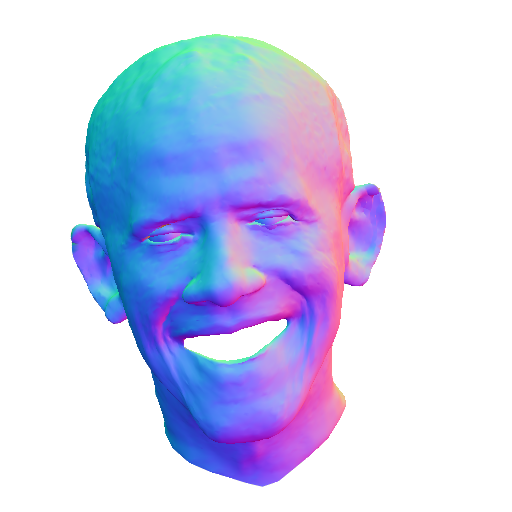}
            & \includegraphics[width=\w]{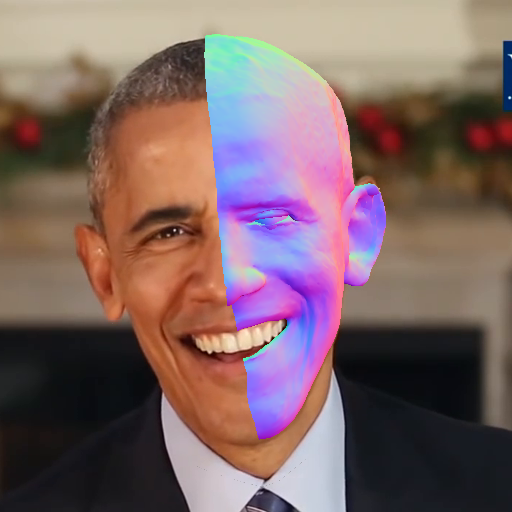}
            \\
            \includegraphics[width=\w]{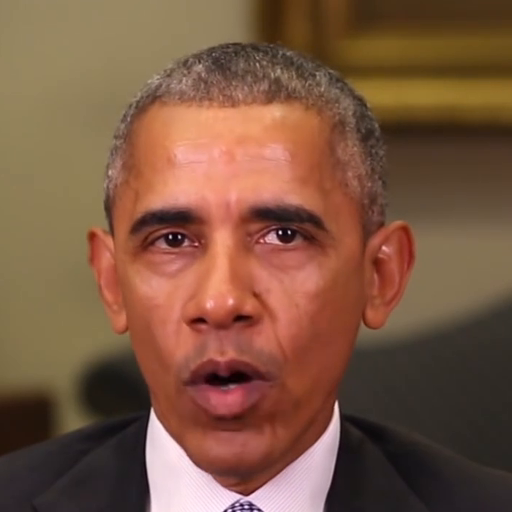} &
            \includegraphics[width=\w]{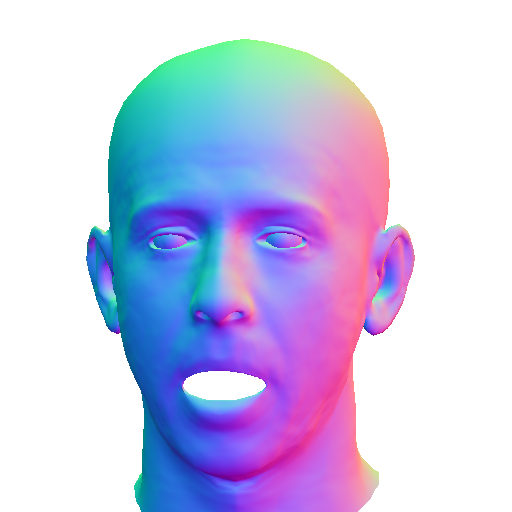} &
            \includegraphics[width=\w]{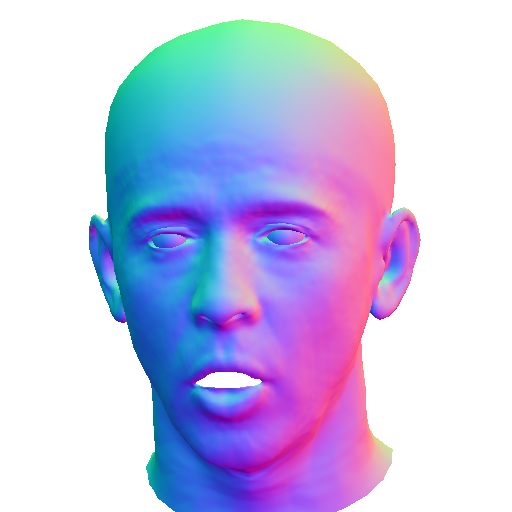} &
            \includegraphics[width=\w]{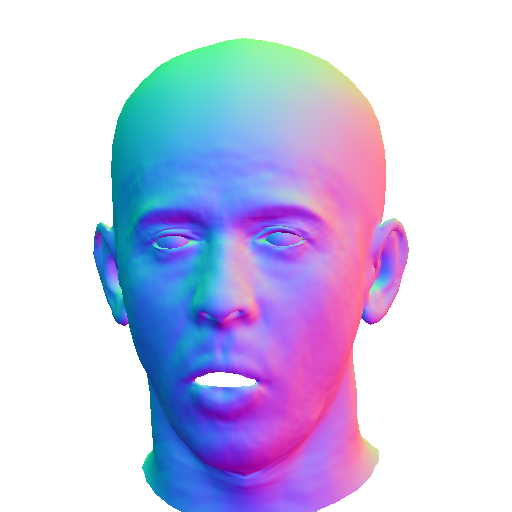} &
            \includegraphics[width=\w]{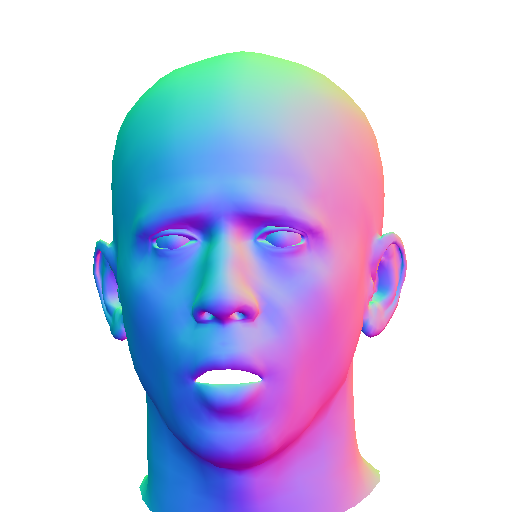} &
            \includegraphics[width=\w]{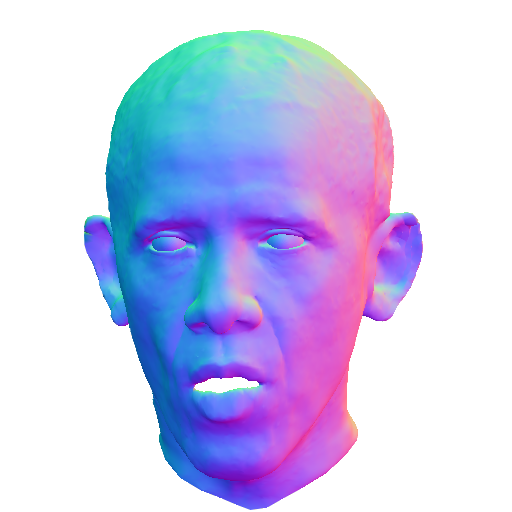}
            & \includegraphics[width=\w]{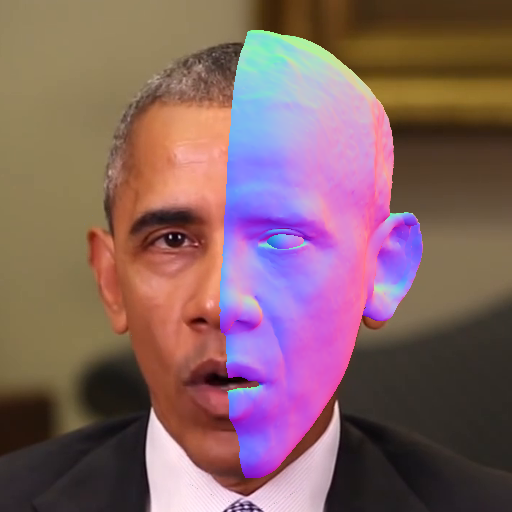}
            \\
            \includegraphics[width=\w]{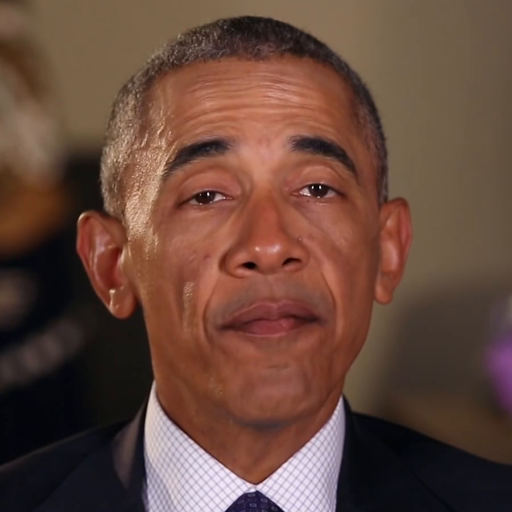} &
            \includegraphics[width=\w]{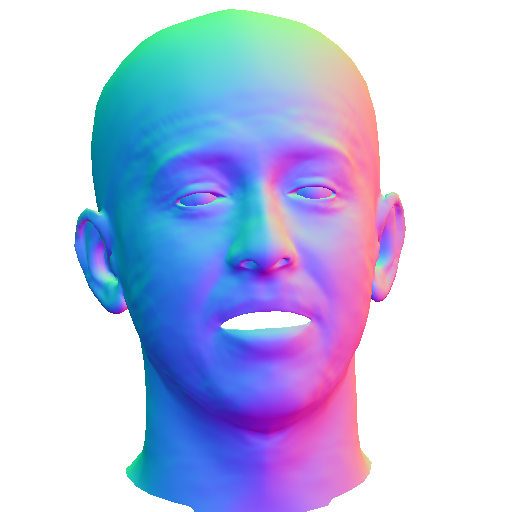} &
            \includegraphics[width=\w]{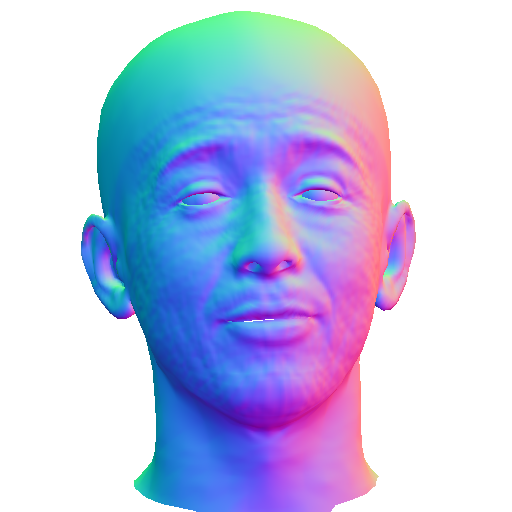} &
            \includegraphics[width=\w]{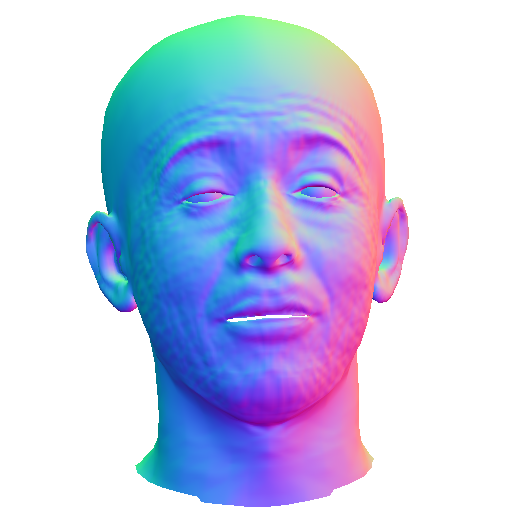} &
            \includegraphics[width=\w]{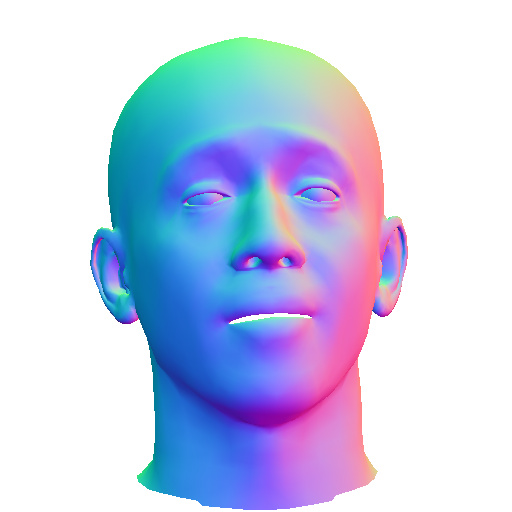} &
            \includegraphics[width=\w]{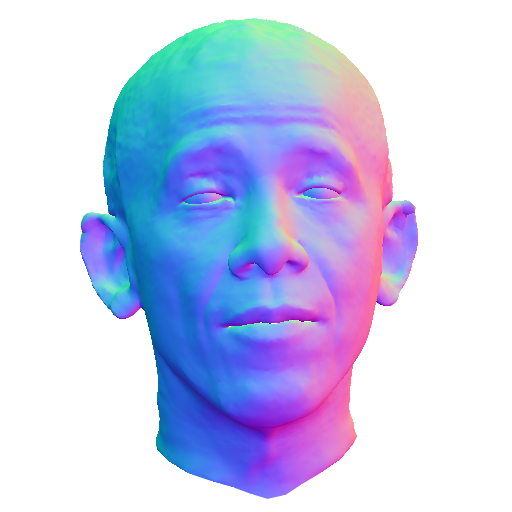}
            & \includegraphics[width=\w]{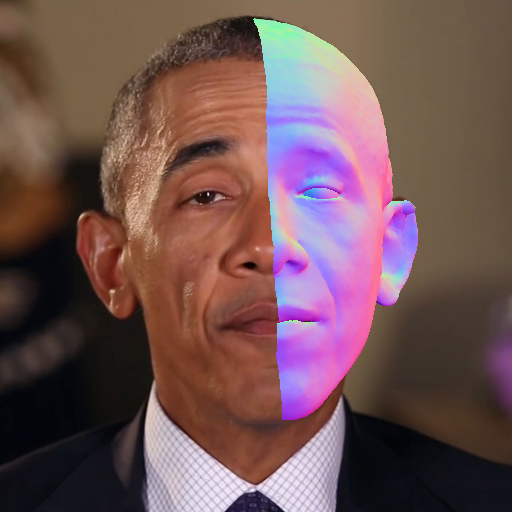}
            \\
            \includegraphics[width=\w]{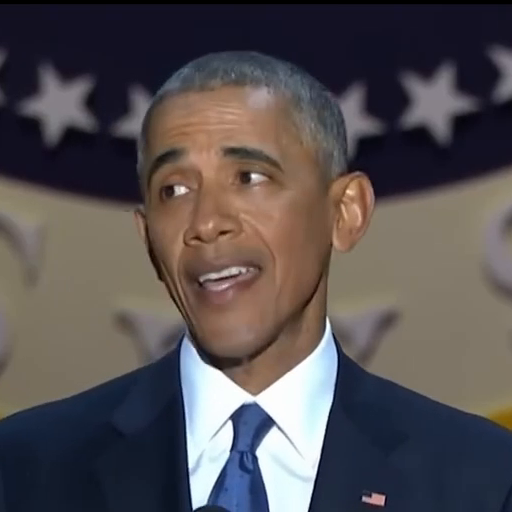} &
            \includegraphics[width=\w]{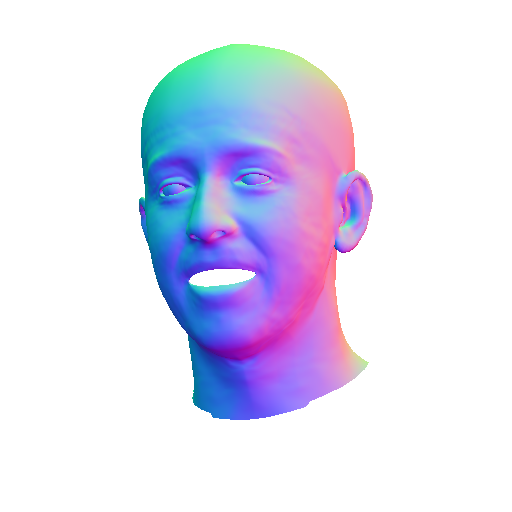} &
            \includegraphics[width=\w]{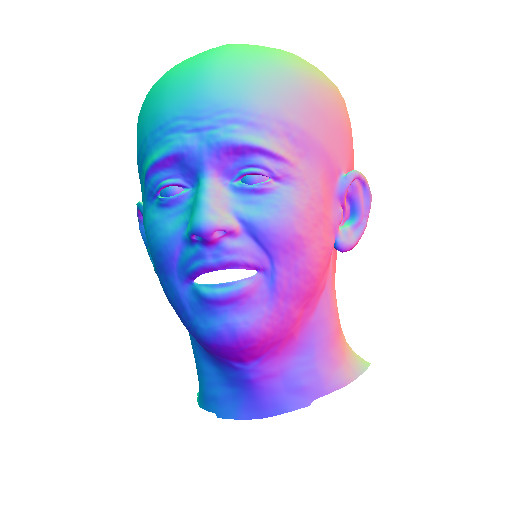} &
            \includegraphics[width=\w]{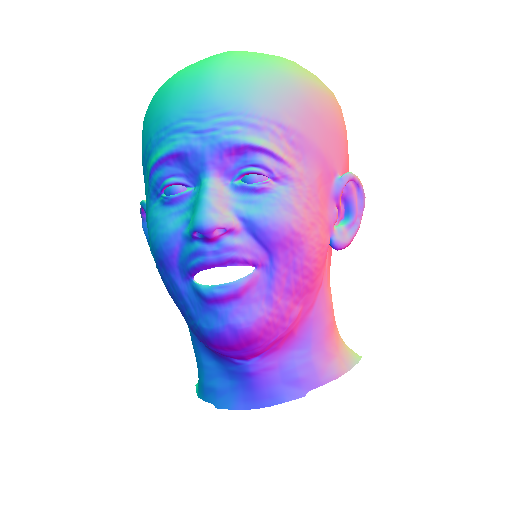} &
            \includegraphics[width=\w]{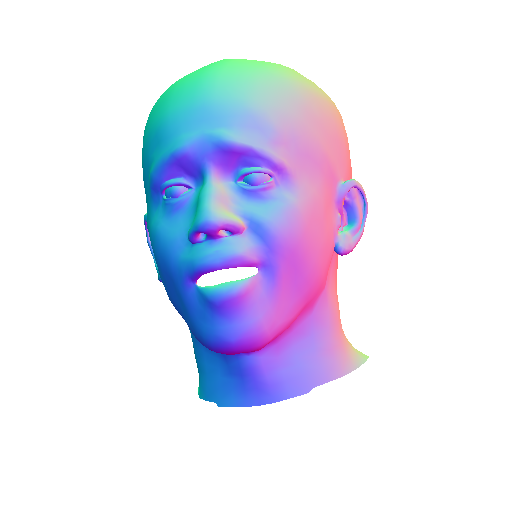} &
            \includegraphics[width=\w]{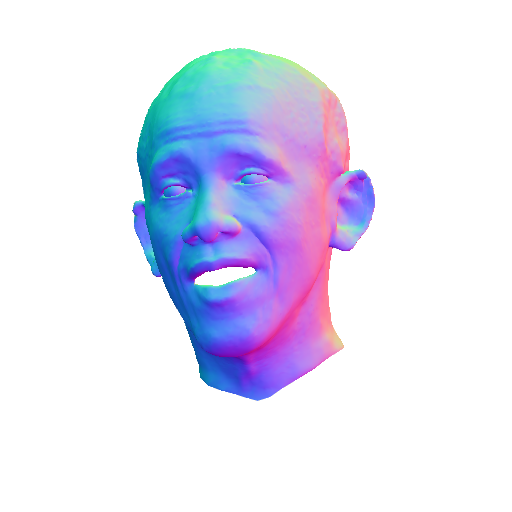}
            & \includegraphics[width=\w]{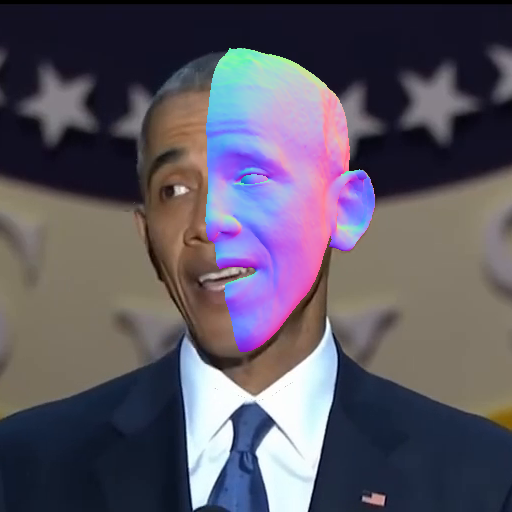}
            \\
            \includegraphics[width=\w]{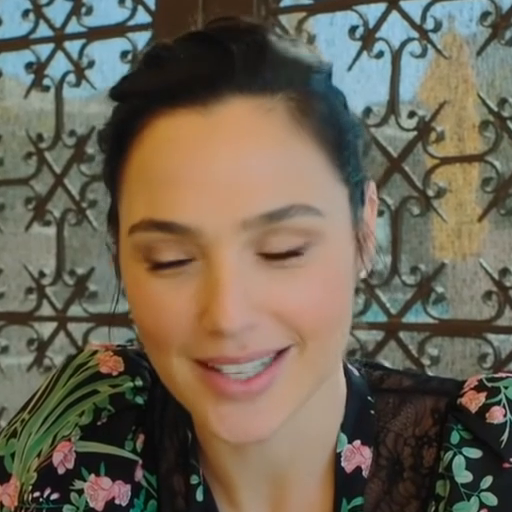} &
            \includegraphics[width=\w]{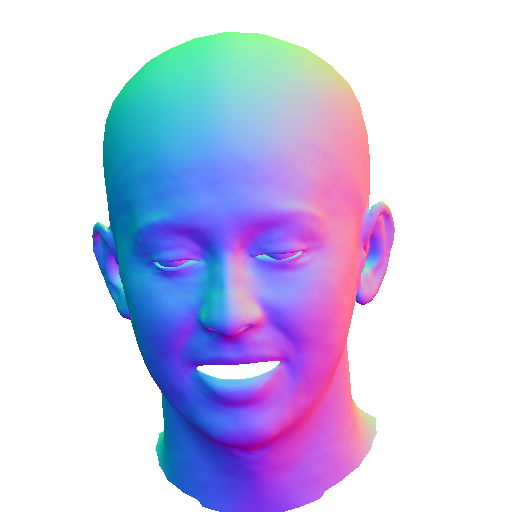} &
            \includegraphics[width=\w]{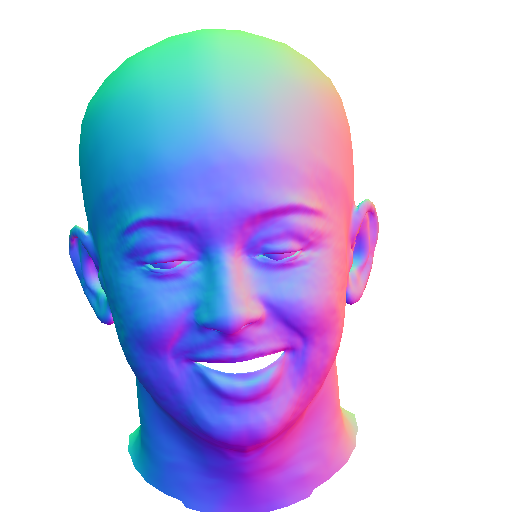} &
            \includegraphics[width=\w]{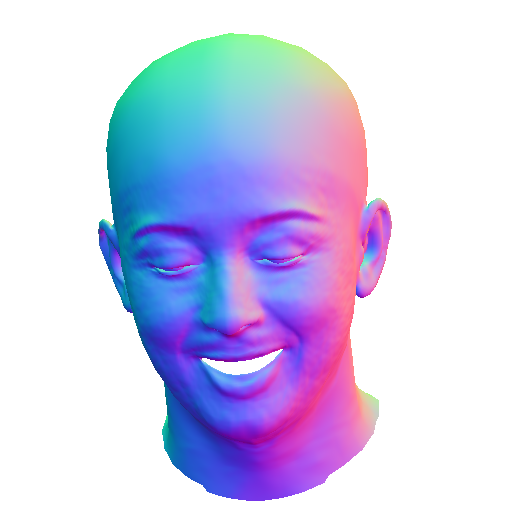} &
            \includegraphics[width=\w]{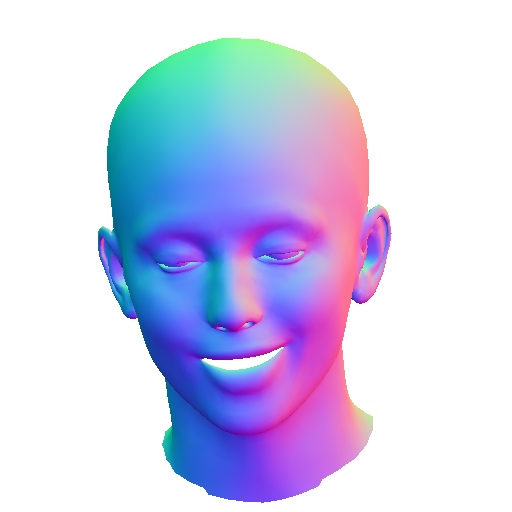} &
            \includegraphics[width=\w]{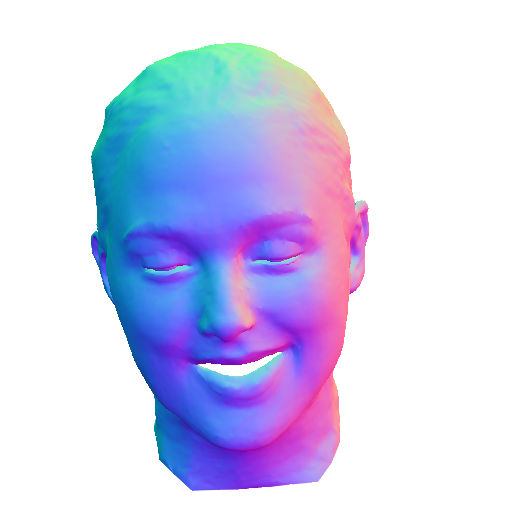}
            & \includegraphics[width=\w]{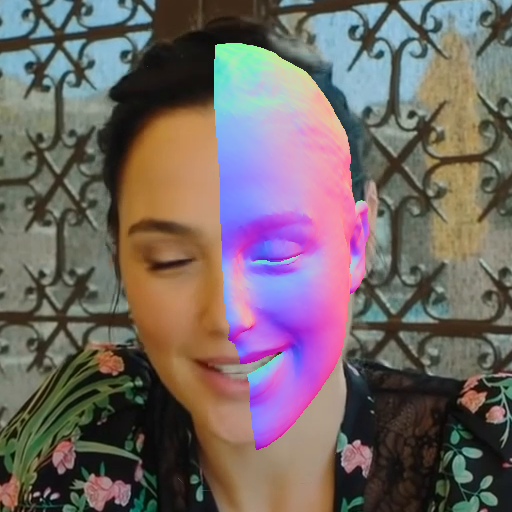}
            \\
            \includegraphics[width=\w]{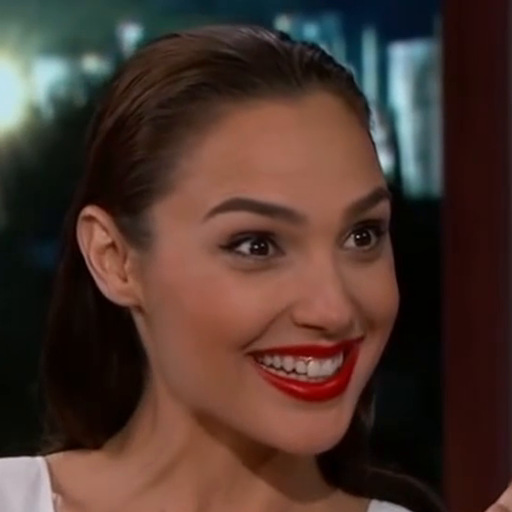} &
            \includegraphics[width=\w]{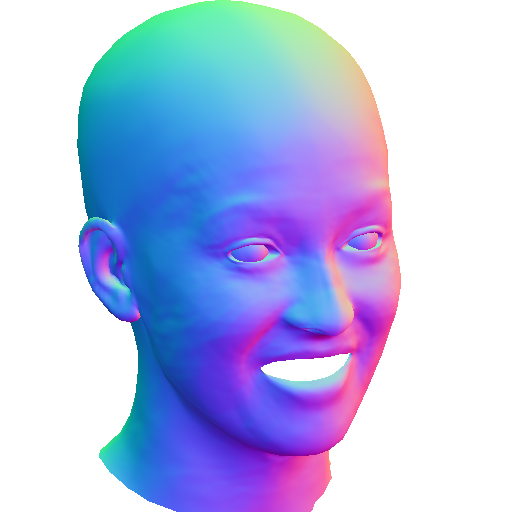} &
            \includegraphics[width=\w]{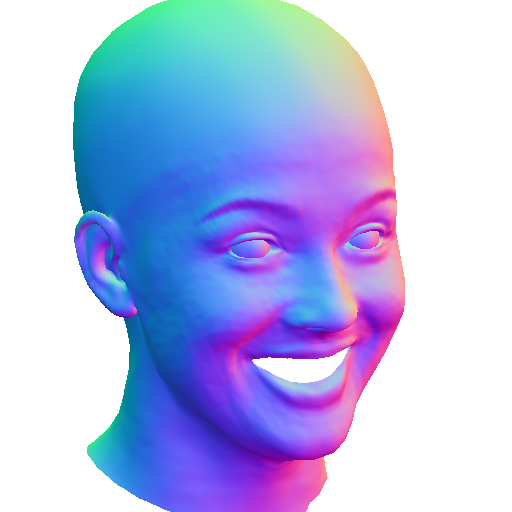} &
            \includegraphics[width=\w]{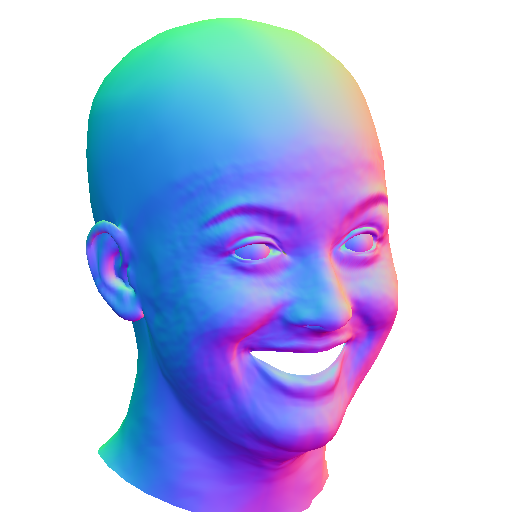} &
            \includegraphics[width=\w]{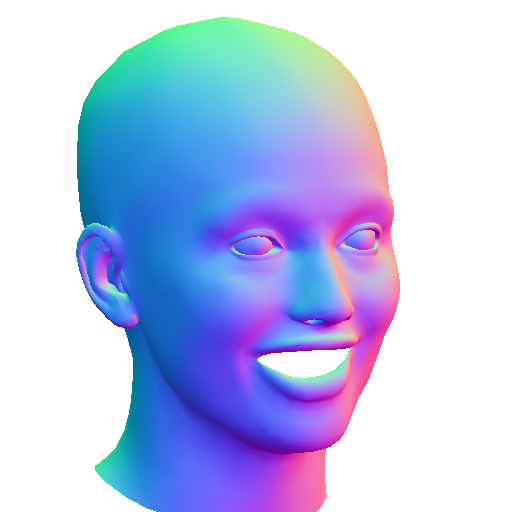} &
            \includegraphics[width=\w]{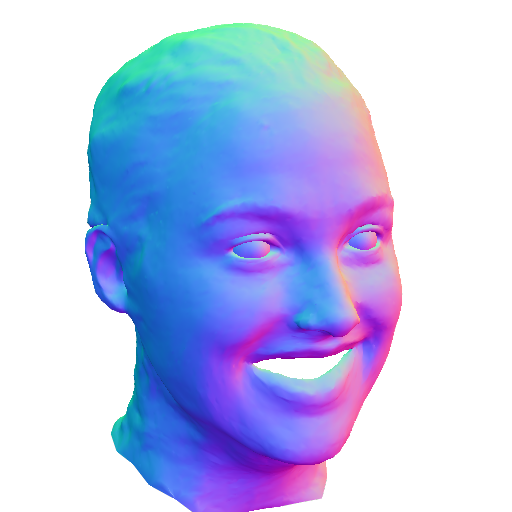}
            & \includegraphics[width=\w]{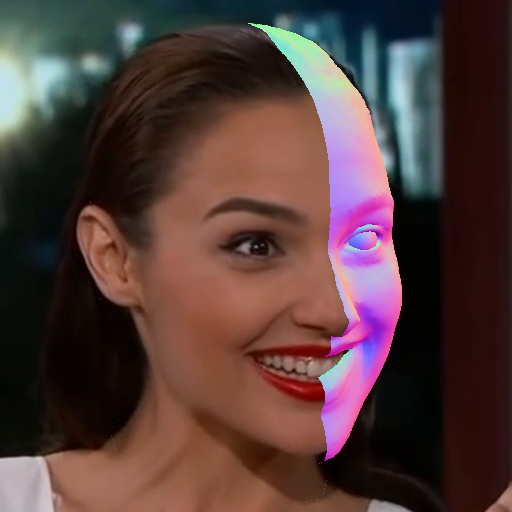}
            \\
            \includegraphics[width=\w]{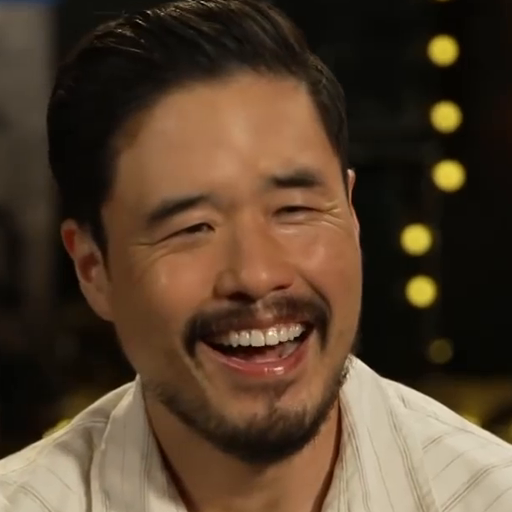} &
            \includegraphics[width=\w]{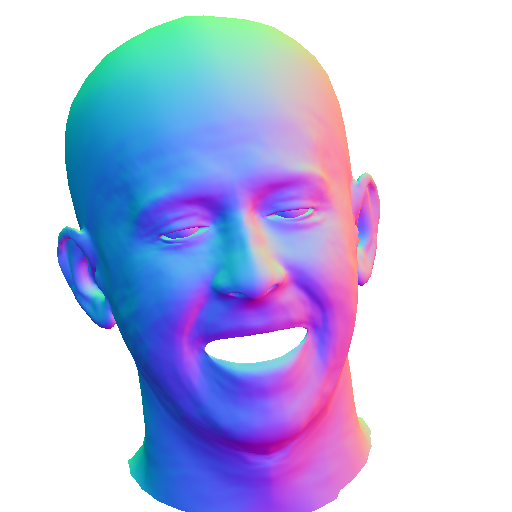} &
            \includegraphics[width=\w]{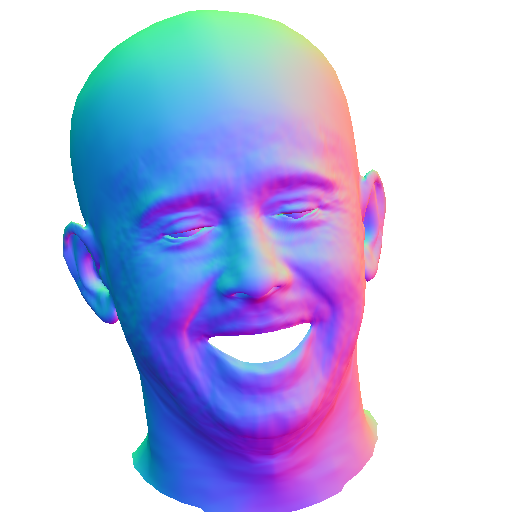} &
            \includegraphics[width=\w]{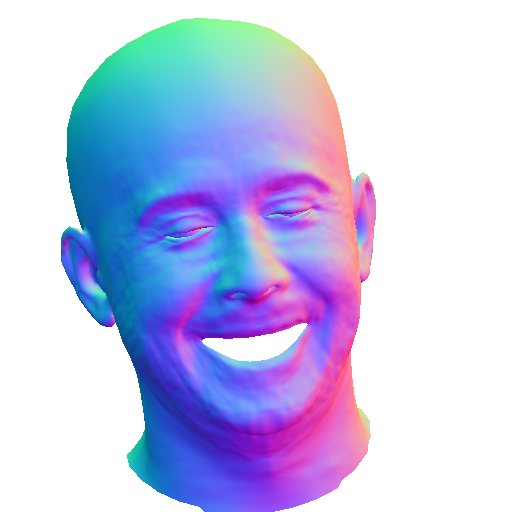} &
            \includegraphics[width=\w]{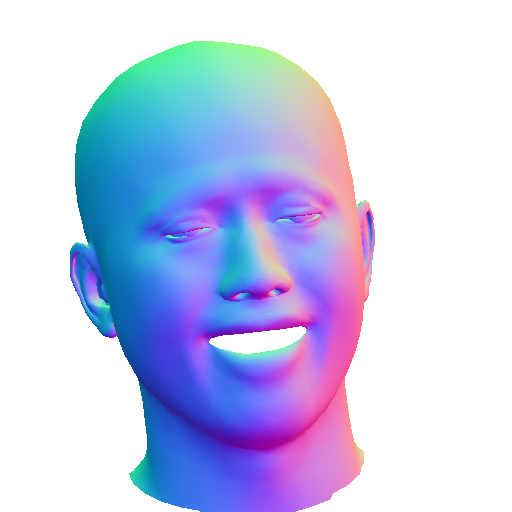} &
            \includegraphics[width=\w]{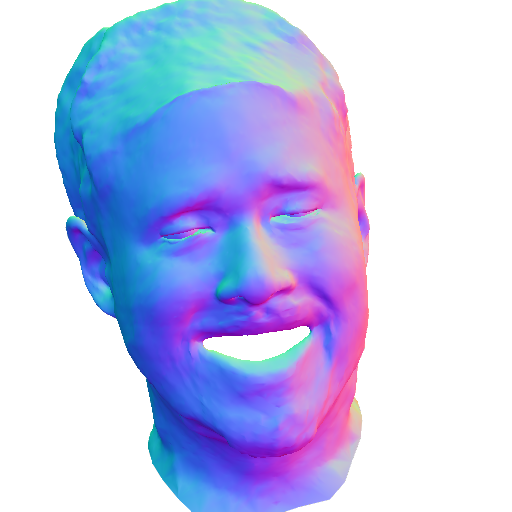}
            & \includegraphics[width=\w]{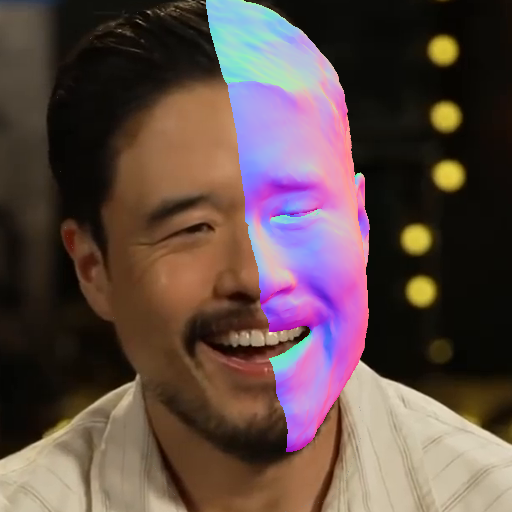}
        \end{tabular}
        &
        \begin{tabular}[t]{ccccccc}
            Input & DECA & EMOCA & EMOCA-t & SMIRK & SPARK
            & SPARK
            \\
            \includegraphics[width=\w]{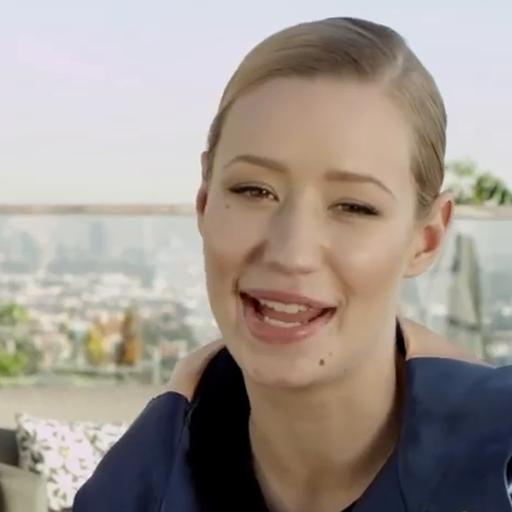} &
            \includegraphics[width=\w]{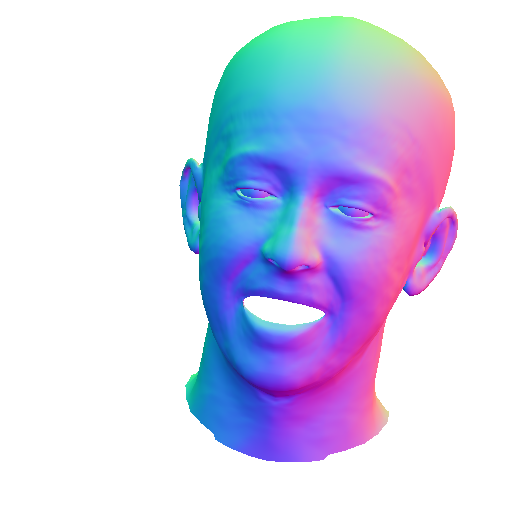} &
            \includegraphics[width=\w]{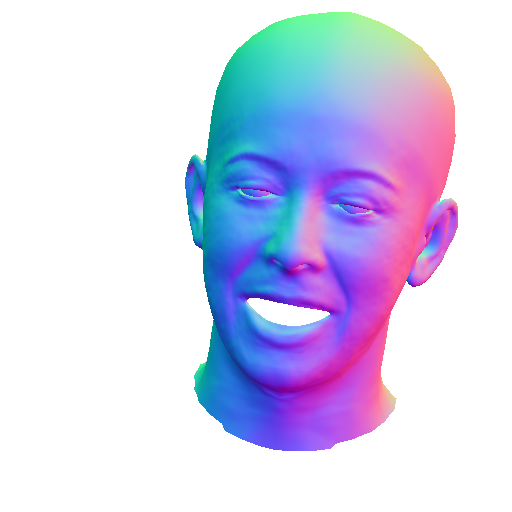} &
            \includegraphics[width=\w]{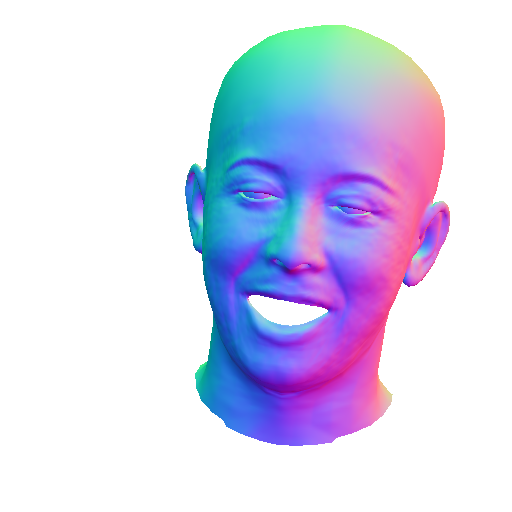} &
            \includegraphics[width=\w]{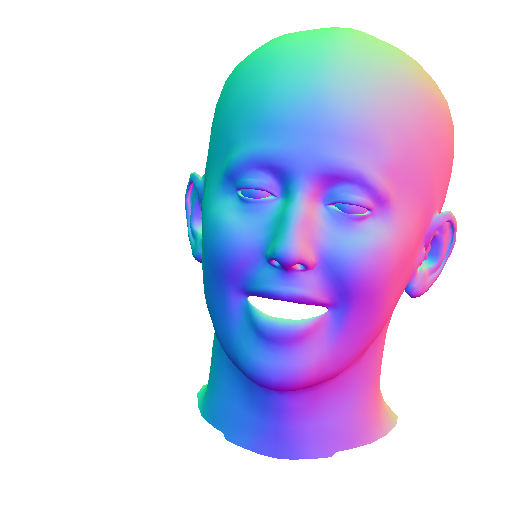} &
            \includegraphics[width=\w]{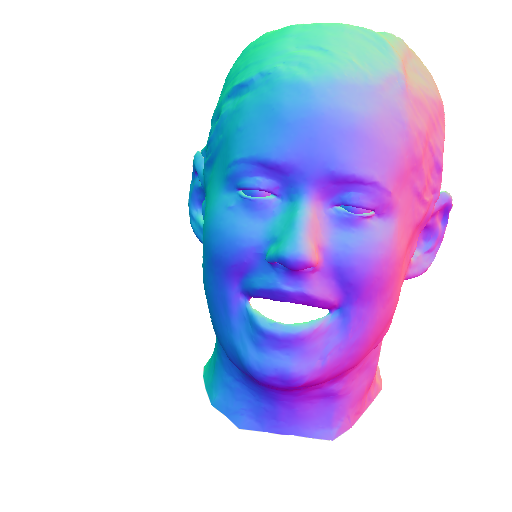}
            & \includegraphics[width=\w]{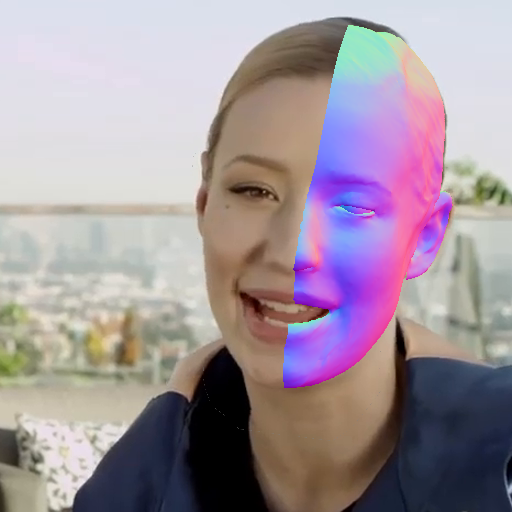}
            \\
            \includegraphics[width=\w]{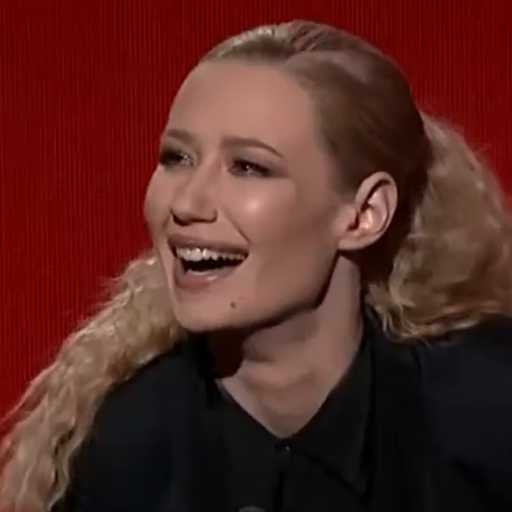} &
            \includegraphics[width=\w]{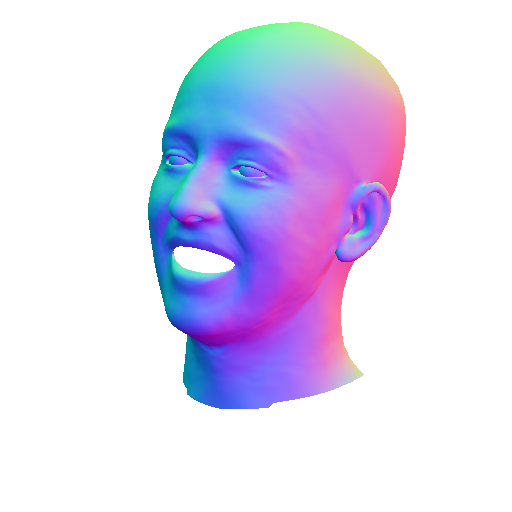} &
            \includegraphics[width=\w]{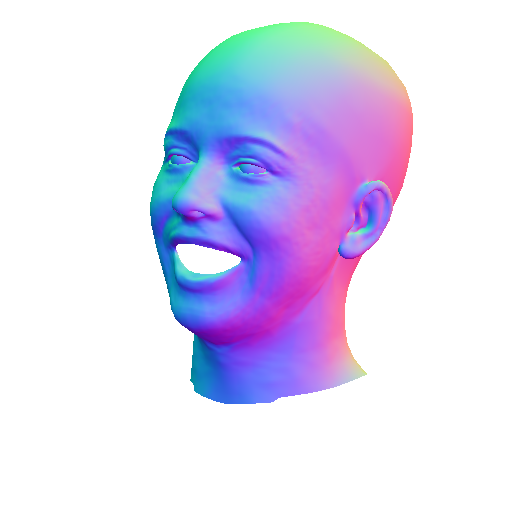} &
            \includegraphics[width=\w]{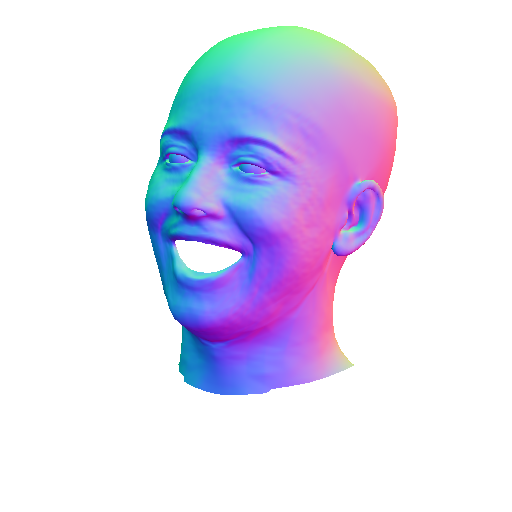} &
            \includegraphics[width=\w]{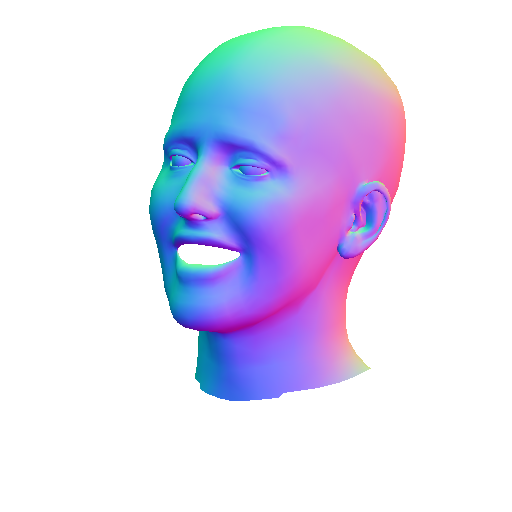} &
            \includegraphics[width=\w]{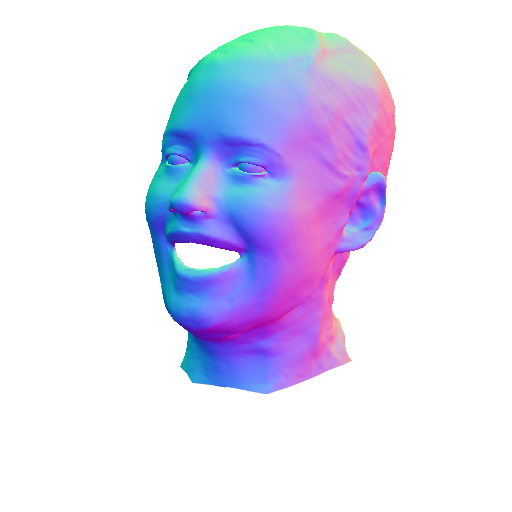}
            & \includegraphics[width=\w]{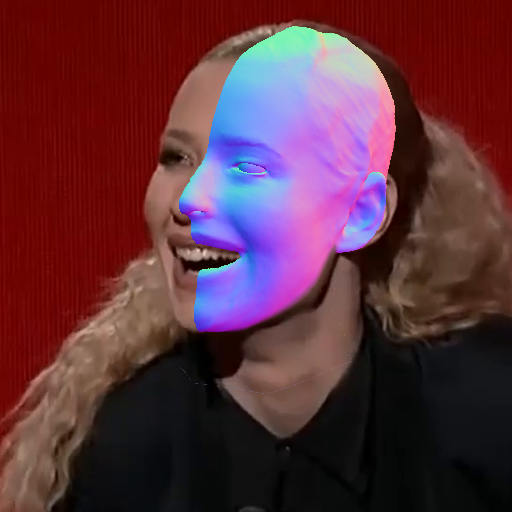}
            \\
            \includegraphics[width=\w]{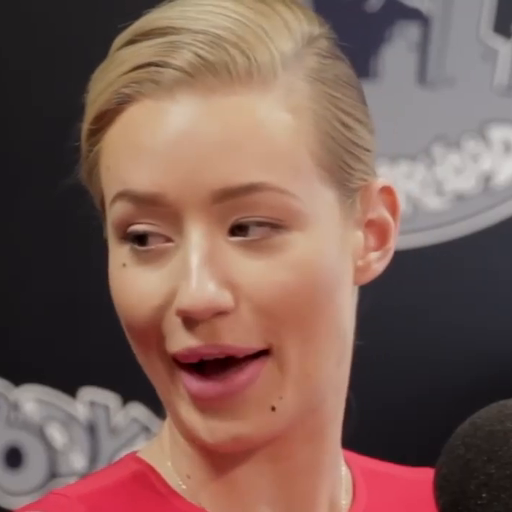} &
            \includegraphics[width=\w]{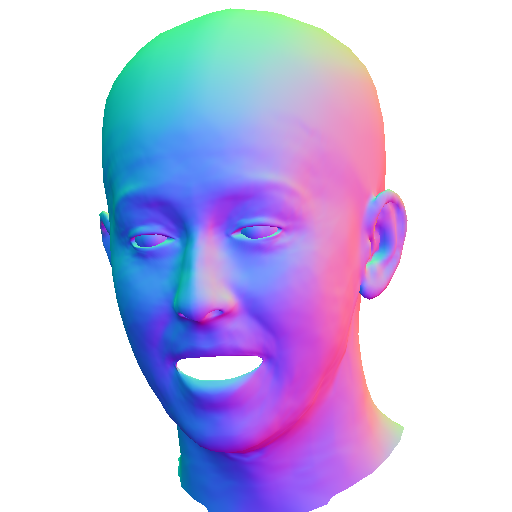} &
            \includegraphics[width=\w]{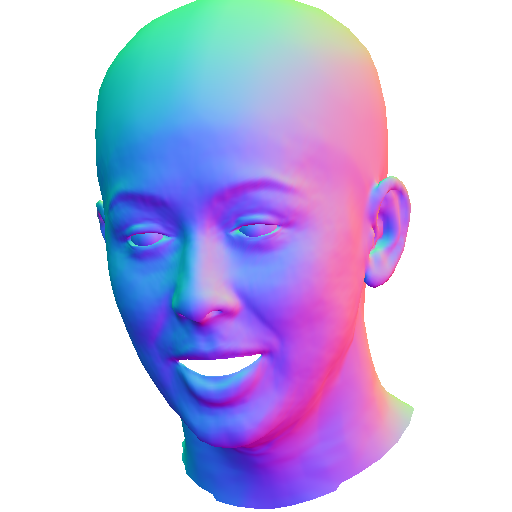} &
            \includegraphics[width=\w]{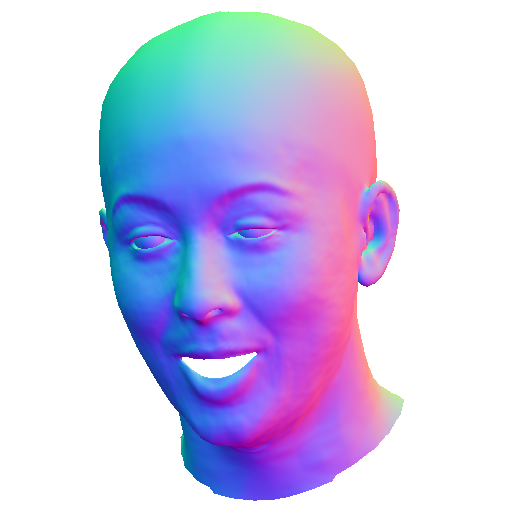} &
            \includegraphics[width=\w]{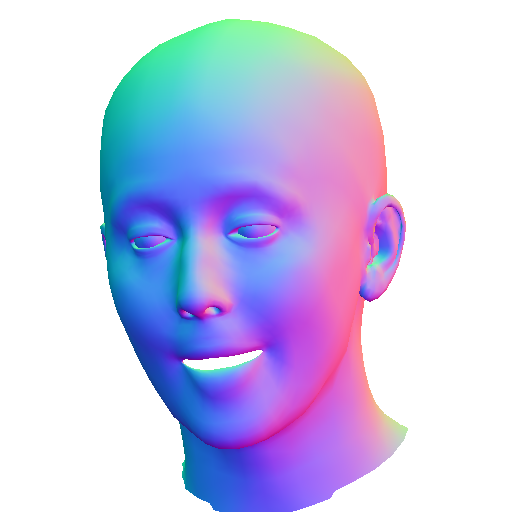} &
            \includegraphics[width=\w]{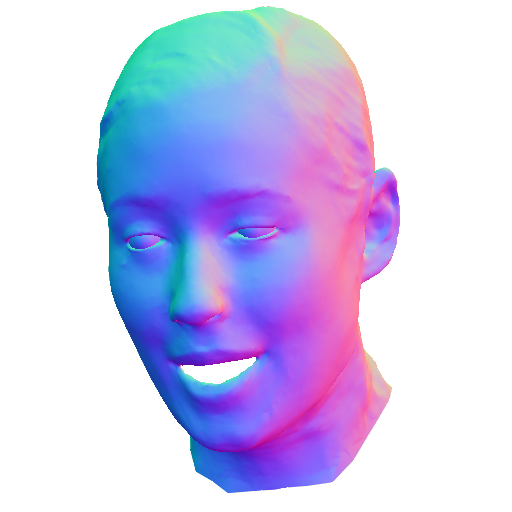}
            & \includegraphics[width=\w]{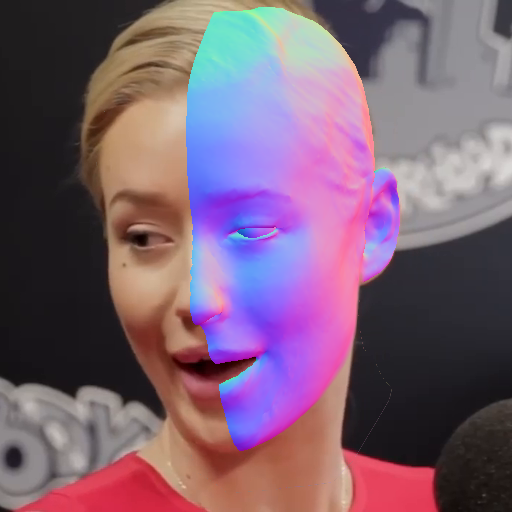}
            \\
            \includegraphics[width=\w]{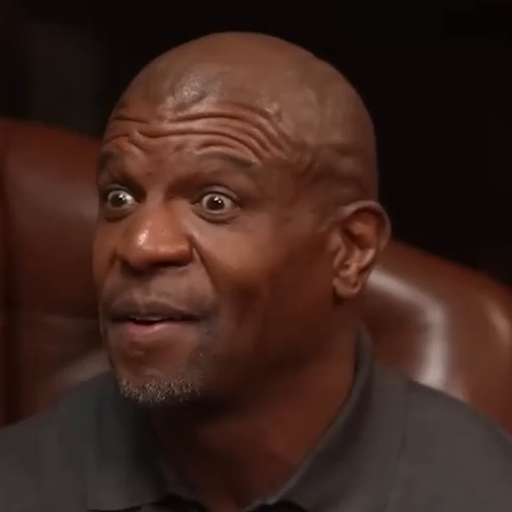} &
            \includegraphics[width=\w]{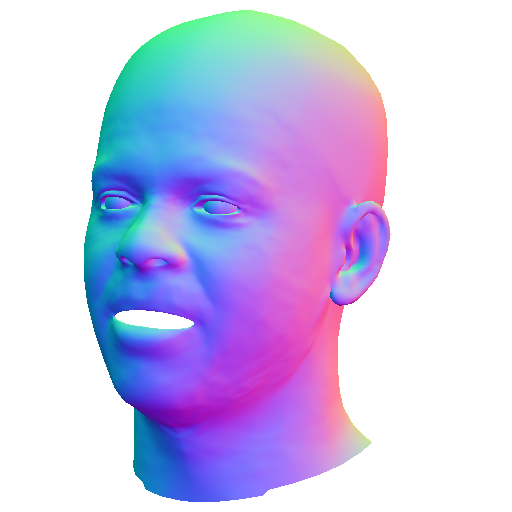} &
            \includegraphics[width=\w]{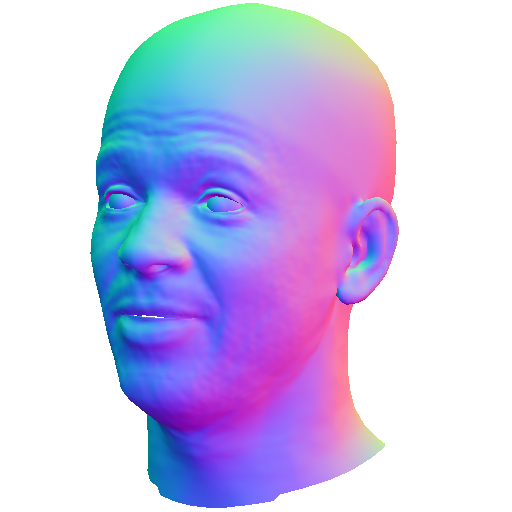} &
            \includegraphics[width=\w]{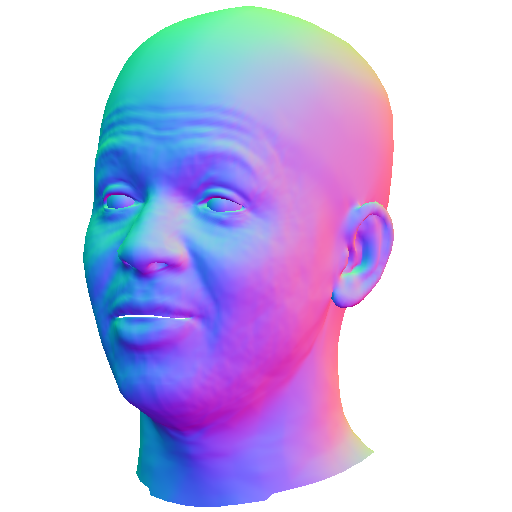} &
            \includegraphics[width=\w]{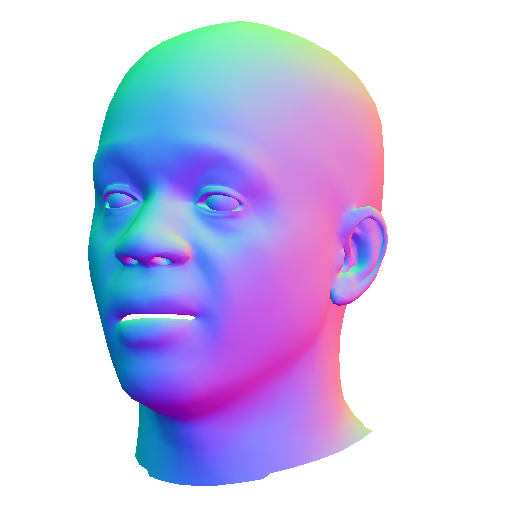} &
            \includegraphics[width=\w]{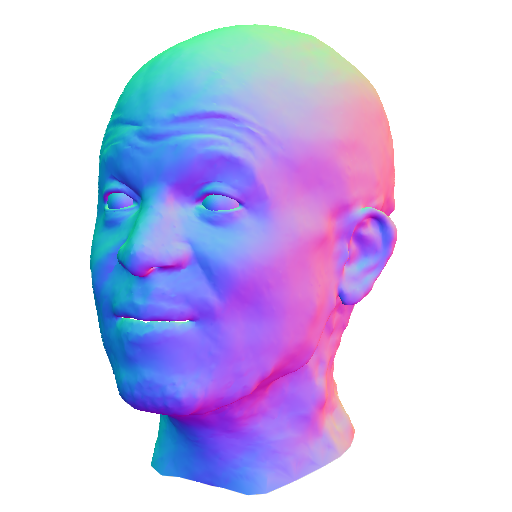}
            & \includegraphics[width=\w]{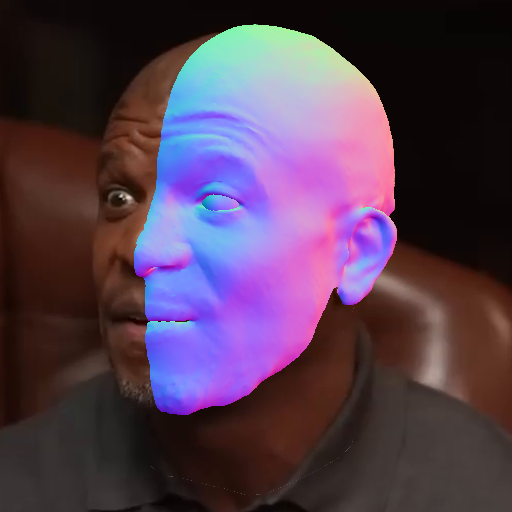}
            \\
            \includegraphics[width=\w]{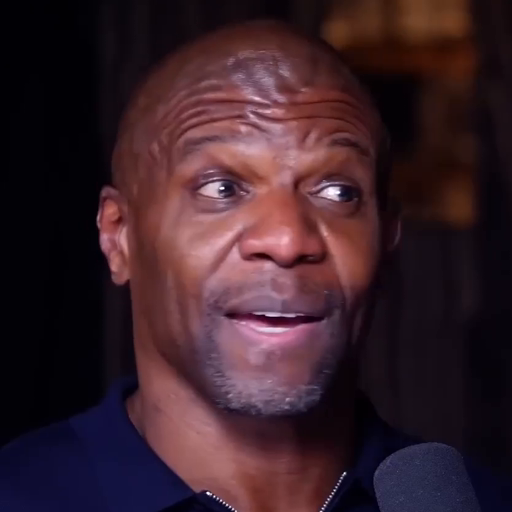} &
            \includegraphics[width=\w]{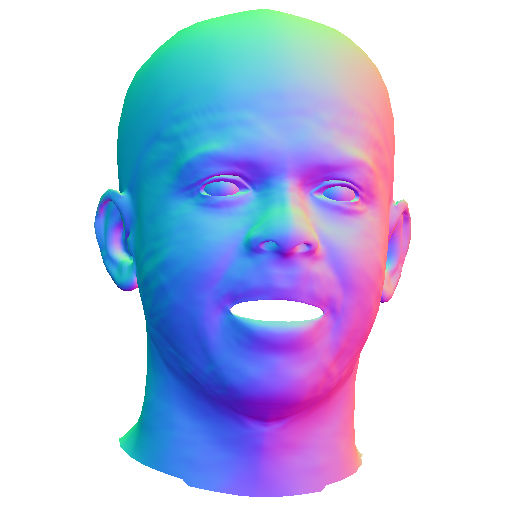} &
            \includegraphics[width=\w]{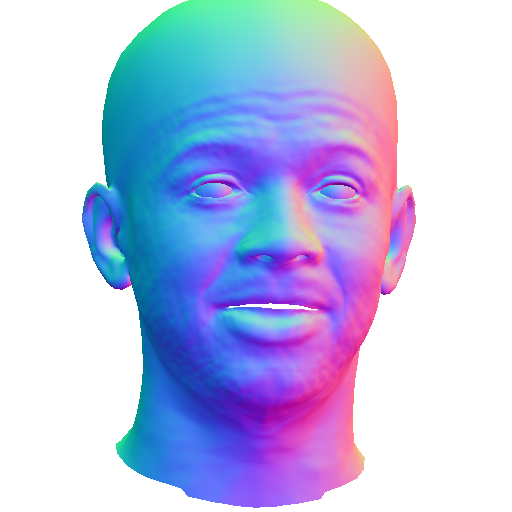} &
            \includegraphics[width=\w]{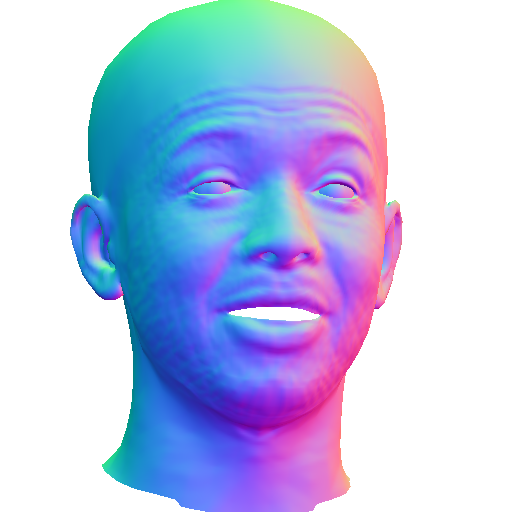} &
            \includegraphics[width=\w]{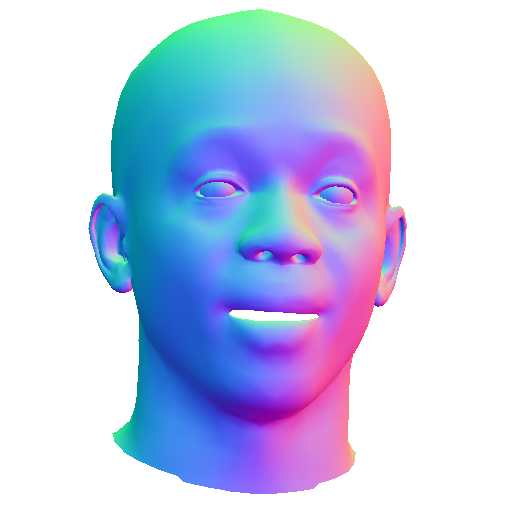} &
            \includegraphics[width=\w]{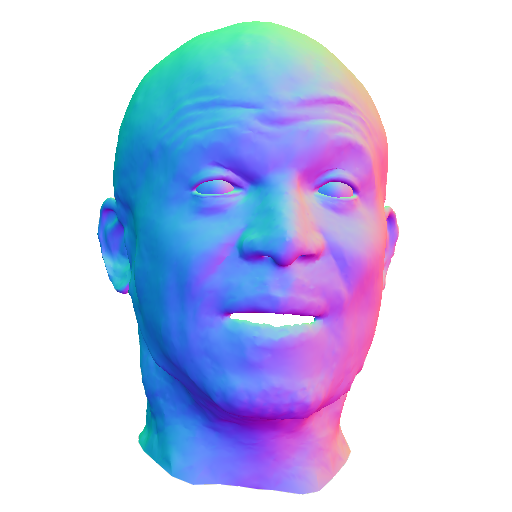}
            & \includegraphics[width=\w]{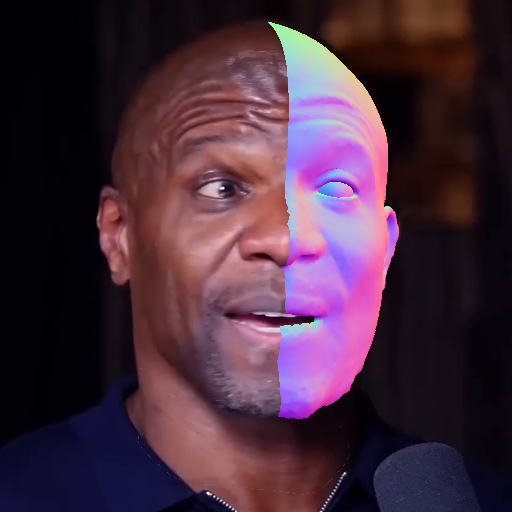}
            \\
            \includegraphics[width=\w]{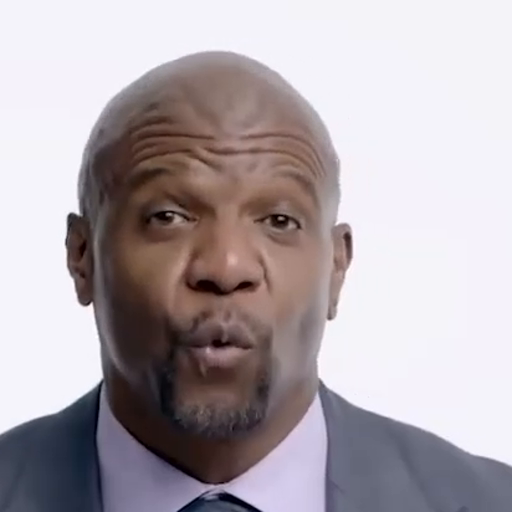} &
            \includegraphics[width=\w]{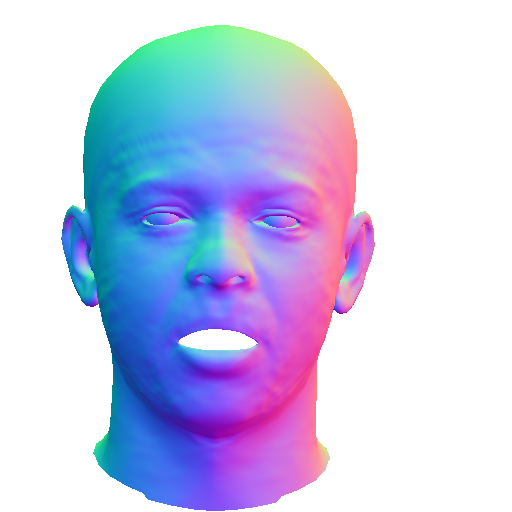} &
            \includegraphics[width=\w]{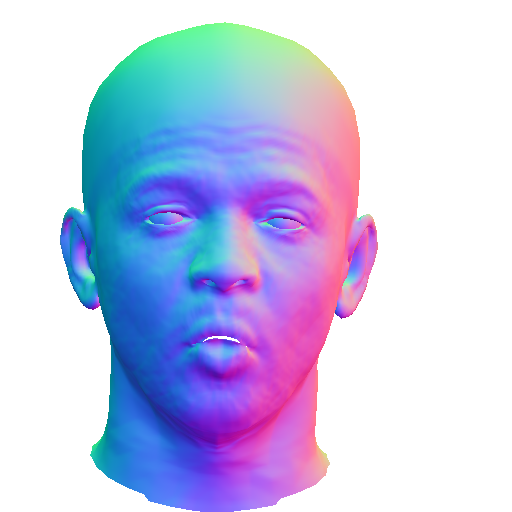} &
            \includegraphics[width=\w]{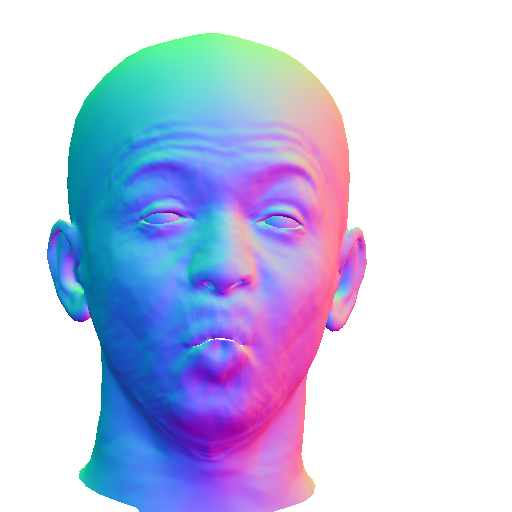} &
            \includegraphics[width=\w]{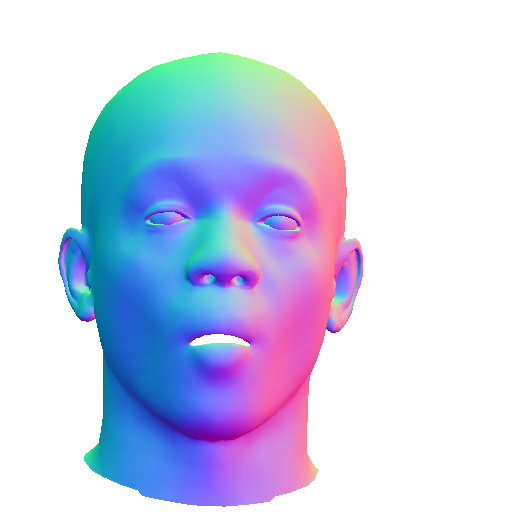} &
            \includegraphics[width=\w]{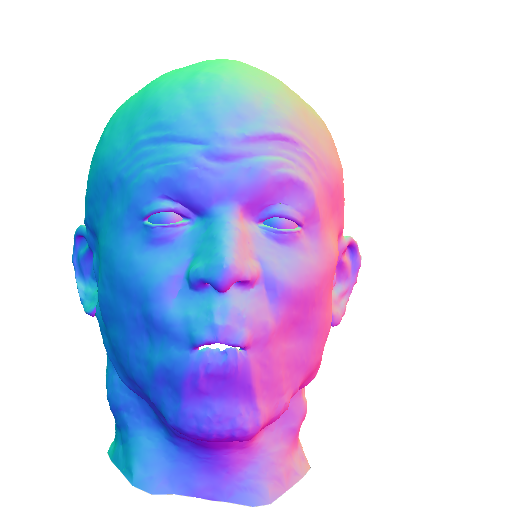}
            & \includegraphics[width=\w]{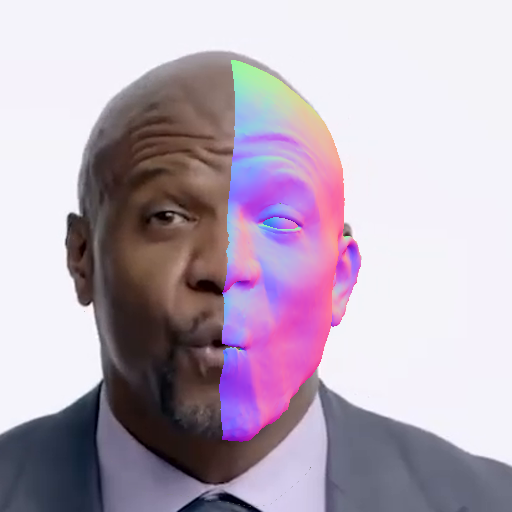}
            \\
            \includegraphics[width=\w]{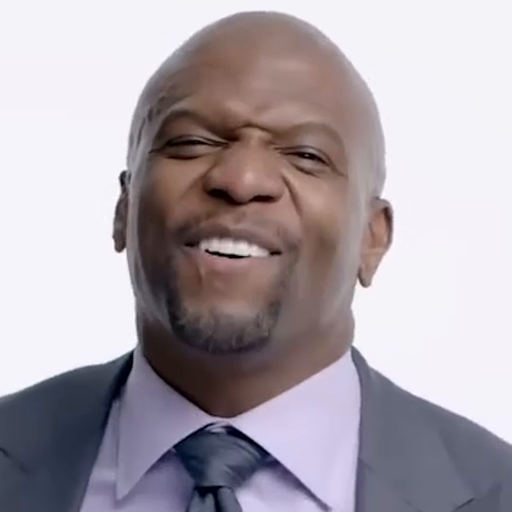} &
            \includegraphics[width=\w]{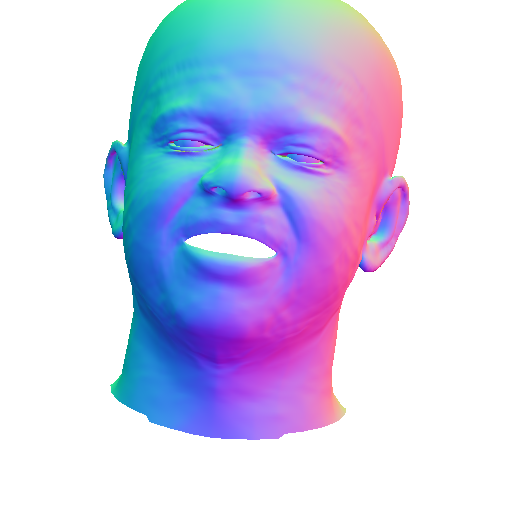} &
            \includegraphics[width=\w]{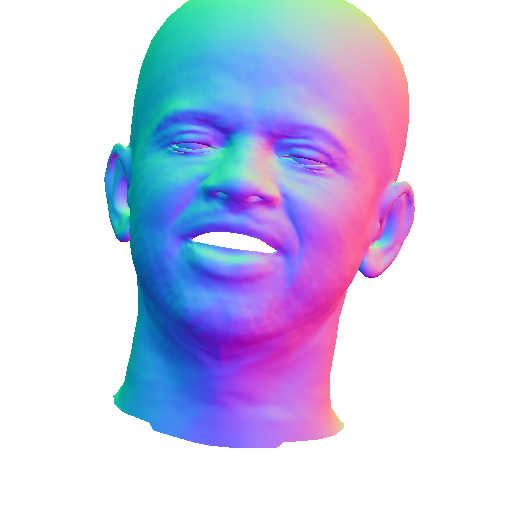} &
            \includegraphics[width=\w]{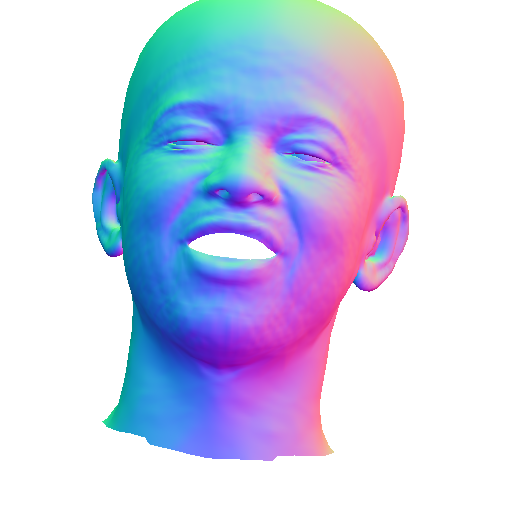} &
            \includegraphics[width=\w]{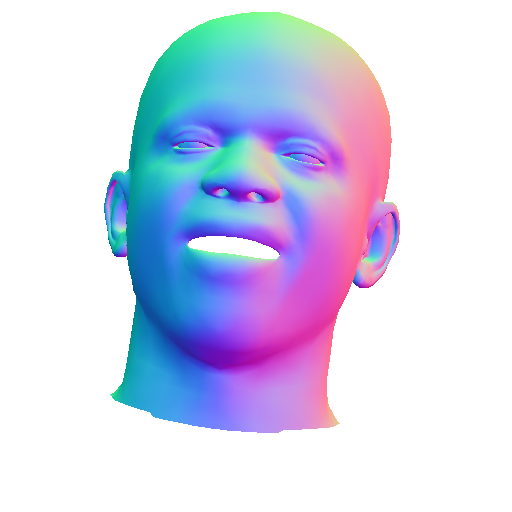} &
            \includegraphics[width=\w]{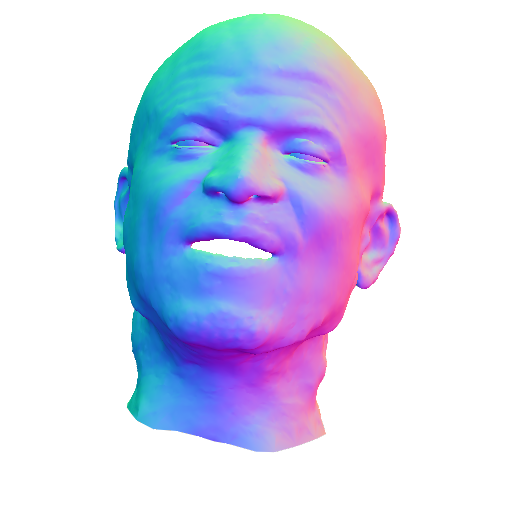}
            & \includegraphics[width=\w]{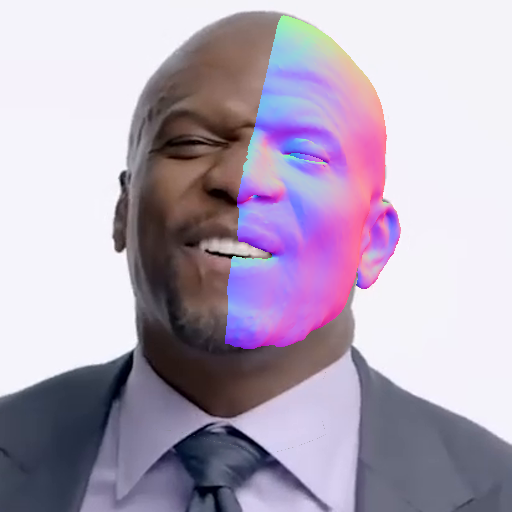}
            \\
            \includegraphics[width=\w]{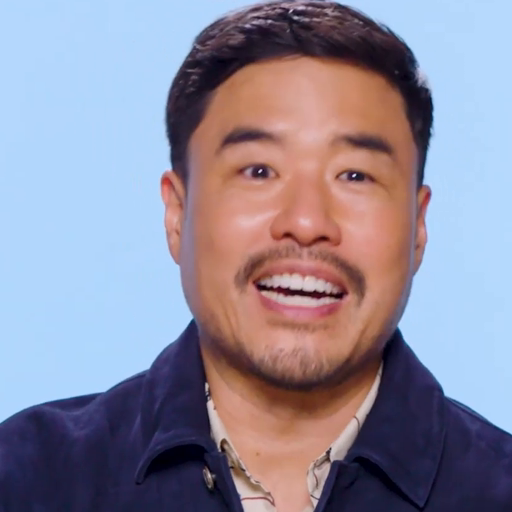} &
            \includegraphics[width=\w]{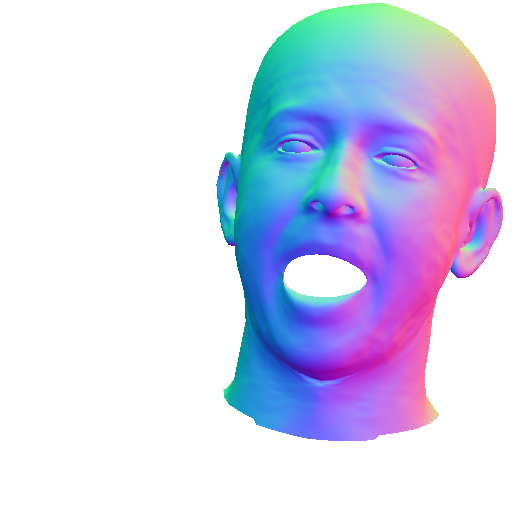} &
            \includegraphics[width=\w]{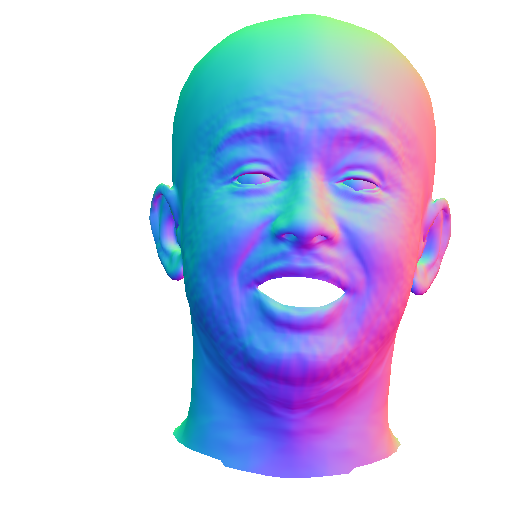} &
            \includegraphics[width=\w]{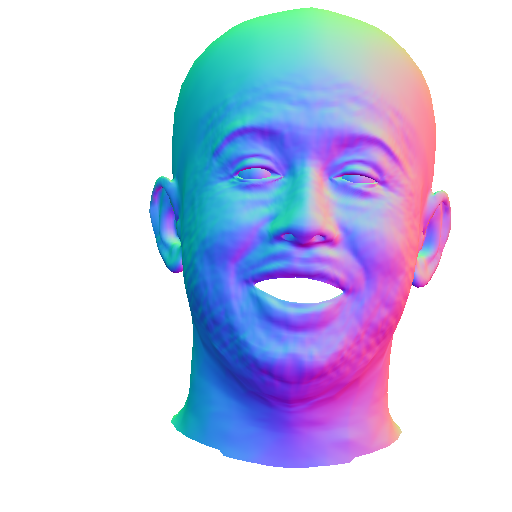} &
            \includegraphics[width=\w]{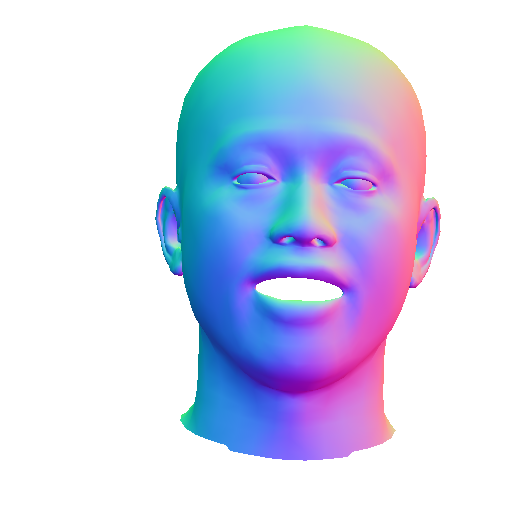} &
            \includegraphics[width=\w]{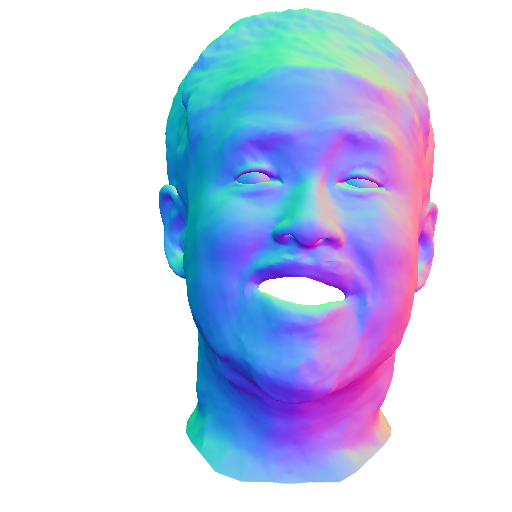}
            & \includegraphics[width=\w]{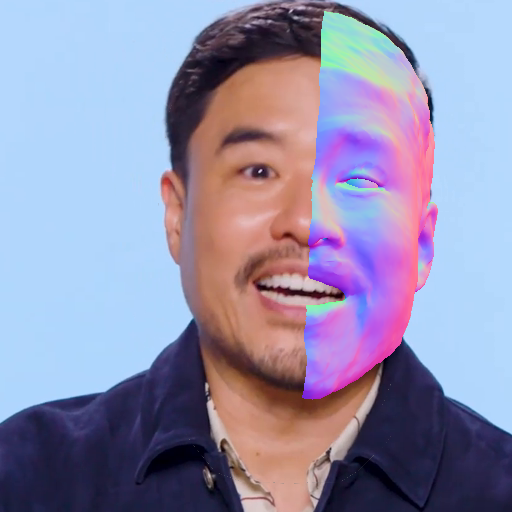}
        \end{tabular}
    \end{tabular}
    }
    \caption{Additional results of our method compared to multiple face reconstruction methods. From left to right: input image, DECA \cite{feng2021learning}, EMOCA \cite{danecek2022EMOCA}, EMOCA fine-tuned, SMIRK \cite{retsinasSMIRK3DFacial2024} and SPARK.}
    \label{fig:qualitative_more}
    \Description{A grid of images showing, from left to right, the ground truth image, the reconstruction using various real-time face tracking methods, our reconstruction, and our result overlaid on top of half of the input face. 16 examples are shown.}
\end{figure*}

\begin{figure*}[t]
    \renewcommand{\w}{0.098\linewidth}
    \setlength{\tabcolsep}{0.0em} % horizontal padding
    \def\arraystretch{0.6}{ % vertical padding
    \begin{tabular}{ccccccccccc}
        Ground truth  & Render & Albedo & Shading & Normals & \hspace{0.02\linewidth} &  Ground truth  & Render & Albedo & Shading & Normals \\
        \includegraphics[width=\w]{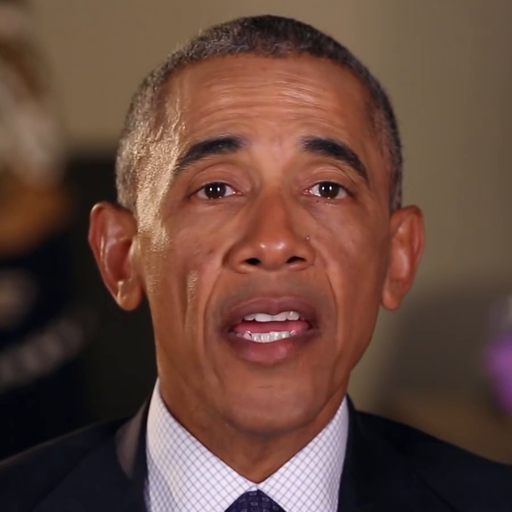} &
        \includegraphics[width=\w]{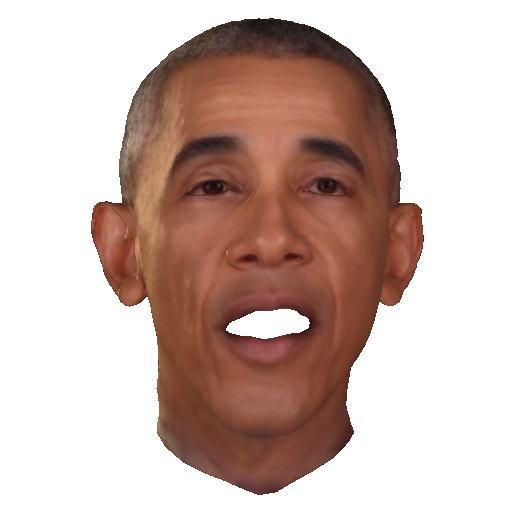} &
        \includegraphics[width=\w]{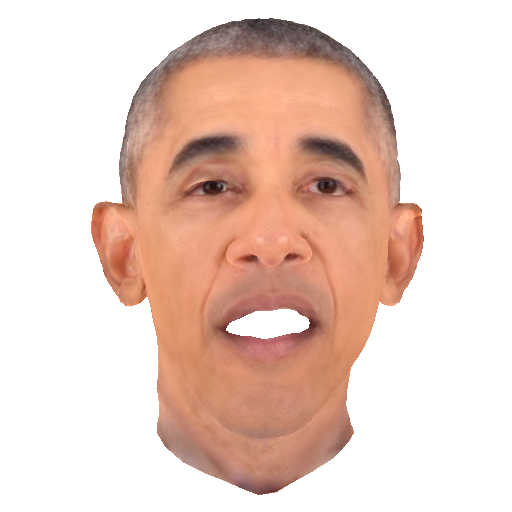} &
        \includegraphics[width=\w]{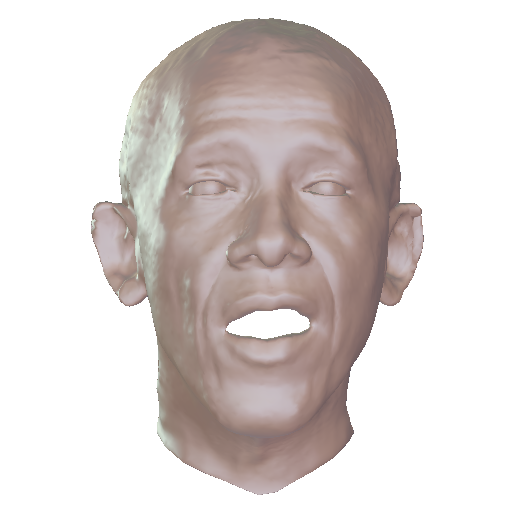} &
        \includegraphics[width=\w]{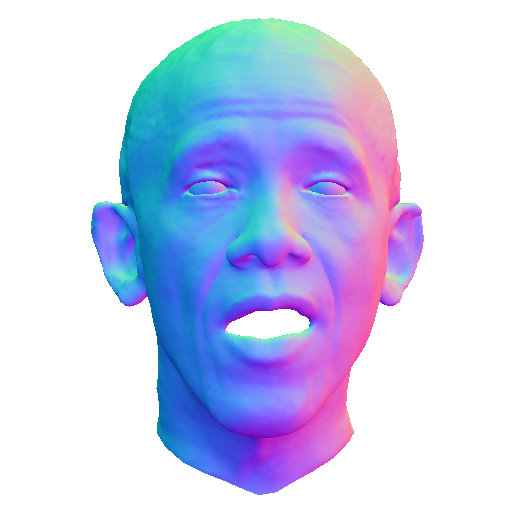} &
        &
        \includegraphics[width=\w]{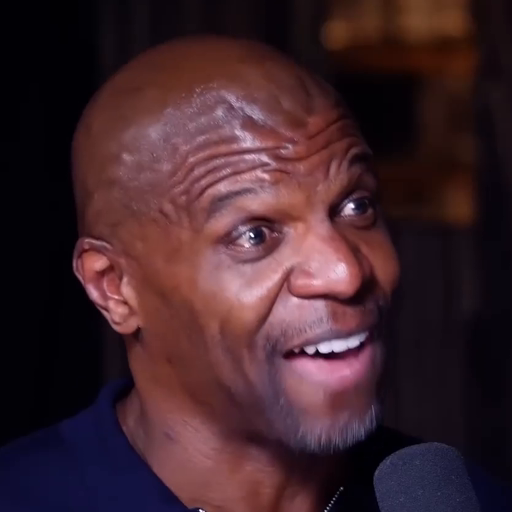} &
        \includegraphics[width=\w]{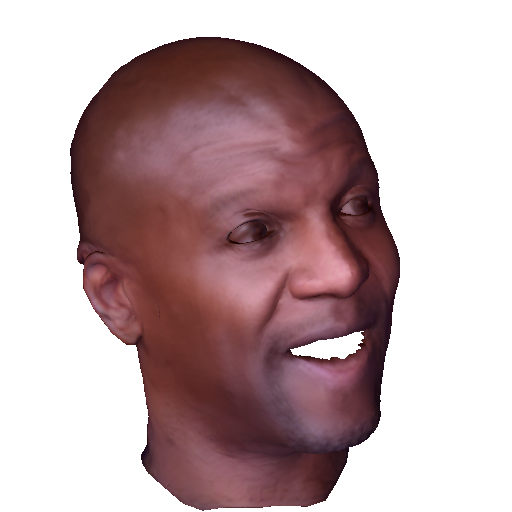} &
        \includegraphics[width=\w]{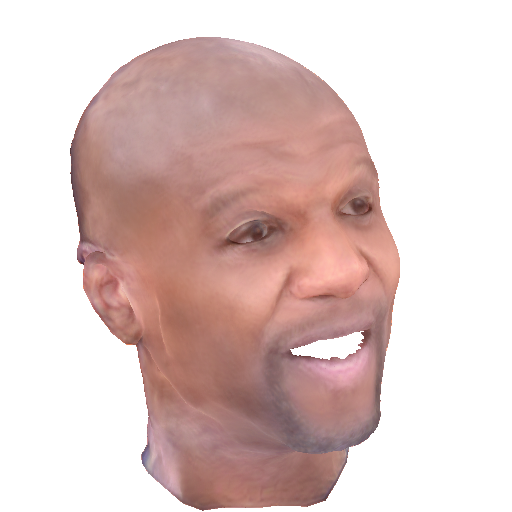} &
        \includegraphics[width=\w]{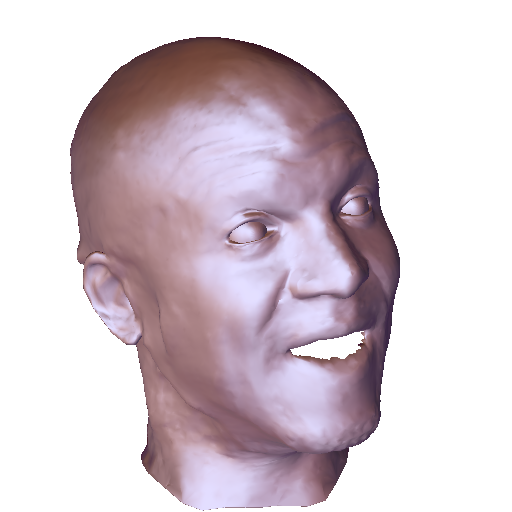} &
        \includegraphics[width=\w]{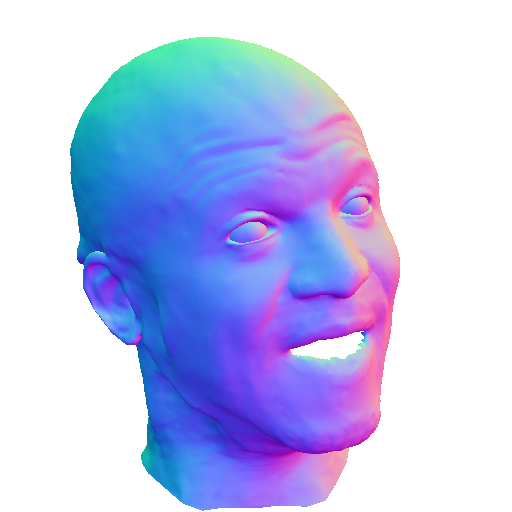}
        \\
        \includegraphics[width=\w]{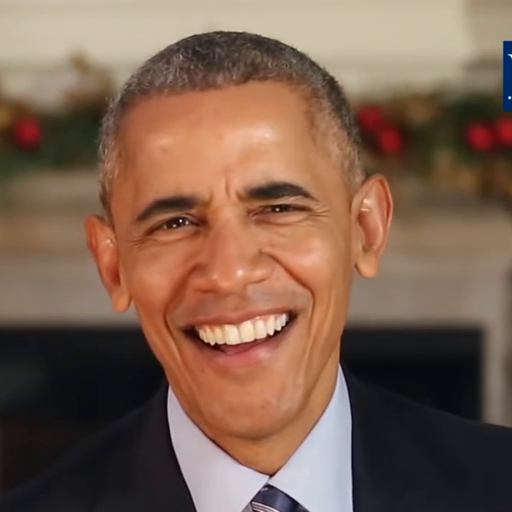} &
        \includegraphics[width=\w]{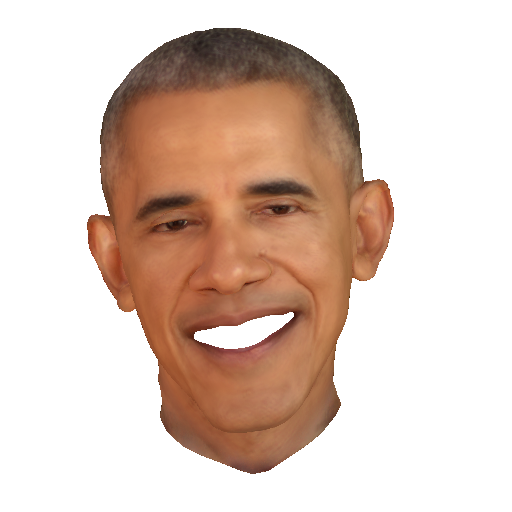} &
        \includegraphics[width=\w]{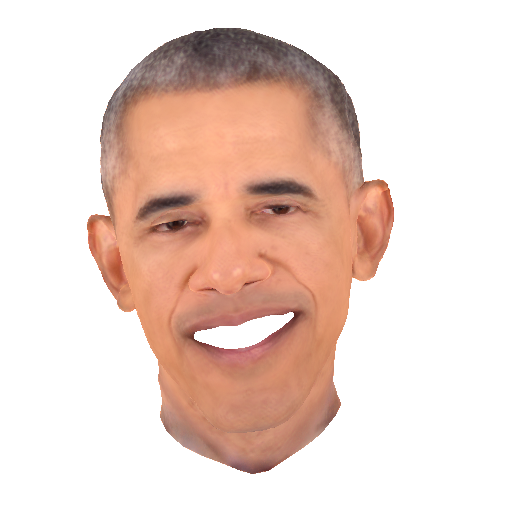} &
        \includegraphics[width=\w]{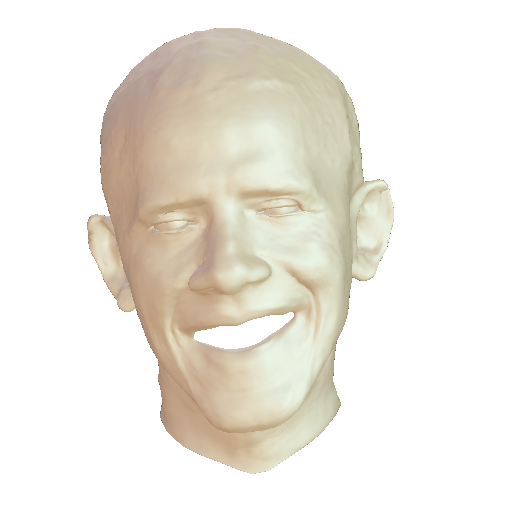} &
        \includegraphics[width=\w]{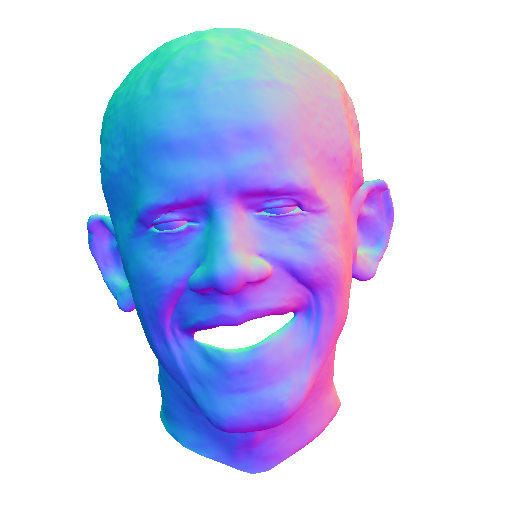} &
        &
        \includegraphics[width=\w]{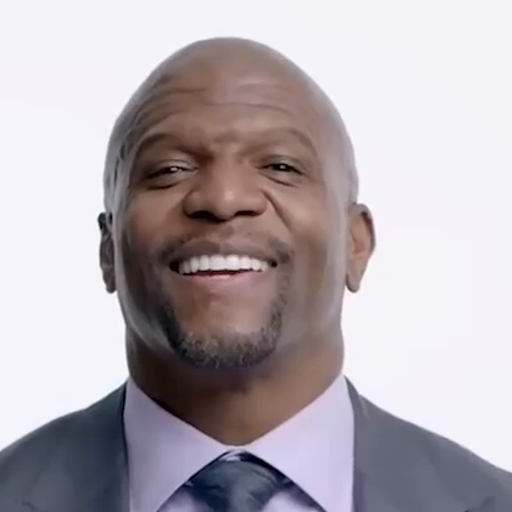} &
        \includegraphics[width=\w]{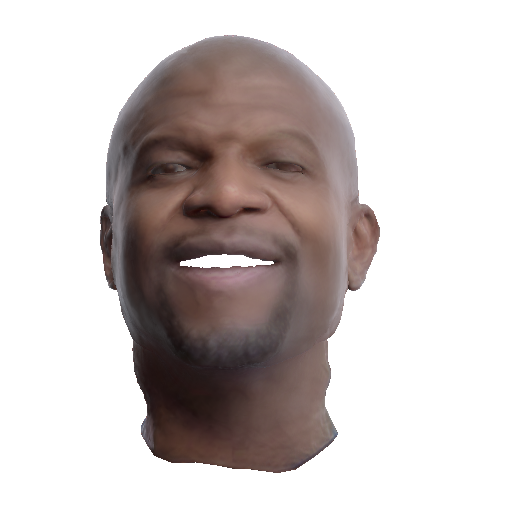} &
        \includegraphics[width=\w]{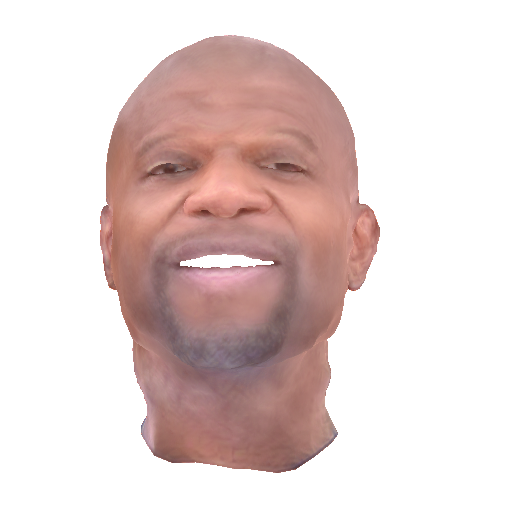} &
        \includegraphics[width=\w]{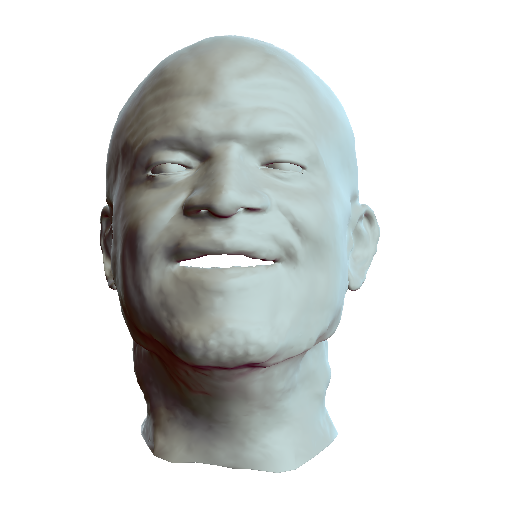} &
        \includegraphics[width=\w]{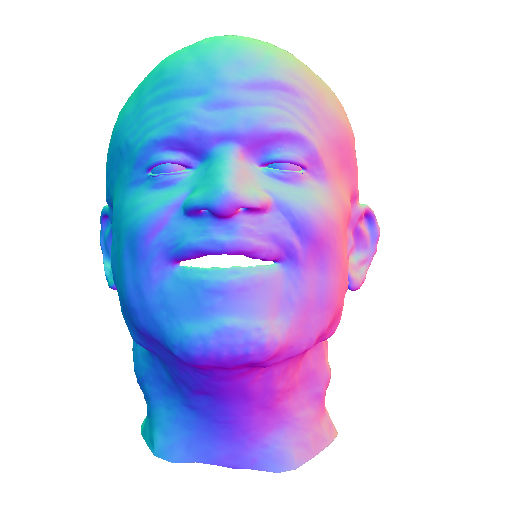}
        \\
        \\
        \includegraphics[width=\w]{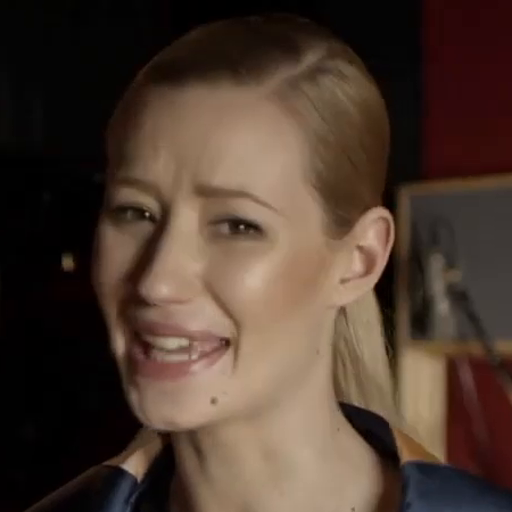} &
        \includegraphics[width=\w]{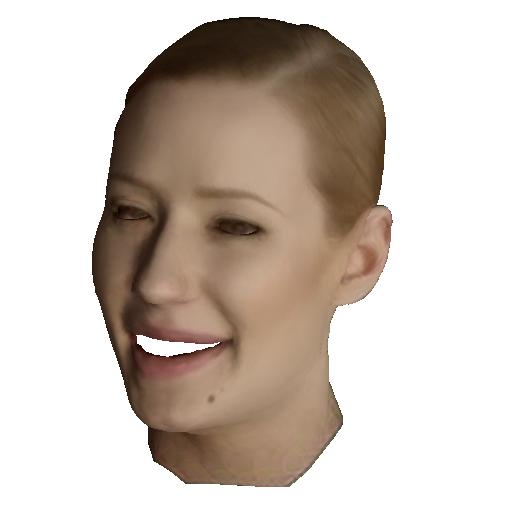} &
        \includegraphics[width=\w]{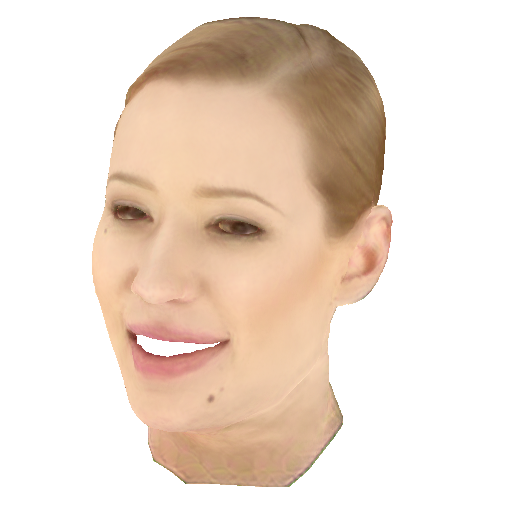} &
        \includegraphics[width=\w]{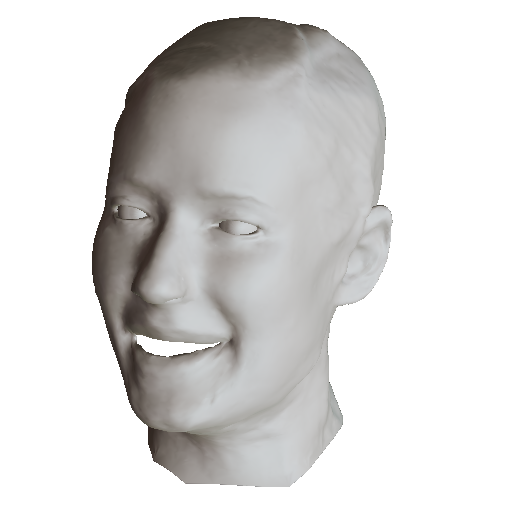} &
        \includegraphics[width=\w]{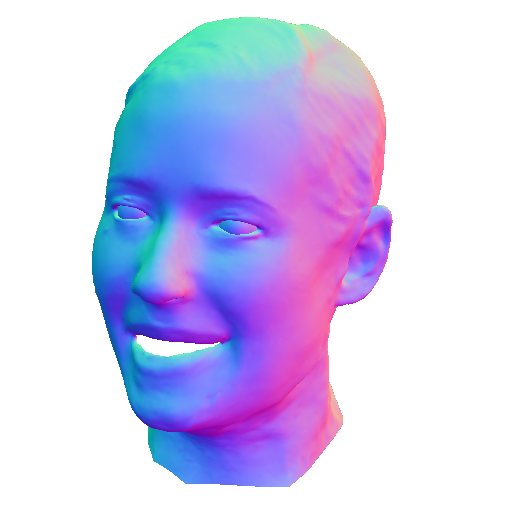} &
        &
        \includegraphics[width=\w]{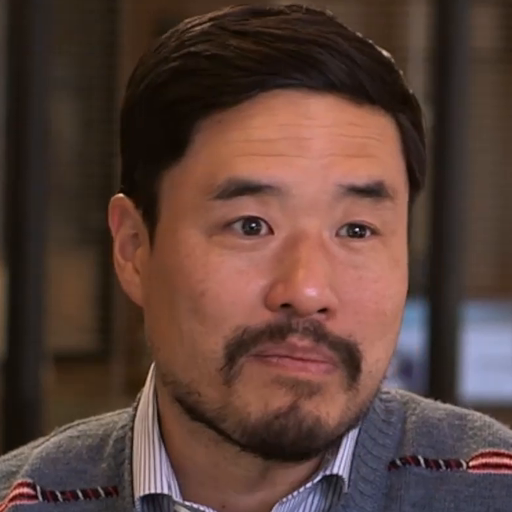} &
        \includegraphics[width=\w]{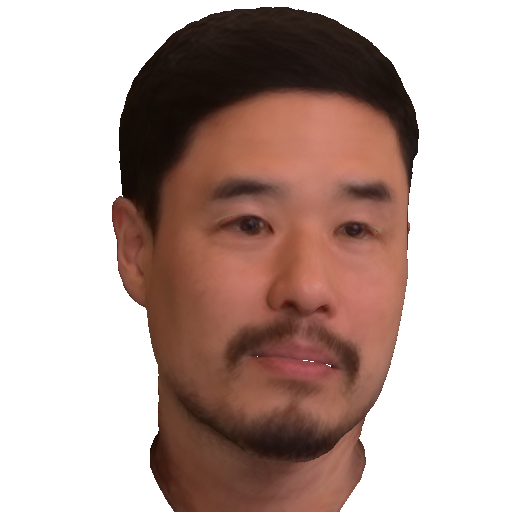} &
        \includegraphics[width=\w]{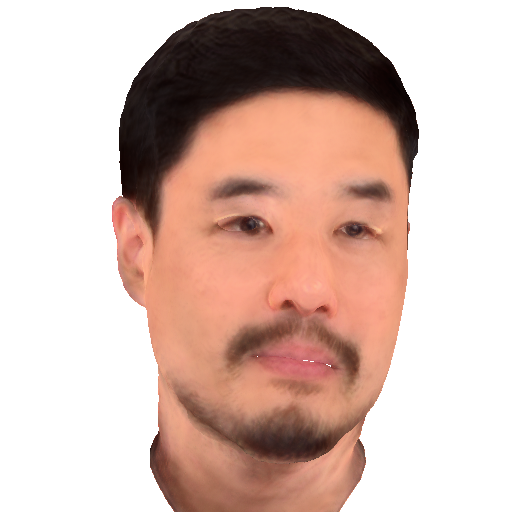} &
        \includegraphics[width=\w]{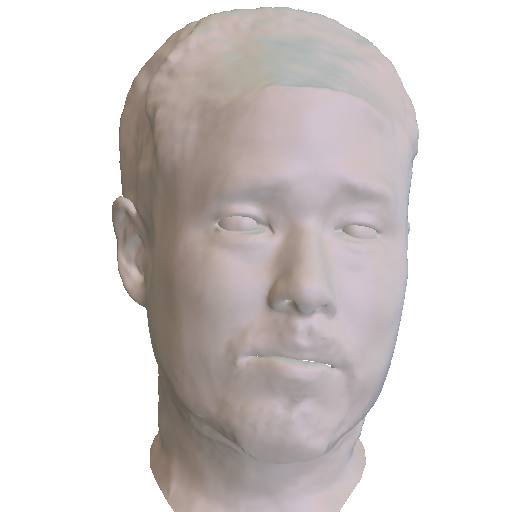} &
        \includegraphics[width=\w]{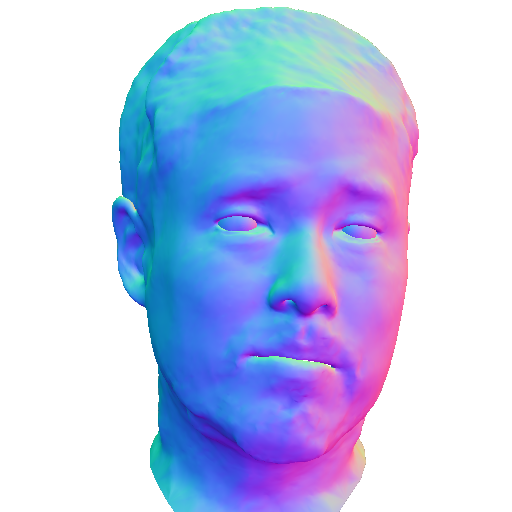}
        \\
        \includegraphics[width=\w]{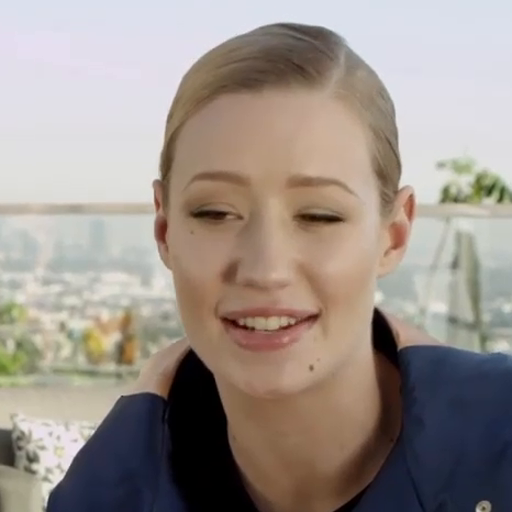} &
        \includegraphics[width=\w]{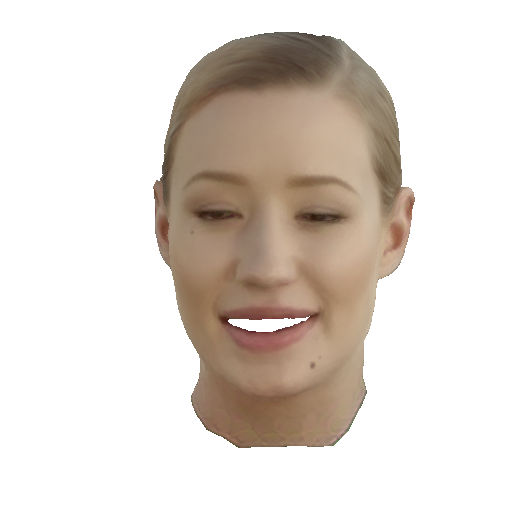} &
        \includegraphics[width=\w]{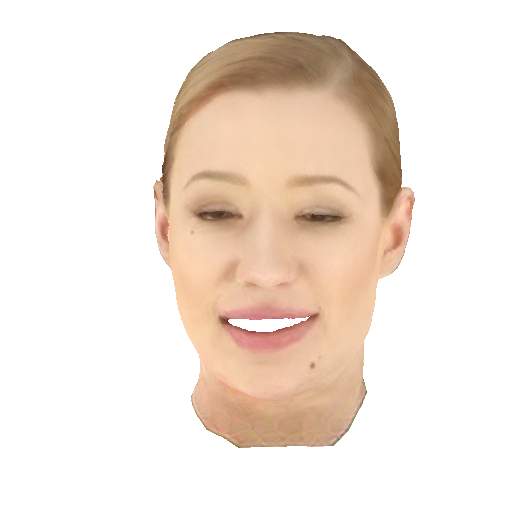} &
        \includegraphics[width=\w]{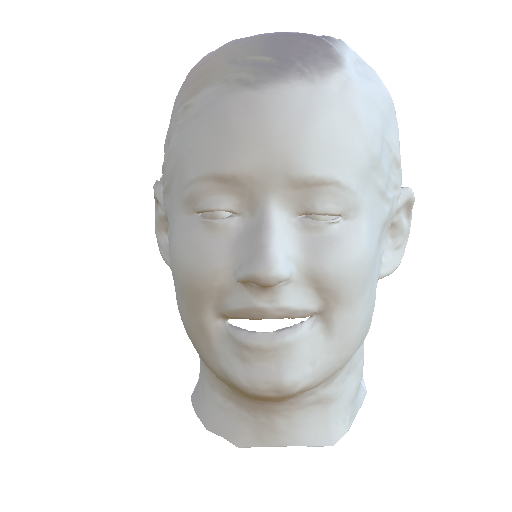} &
        \includegraphics[width=\w]{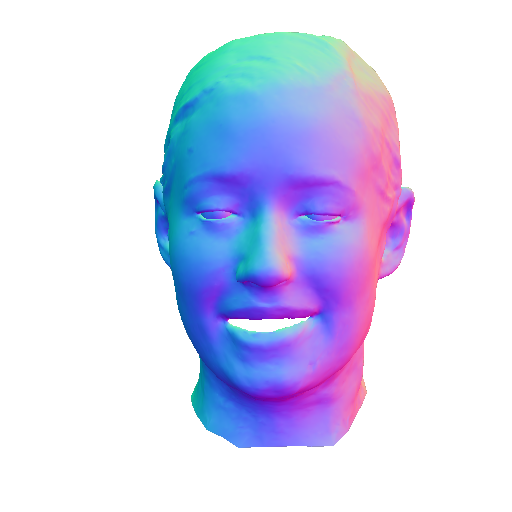} &
        &
        \includegraphics[width=\w]{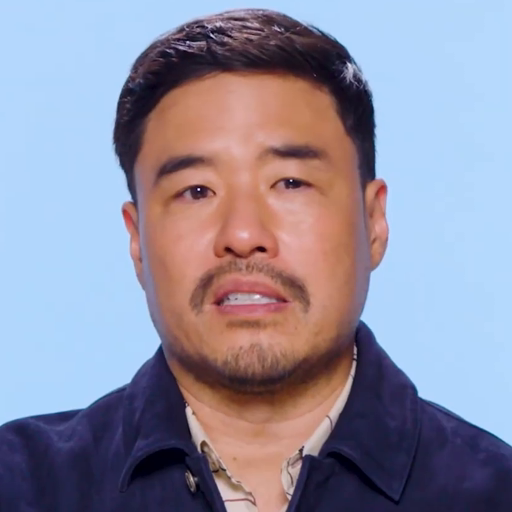} &
        \includegraphics[width=\w]{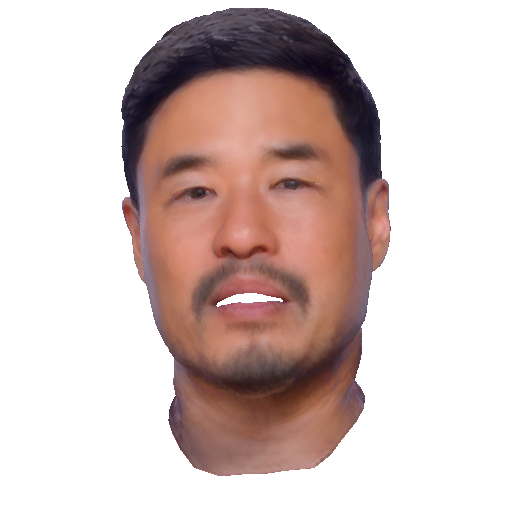} &
        \includegraphics[width=\w]{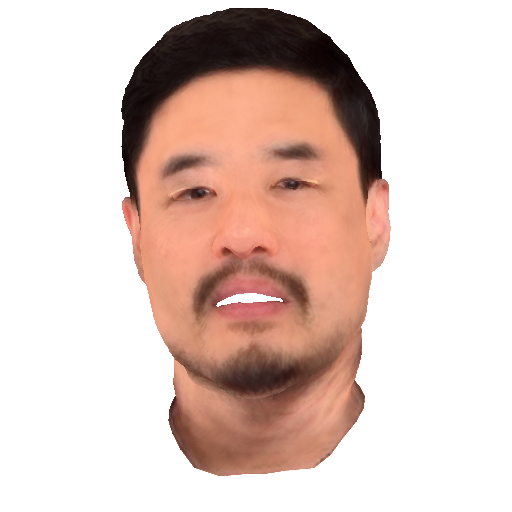} &
        \includegraphics[width=\w]{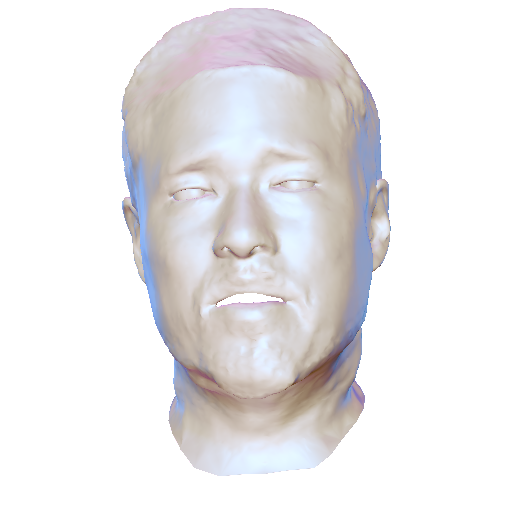} &
        \includegraphics[width=\w]{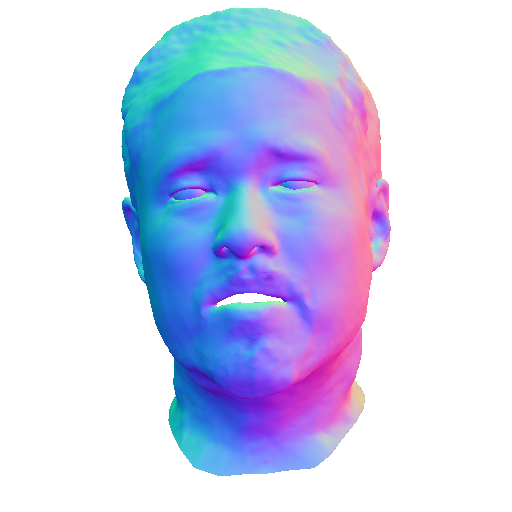}
    \end{tabular}
    }
    \caption{Qualitative results from \avatarStage{}. For a given subject, only the shading varies between sequences. While we only show samples for two sequences per subject, the method was trained with 4 to 6 sequences for each subject.}
    \label{fig:qualitative_multiflare}
    \Description{A grid of images showing the results of the avatar reconstruction stage of our method, MultiFLARE. From left to right are shown: the ground truth image, the final render, the albedo, shading and normals.}
\end{figure*}

\end{document}

\title{SPARK: Self-supervised Personalized Real-time Monocular Face Capture \ **Supplementary Material**}


\author{Kelian Baert}
\affiliation{%
  \institution{Technicolor Group}
  \city{Paris}
  \country{France}
}
\affiliation{%
  \institution{University Rennes}
  \city{Rennes}
  \country{France}
}
\email{kelian.baert@technicolor.com}

\author{Shrisha Bharadwaj}
\affiliation{%
  \institution{Max Planck Institute for Intelligent Systems}
  \city{Tübingen}
  \country{Germany}}
\email{shrisha.bharadwaj@tuebingen.mpg.de}

\author{Fabien Castan}
\affiliation{%
  \institution{Technicolor Group}
  \city{Paris}
  \country{France}
}
\email{fabien.castan@technicolor.com}

\author{Benoit Maujean}
\affiliation{%
  \institution{Technicolor Group}
  \city{Paris}
  \country{France}
}
\email{benoit.maujean@technicolor.com}

\author{Marc Christie}
\affiliation{%
  \institution{University Rennes, IRISA, CNRS, Inria}
  \city{Rennes}
  \country{France}}
\email{marc.christie@irisa.fr}

\author{Victoria F. Abrevaya}
\affiliation{%
  \institution{Max Planck Institute for Intelligent Systems}
  \city{Tübingen}
  \country{Germany}}
\email{victoria.abrevaya@tuebingen.mpg.de}

\author{Adnane Boukhayma}
\affiliation{%
  \institution{Inria, University Rennes, IRISA, CNRS}
  \city{Rennes}
  \country{France}}
\email{adnane.boukhayma@gmail.com}

\renewcommand{\shortauthors}{Baert et al.}

\maketitle

%%%%%%%%%%% sections %%%%%%%%%%
\section{Introduction}

In this supplemental document, we provide additional details on our training process, sources for our dataset and detailed quantitative results for each subject.
\section{Dataset}

We compile our dataset from online videos, for which we provide sources in Table \ref{tab:dataset}. We manually cut the videos to only include a single viewpoint of the subject, removing frames where the person is heavily occluded or not visible at all. In several cases, we sample multiple sequences from the same source video. For the purposes of our work, we treat those as different videos because of the entirely different illumination conditions.  

\section{Training details}

In this section, we provide more information on the training of both stages of our method.

\subsection{\avatarStage{}}

We train \avatarStage{} for 3000 iterations using a batch size of 4 images. The canonical geometry is remeshed \cite{Botsch2004ARA} to increase its resolution after 500 iterations. We use an Adam optimizer~\cite{kingma14} with a learning rate of $2\mathrm{e}{-4}$ for the deformation network $\mathcal{D}$ and canonical vertices $x_c$, $1e-3$ for the material network $\mathcal{M}$ and light MLPs $\mathcal{L}_i$. We also optimize pose and expression parameters with a learning rate of $1\mathrm{e}{-4}$. Training takes about 20 minutes on an NVIDIA RTX A5000 GPU.

\paragraph{Deformation network}
The deformation network $\mathcal{D}$ is initialized through a quick supervised learning, minimizing $||\mathcal{D}(\gamma(\text{x})) - \mathcal{E}||_2$ at the positions of the FLAME~\cite{FLAME:SiggraphAsia2017} template vertices. The deformer will initially act as a continuous representation of the FLAME expression basis $\mathcal{E}$ independent of topology. Note that we only observed a marginal difference when subdividing the FLAME template mesh for this stage, as the MLP's inductive bias already provides the deformer interpolation ability. We use the Adam optimizer with a learning rate of $2\mathrm{e}{-4}$ and train for 5000 iterations (using all vertex positions of the template mesh at every iteration), requiring under a minute on a NVIDIA RTX A5000 GPU.

\paragraph{Progressive multi-resolution hash encoding}
We embed the input of the material network $\mathcal{M}$ using multi-resolution hash encoding \cite{mueller2022instant}, with 16 resolutions levels ranging from 16 to 4096 and 2 features per level. These increasingly detailed levels are progressively enabled \cite{li2023neuralangelo}: one every 250 training iterations, and the remaining 8 at iteration 2000.

\paragraph{Network architectures}
The material network $\mathcal{M}$ and the illumination networks $\mathcal{L}_i$ all have 3 hidden layers of size 64, with ReLU activations. The final prediction of $\mathcal{M}$ is activated using Softplus ($\beta=100$), while $\mathcal{L}_i$ outputs use a Sigmoid activation. The deformer $\mathcal{D}$ has 4 hidden layers of size 128 with Softplus activations ($\beta=100$).

\subsection{Tracker adaptation}

We perform the transfer-learning stage using the EMOCA v2 \cite{danecek2022EMOCA, filntisis2023spectre} encoder. We freeze most of the encoder and only update the weights of the last ResNet-50 \cite{heDeepResidualLearning2016} block and the MLP head, for both the coarse shape encoder and the expression encoder. We train for a maximum of 30 epochs using the Adam optimizer with a learning rate of $1e-5$ and a batch size of $16$, using early stopping to minimize the risk of overfitting. We set weights $\lambda_{\text{emo}}$, $\lambda_{\text{eye}}$, $\lambda_{\text{mc}}$, $\lambda_{\psi}$ and $\lambda_{\text{lipr}}$ to the values provided by the authors' implementation and increase $\lambda_{\text{pho}}$ from $2$ to $100$. Doing so, we take advantage of our more accurate pre-estimated appearance model from \avatarStage{}. On our datasets, training is completed in 20 minutes on average, depending on early stopping.

The cumulative training time for both the \avatarStage{} and the transfer-learning stage is less than 45 minutes. Once trained, the expression basis from the deformation network can be precomputed and the convolutional encoder, a ResNet-50 with a MLP head, can perform inference in real-time on modern hardware.

\section{Detailed Results}

\begin{figure}
    \includegraphics[width=1.0\linewidth]{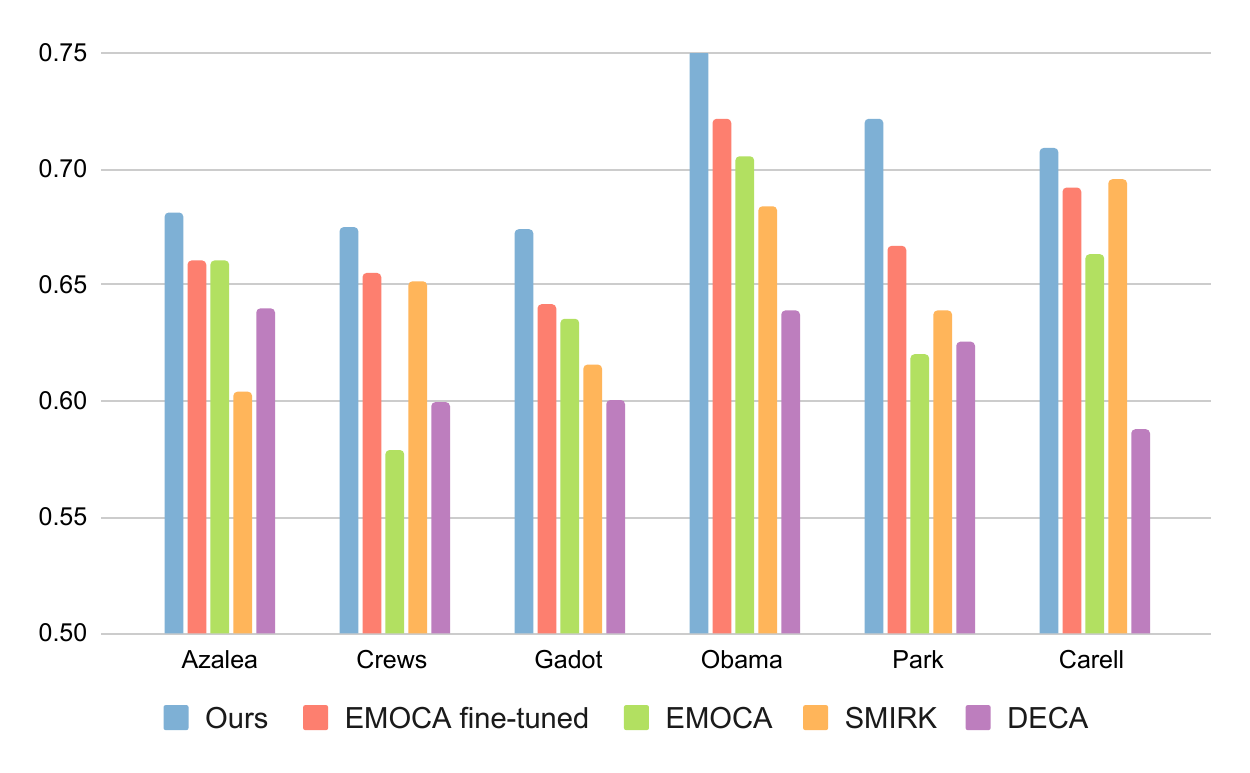}
    \includegraphics[width=1.0\linewidth]{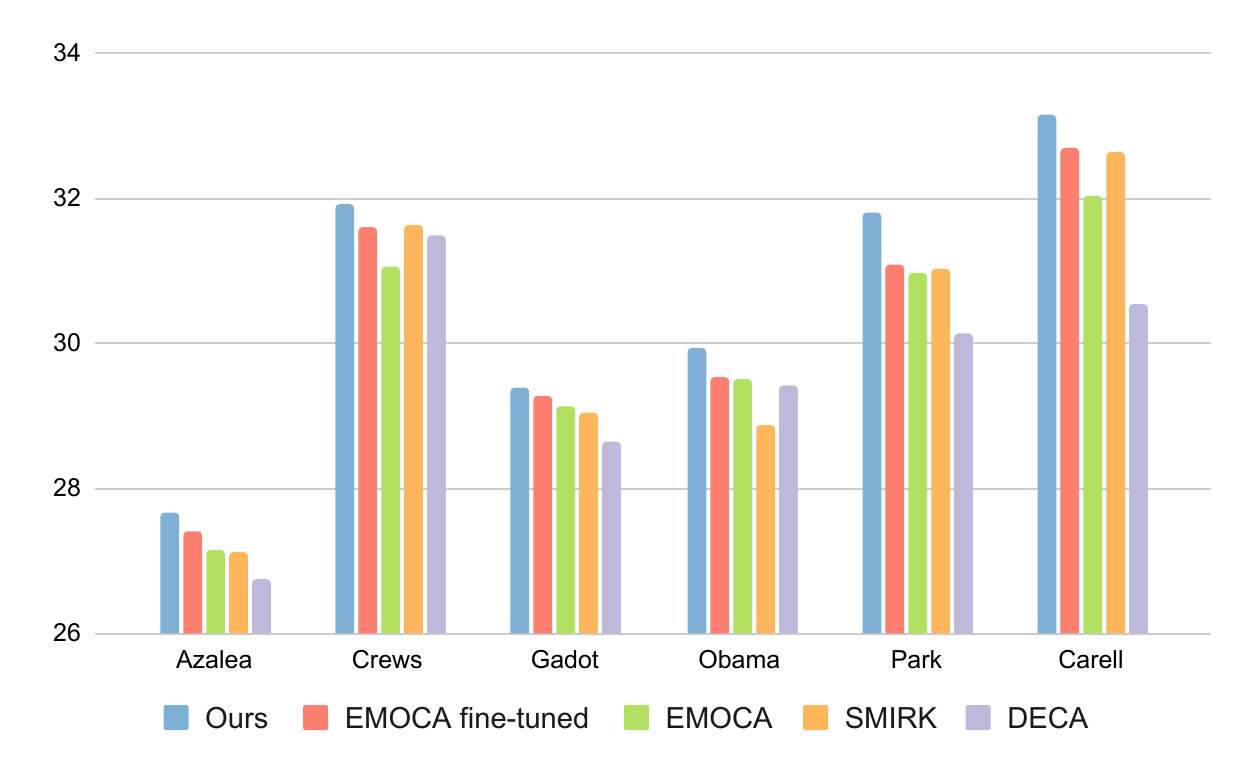}
    \caption{Results on 6 datasets for our semantic IoU (top) and geometry-based image warping PSNR (bottom) metrics, compared against several state-of-the-art methods. For each dataset, we perform cross-validation and average the results.}
    \label{fig:results_per_subject}
\end{figure}

\begin{figure}
    \includegraphics[width=1.0\linewidth]{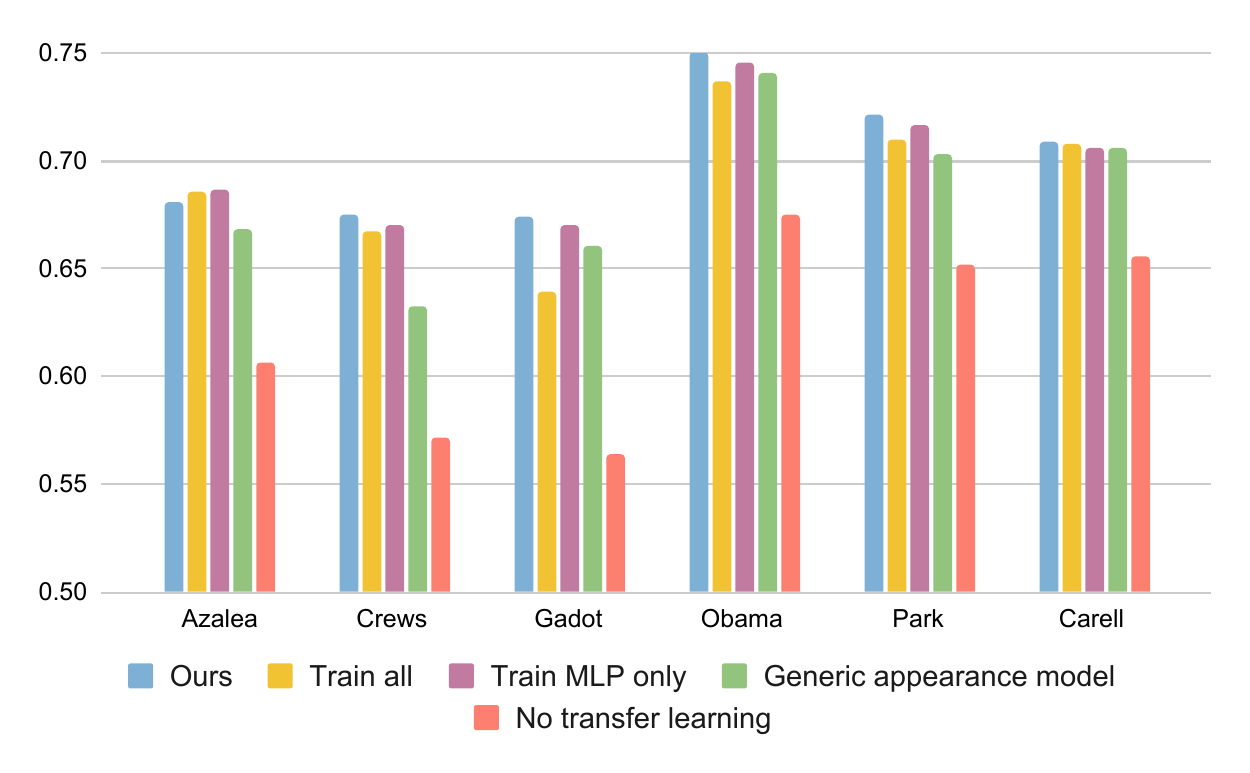}
    \includegraphics[width=1.0\linewidth]{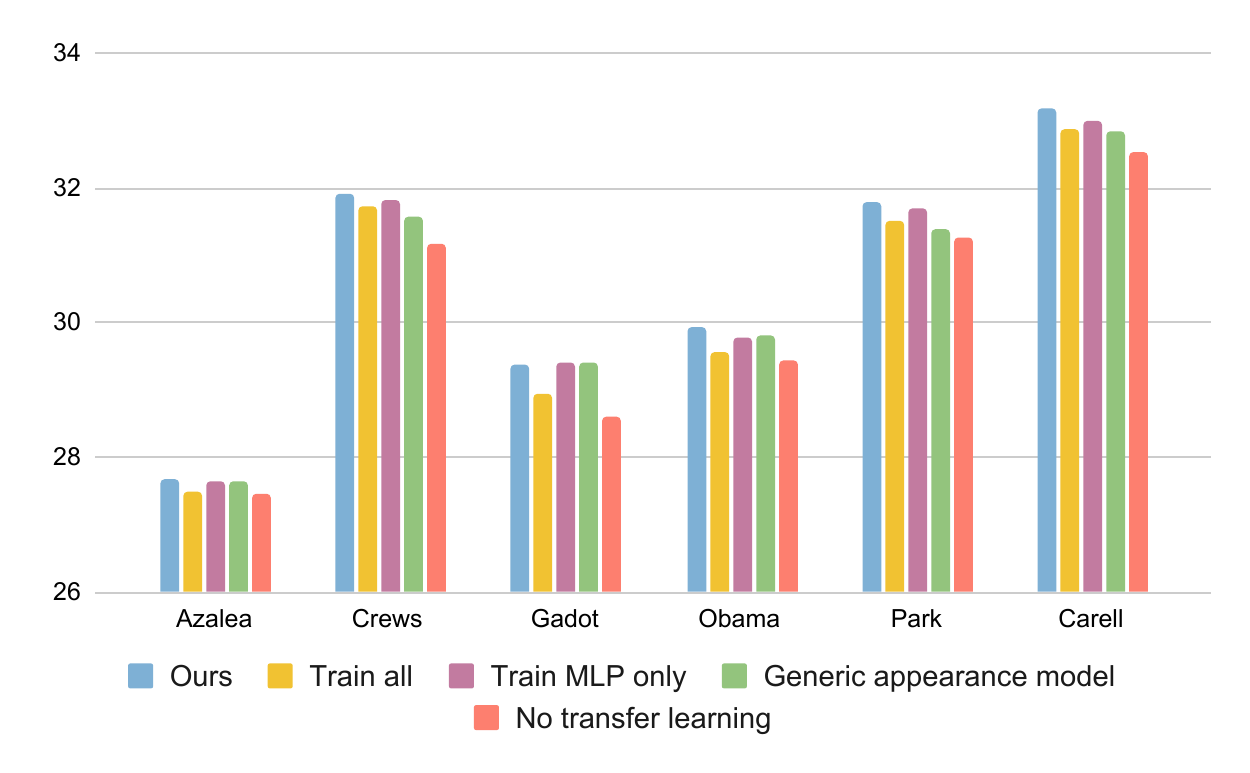}
    \caption{Results of our ablation study on 6 datasets for our semantic IoU (top) and geometry-based image warping PSNR metrics (bottom). For each dataset, we perform cross-validation and average the results.}
    \label{fig:results_ablation_per_subject}
\end{figure}

We report detailed per-subject results of our final personalized method for the semantic IoU and image warping metrics. Fig.~\ref{fig:results_per_subject} shows a comparison of our method against state-of-the-art approaches for monocular face reconstruction. Fig.~\ref{fig:results_ablation_per_subject} shows an ablation study of our method for various training schemes in the tracker adaptation stage.

\begin{table*}
    \setlength{\tabcolsep}{0.25em} % horizontal padding
    \def\arraystretch{0.9}{ % vertical padding
    \begin{tabular}{lccl}
        \toprule
        Subject & Video & Cut duration & Source \\
        \midrule
        \multirow{10}{*}{Barack Obama} & 1 & 01:00 & \url{https://www.youtube.com/watch?v=7G5kMmnAp_8} \\
    	& 2  & 00:43 & \url{https://www.youtube.com/watch?v=1CW6NbZvR_w} \\
    	& 3	 & 01:00 & \url{https://www.youtube.com/watch?v=lbwlVwNWLzU} \\
    	& 4	 & 01:00 & \url{https://www.youtube.com/watch?v=PtKBjuQblqI} \\
    	& 5	 & 01:00 & \url{https://www.youtube.com/watch?v=-XecjJTorNs} \\
    	& 6	 & 00:56 & \url{https://www.youtube.com/watch?v=_iKNE4ndBdg} \\
    	& 7	 & 01:00 & \url{https://www.youtube.com/watch?v=siyBp8Csugk} \\
    	& 8	 & 01:00 & \url{https://www.youtube.com/watch?v=aip0BAWrdLw} \\
    	& 9	 & 01:00 & \url{https://www.youtube.com/watch?v=1Z9tiSqHzkg} \\
    	& 10 & 01:00 & \url{https://www.youtube.com/watch?v=BIoBwblIxws} \\
    	\midrule
    	\multirow{7}{*}{Iggy Azalea} & 1 & 00:29 & \multirow{4}{*}{\url{https://www.youtube.com/watch?v=AMh5f8xRLRE}} \\
    	& 2	& 00:40 & \\
    	& 3	& 00:27 & \\
    	& 4	& 00:44 & \\
    	& 5	& 00:43 & \url{https://www.youtube.com/watch?v=-jlPXv6Ehks} \\
    	& 6	& 00:30 & \url{https://www.youtube.com/watch?v=YvB1o9wCO80} \\
    	& 7	& 00:40 & \url{https://www.youtube.com/watch?v=AogzdSQ5LVE} \\
    	\midrule
    	\multirow{6}{*}{Gal Gadot} & 1 & 00:32 & \multirow{4}{*}{\url{https://www.youtube.com/watch?v=OJJMVLPdAwY}} \\
    	& 2	& 00:13 & \\
    	& 3	& 00:29 & \\
    	& 4	& 00:30 & \\
    	& 5	& 00:17 & \url{https://www.youtube.com/watch?v=TJVf3KZAIG4} \\
    	& 6	& 00:12 & \url{https://www.youtube.com/watch?v=iX01L8wmhBk} \\
    	\midrule
    	\multirow{10}{*}{Terry Crews} & 1 & 01:12 & \multirow{2}{*}{\url{https://www.youtube.com/watch?v=eM1XfAsGnHI}} \\
    	& 2	 & 00:42 & \\
    	& 3	 & 00:47 & \url{https://www.youtube.com/watch?v=BvA_AinclK8} \\
    	& 4	 & 00:33 & \url{https://www.youtube.com/watch?v=fJjsbgk6e0A} \\
    	& 5	 & 00:59 & \url{https://www.youtube.com/watch?v=Im8itGAyZ6M} \\
    	& 6	 & 01:07 & \url{https://www.youtube.com/watch?v=hJeg601TIU4} \\
    	& 7	 & 00:49 & \url{https://www.youtube.com/watch?v=iux7NZ56Ei4} \\
    	& 8	 & 01:06 & \url{https://www.youtube.com/watch?v=PnoCfKY7L-s} \\
    	& 9	 & 00:28 & \url{https://www.youtube.com/watch?v=v0F3BhonHMM} \\
    	& 10 & 00:58 & \url{https://www.youtube.com/watch?v=PmbXC3dlK_s} \\
    	\midrule
    	\multirow{8}{*}{Randall Park} & 1 & 00:32 & \url{https://www.youtube.com/watch?v=ITzB_I0y_EM} \\
    	& 2	& 00:59 & \url{https://www.youtube.com/watch?v=LKqvgRAZjZg} \\
    	& 3	& 01:13 & \url{https://www.youtube.com/watch?v=nbfsEmeA9n4} \\
    	& 4	& 01:03 & \url{https://www.youtube.com/watch?v=wepD5rUZf8k} \\
    	& 5	& 00:43 & \url{https://www.youtube.com/watch?v=qT7nP4zQ05Y} \\
    	& 6	& 00:59 & \url{https://www.youtube.com/watch?v=mTPchtLBIJI} \\
    	& 7	& 00:50 & \url{https://www.youtube.com/watch?v=nbY34h54av0} \\
    	& 8	& 00:58 & \url{https://www.youtube.com/watch?v=2eQN4nCW25c} \\
    	\midrule
    	\multirow{12}{*}{Steve Carell} & 1 & 00:23 & \multirow{12}{*}{\textit{The Office} season 2} \\
    	& 2	 & 00:16 & \\
    	& 3	 & 00:22 & \\
    	& 4	 & 00:10 & \\
    	& 5	 & 00:15 & \\
    	& 6	 & 00:09 & \\
    	& 7	 & 00:08 & \\
    	& 8	 & 00:27 & \\
    	& 9	 & 00:18 & \\
    	& 10 & 00:30 & \\
    	& 11 & 00:19 & \\
    	& 12 & 00:29 & \\
        \bottomrule
    \end{tabular}
    }
    \caption{Sources for all videos of our dataset.}
    \label{tab:dataset}
\end{table*}

%%%% insert bib %%%%%%
% Bibliography
\bibliographystyle{ACM-Reference-Format}
\bibliography{bibliography}